\RequirePackage{fix-cm}
\documentclass[smallextended]{svjour3}       % onecolumn (second format)
\smartqed  % flush right qed marks, e.g. at end of proof
\usepackage{graphicx}
\usepackage{color}
\usepackage{subfigure}
\usepackage{longtable}
\usepackage{lscape}
\usepackage{multirow}
\usepackage{color}

\begin{document}

\title{Applications of Artificial Neural Networks in Microorganism Image Analysis: 
A Comprehensive Review from Conventional Multilayer Perceptron to 
Popular Convolutional Neural Network and Potential Visual Transformer}

\author{Jinghua Zhang $^{1, 3}$         \and
        Chen Li$^{1, *, \#}$ \and
        Yimin Yin $^{2}$\and 
        Jiawei Zhang $^{1}$\and
        Marcin Grzegorzek$^3$ %etc.
}

\titlerunning{Applications of Artificial Neural Networks in Microorganism Image Analysis...
} % if too long for running head

\institute{1. Microscopic Image and Medical Image Analysis Group, 
College of Medicine and Biological Information Engineering, 
Northeastern University, Shenyang, China \\
2. School of Mathematics and Statistics, Hunan First Normal University, Changsha, China \\ 
3. Institute for Medical Informatics, University of Luebeck, Luebeck, Germany \\ 
$*$. Co-first author: Chen Li \\
$\#$. Corresponding author: Chen Li, \email{lichen201096@hotmail.com}       
}

\date{Received: date / Accepted: date}

\maketitle

\begin{abstract}
Microorganisms are widely distributed in the human daily living environment. They play an essential role in environmental pollution control, disease prevention and treatment, and food and drug production. The analysis of microorganisms is essential for making full use of different microorganisms. The conventional analysis methods are laborious and time-consuming. Therefore, the automatic image analysis based on artificial neural networks is introduced to optimize it. However, the automatic microorganism image analysis faces many challenges, such as the requirement of a robust algorithm caused by various application occasions, insignificant features and easy under-segmentation caused by the image characteristic, and various analysis tasks. Therefore, we conduct this review to comprehensively discuss the characteristics of microorganism image analysis based on artificial neural networks. In this review, the background and motivation are introduced first. Then, the development of artificial neural networks and representative networks are presented. After that, the papers related to microorganism image analysis based on classical and deep neural networks are reviewed from the perspectives of different tasks. In the end, the methodology analysis and potential direction are discussed.

\keywords{Microorganism image analysis \and Classical neural network \and Deep neural network \and Transfer learning}

\end{abstract}

\section{Introduction}
\label{intro}
\begin{figure}[htbp!]
\centering
\includegraphics[width=0.98\linewidth]{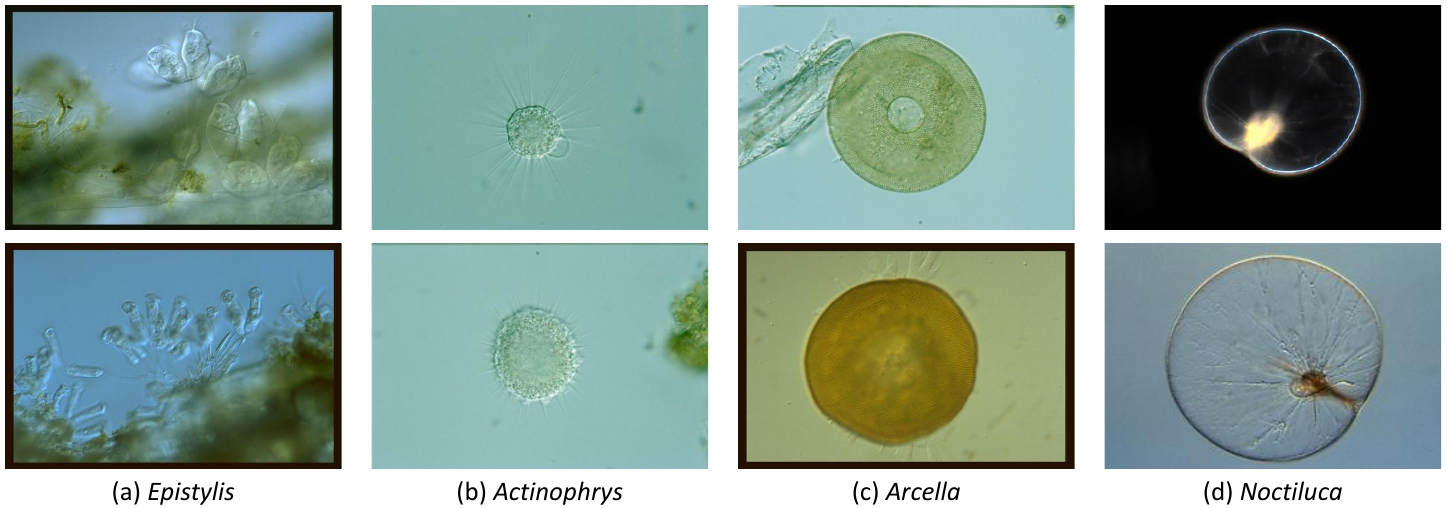}
\caption{The environmental microorganism images provided in~\cite{Li-2021-EMDS}.}
\label{fig1}
\end{figure}

Microorganisms are tiny living organisms that can appear as unicellular, multicellular, and acellular types~\cite{Madigan-1997-BBM}. Their examples are shown in Fig.~\ref{fig1}. Some microorganisms are benefiting, such as \emph{Lactobacteria} can decompose substances to give nutrients to plants~\cite{Kulwa-2019-ASSM}, \emph{Actinophrys} can digest the organic waste in sludge and increase the quality of freshwater~\cite{Zhang-2020-AMCF}, and \emph{Rhizobium leguminosarum} can help soybean to fix nitrogen and supply food to human beings~\cite{Bagyaraj-2007-AM}. However, there are also many harmful microorganisms, such as \emph{Mycobacterium tuberculosis} can lead to disease and death~\cite{Gillespie-2012-MMIA} and the novel coronavirus disease 2019 (COVID-19) constitutes a public health emergency globally~\cite{Rahaman-2020-ICSC}. Therefore, microorganism research plays a vital role in pollution monitoring, environmental management, medical diagnosis, agriculture, and food production~\cite{Kulwa-2019-ASSM,Li-2019-ASAC}, and the analysis of microorganisms is essential for related research and applications~\cite{Li-2020-MAMR}.

In general, microorganism analysis methods can be summarised into four categories: chemical (e.g., chemical component analysis), physical (e.g., spectrum analysis), molecular biological (e.g., DNA and RNA analysis), and morphological (e.g., manual observation under a microscope) methods~\cite{Li-2019-ASAC}. Their main advantages and disadvantages are compared in Table~\ref{table1}. The chemical method is highly accurate but often results in secondary pollution of chemical reagents~\cite{Li-2019-ASAC}. The physical method also has high accuracy, but it requires expensive equipment~\cite{Li-2019-ASAC}. The molecular biological method distinguishes microorganisms by sequence analysis of the genome~\cite{Yamaguchi-2015-SDCR}. This strategy needs expensive equipment, plenty of time, and professional researchers. The morphological method is the most direct and brief approach, where a microorganism is observed under a microscope and recognized manually based on its shape~\cite{Pepper-2011-EM}. The morphological method is the most cost-effective of the above methods, but it is still laborious, tedious, and time-consuming~\cite{Zhang-2020-AMCF}. Besides, the objectivity of this manual analyzing process is unstable, depending on the experience, workload, and mood of the biologist significantly. Therefore, developing an automatic microorganism image analysis system is of great significance. Nevertheless, it faces many challenges. As the microorganism examples are shown in Fig.~\ref{fig1}, It can be found that the backgrounds of \emph{Epistylis} contain a lot of noise and impurities. \emph{Actinophrys} has lots of filopodia that are easy to be under-segmented. The examples of \emph{Arcella} and \emph{Noctiluca} demonstrate that different light conditions can lead to entirely different color features. The high transparency of \emph{Noctiluca} results in inconspicuous texture features. However, broad application occasions require the development of flexible and robust \emph{Microorganism Image Analysis} (MIA) algorithms. Additionally, the related algorithms are various due to the numerous analysis tasks.

\begin{table}[htbp!]
\caption{A comparison of traditional methods for microorganism analysis~\cite{Li-2016-CMIA}.}
\label{table1}
\renewcommand\arraystretch{1.6}
\setlength{\tabcolsep}{5pt}
\centering
\begin{tabular}{ccc}
\hline
\textbf{Methods}                                                      & \textbf{Advantages} & \textbf{Disadvantages}                                                                                                    \\ \hline
Chemical method                                                         & High accuracy       & Secondary pollution of chemical reagent                                                                                   \\
Physical method                                                       & High accuracy       & Expensive equipment                                                                                                       \\
\begin{tabular}[c]{@{}c@{}}Molecular biological\\ method\end{tabular} & High accuracy       & \begin{tabular}[c]{@{}c@{}}Secondary pollution of chemical reagent,\\ expensive equipment, long testing time\end{tabular} \\
Morphological method                                                  & Short testing time  & Long training time for skillful operators                                                                                 \\ \hline
\end{tabular}
\end{table}

\emph{Artificial Intelligence} (AI) technology has developed rapidly in recent years \cite{Zhou-2020-ACRB}. It achieves outstanding performance in many fields of image analysis and processing, such as autonomous driving~\cite{Sallab-2017-DRLF,Chen-2017-M3OD,Wang-2019-PVDE}, face recognition~\cite{Liu-2017-SDHE,Wang-2018-CLMC,Deng-2019-AAAM}, and disease diagnosis~\cite{Sun-2020-GHIS,Xue-2020-AATF,Zhao-2020-ASSD,Ai-2021-ASRG}. AI can undertake laborious and time-consuming work and quickly extract valuable information from image data. Therefore, AI shows potential in MIA. Besides, AI has a robust objective analysis ability in MIA and can avoid subjective differences caused by manual analysis. To some extent, the misjudgment of biologists can be reduced, and the efficiency can be improved.

As an essential part of artificial intelligence technology, \emph{Artificial Neural Network} (ANN) is originally designed according to the biological neuron~\cite{Mcculloch-1943-ALCI}. Due to the limitation of computer performance, the difficulty of training, and the popularization of \emph{Support Vector Machine} (SVM), the early ANN development once fall into a state of stagnation~\cite{Yamashita-2016-AIGD}. After that, with the improvement of computer performance, \emph{Convolutional Neural Network} (CNN) shows an overwhelming advantage in image recognition~\cite{Krizhevsky-2017-ICDC}, and ANN is paid attention to again and developed rapidly. We find that ANNs are widely used in MIA thorough investigation because they can learn valuable patterns from enormous data and features.

\subsection{Artificial Intelligence Methods for Microorganism Image Analysis}

\begin{figure}[htbp!]
\centering
\includegraphics[width=1\linewidth]{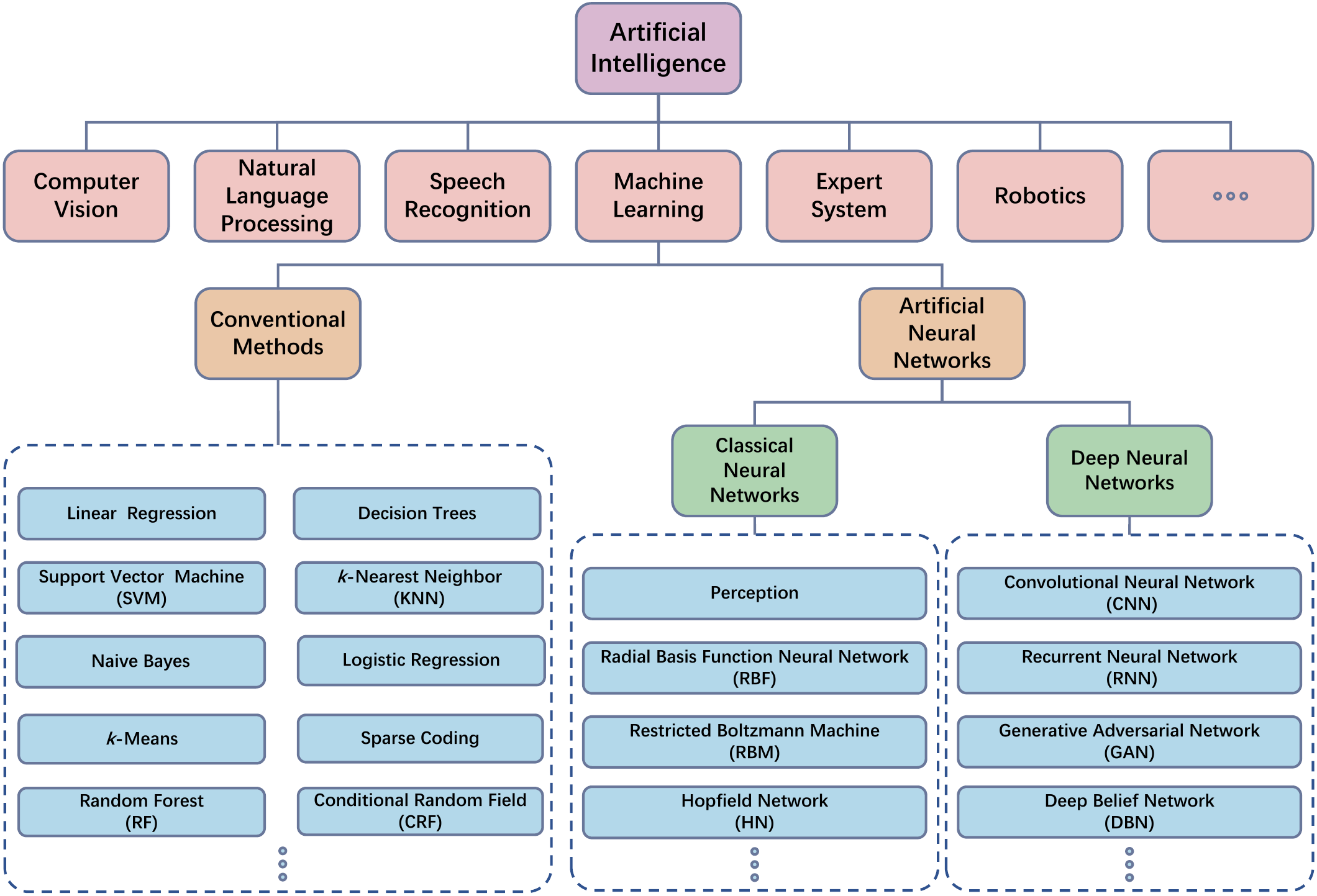}
\caption{The structure of ML in the AI.}
\label{fig2}
\end{figure}

AI is an umbrella term encompassing the techniques for a machine to mimic or go beyond human intelligence, primarily in cognitive capabilities~\cite{Robertson-2018-DIAB}. Fig.~\ref{fig2} provides the structure of AI technology~\cite{Linnosmaa-2020-MLSC,Zhou-2020-ACRB}. AI has several important sub-domains, such as \emph{Machine learning} (ML), computer vision, natural language processing. Among them, ML technology has been widely used in the MIA task, such as \emph{Environmental Microorganism} (EM) segmentation~\cite{Li-2020-MAMR,Zhang-2020-AMCF}, \emph{Herpesvirus} detection~\cite{Devan-2019-DHCT}, and \emph{Tuberculosis Bacilli} (TB) classification~\cite{Rulaningtyas-2011-ACTB,Osman-2012-OSEL}. 

As shown in Fig.~\ref{fig2}, ML can be grouped into conventional methods and ANNs. In conventional methods, SVM, \emph{$k$-Nearest Neighbor} (KNN), \emph{Random Forest} (RF), and other methods have been applied to the MIA task. For example, work~\cite{Li-2015-ACIA} proposes an automatic EM classification system based on content-based image analysis techniques. Four features (histogram descriptor, geometric feature, Fourier descriptor, and internal structure histogram) are extracted from the Sobel edge detector-based segmentation result for training SVM to perform the classification task. For ten EM classes tested in this work, the mean of average precisions obtained by the system amounts to 94.75\%~\cite{Kulwa-2019-ASSM}. An automatic identification approach of TB is proposed in~\cite{Forero-2003-AITT}. The Canny edge technique is applied for edge detection, followed by non-maxima suppression and hysteresis threshold operations. After that, a morphological closing operation is applied. Then compactness and eccentricity features are extracted in one branch, and the same segmented images are passed through $k$-means clustering in the second branch. Each branch independently goes to the classification part using the nearest neighbor classifier. The average results obtained from the two branches are 93.30\% and 100\% sensitivity for the first and second branches~\cite{Kulwa-2019-ASSM}. In~\cite{Di-2011-UZAS}, a MIA work based on the ZooImage automated system is introduced, where 1437 objects of 11 taxa are classified using four shape features and a RF classifier. The classification accuracy of 83.92\% is finally achieved~\cite{Li-2016-CMIA}.

ANNs also play a vital role in the MIA task. Fig.~\ref{fig2} shows that they include classical neural networks and deep neural networks. In the early years, due to the computer performance limitation, classical neural networks, represented by MLP, \emph{Radial Basis Function Neural Network} (RBF), and \emph{Probabilistic Neural Network} (PNN), are applied to the MIA tasks. For example, in~\cite{Culverhouse-1996-ACFD}, human experts' performance in identifying \emph{Dinoflagellates} is compared to that achieved by two ANN classifiers (MLP and RBF) and two other statistical techniques, KNN and \emph{Quadratic Discriminant Analysis} (QDA). The data set used for training and testing comprised a collection of 60 features that are extracted from the specimen images. The result shows that the ANN classifiers outperform classical statistical techniques. Extended trials show that the human experts achieve 85\% accuracy while the RBF achieves the best performance of 83\%, the MLP 66\%, KNN 60\%, and the QDA 56\%. The work~\cite{Kumar-2010-RDMU} uses PNN to select the best identification parameters of the features extracted from the microorganism images. PNN is then used to classify the microorganisms with a 100\% accuracy using nine identification parameters.

Later, with the significant improvement of computer performance, the development of neural network theory, and the proposal of CNN, deep neural networks show an overwhelming advantage in image analysis, including microbial image analysis. For example, the work~\cite{Matuszewski-2018-MATS} uses U-Net to perform the \emph{Rift Valley} virus segmentation and achieves a Dice score of 90\% and \emph{Intersection Over Union} (IOU) of 83.1\%. In~\cite{Wahid-2019-DCNN}, transfer learning based on Xception is applied to perform the bacterial classification. Seven varieties of bacteria for recognition that might be lethal for humans are chosen for the experiment. The performance is evaluated on 230 bacteria images of seven varieties from the test dataset, which shows promising performance with approximately 97.5\% prediction accuracy in bacteria image classification.

In conclusion, we can find that conventional machine learning methods and classical neural network methods have similar workflows in the MIA task. They typically rely heavily on feature engineering~\cite{Al-2015-PEHC}. These workflows usually contain image acquisition, image preprocessing, segmentation, feature extraction, classifier design, and evaluation. The reliability of accuracy depends on the design and extraction of features~\cite{Al-2015-PEHC}. In recent years, with the development and popularization of CNN, which is one of the most important parts of deep neural networks, the MIA task can work without feature engineering. Compared with the classical ANN method, CNN can directly extract valuable features from the image through the convolutional kernel. This kind of ability makes the research and application of CNN in MIA increase rapidly and obtain overwhelming advantages.

\subsection{Motivation of This Review}
This paper focuses on the development and application of ANNs in the MIA task. A comprehensive overview of techniques for the image analysis of microorganisms using classical and deep neural networks is presented. The motivation is to clarify the development history of ANNs, understand the popular technology and trend of ANN applications in the MIA field. Besides, this paper also discusses potential techniques for the image analysis of microorganisms by ANNs. To our best knowledge, some review papers summarize research related to the MIA task, such as papers~\cite{Kulwa-2019-ASSM,Li-2020-ARCM,Li-2019-ASAC,Li-2018-ABRC,Zhang-2021-ACRI}. In the following part, we go through these review papers.

The review~\cite{Kulwa-2019-ASSM} comprehensively analyzes the various studies focusing on microorganism image segmentation methods from 1989 to 2018. These methods include classical methods (e.g., edge-based, threshold-based, and region-based segmentation methods) and machine learning-based methods (supervised and unsupervised machine learning methods). About 85 related papers are summarized in this review. The ANN-based segmentation method is only one part of this review. In the review~\cite{Li-2020-ARCM}, to clarify the potential and application of different clustering techniques in MIA, the related works from 1997 to 2017 are summarized while pinning out the specific challenges on each work (area) with the corresponding suitable clustering algorithm. More than 20 related research papers are summarized in this review. The review~\cite{Li-2019-ASAC} summarizes the development history of microorganism classification using content-based microscopic image analysis approaches. It introduces the classification methods from different application domains, including agricultural, food, environmental, industrial, medical, water-borne, and scientific microorganisms. Around 240 related works are summarized. The classification methods discussed in this review contain ANNs and many ML methods like SVM, KNN, and RF. \cite{Zhang-2021-ACRI} proposes a comprehensive review focusing on microorganism counting based on image analysis approaches. It summarizes more than 144 related papers from 1980 to 2020. The image analysis approaches include both classical image analysis methods and deep learning methods. ANN is only a part of the methods discussed in this review.

To sum up, the review~\cite{Li-2020-ARCM} only summarizes the MIA method based on clustering techniques. The review papers~\cite{Kulwa-2019-ASSM,Li-2019-ASAC,Zhang-2021-ACRI} focus on segmentation, classification, and counting tasks. Although the methods discussed include some ANN methods, they are not the central part of these papers. Hence, to comprehensively understand the development history, popular technology, and trend of ANNs in the MIA field, we conduct this review based on our previous work~\cite{Li-2018-ABRC}.

Our previous work~\cite{Li-2018-ABRC} proposes a brief review for content-based MIA using classical and deep neural networks. It briefly summarizes 55 related papers from 1992 to 2017. To illustrate the differences, the characteristics of our current review and previous work are summarized in Tab.~\ref{table1.5}. The current review summarizes 96 papers related to classification, segmentation, detection, counting, feature extraction, image enhancement, and data augmentation from 1992 to 2020. The current review comprehensively analyzes the applications of ANN in the MIA field by providing the introduction of representative ANNs and the discussion of the method in each paper. Besides, it also provides summary tables, statistic analysis, and potential directions.

\begin{table}[htbp!]
\caption{A comparison of our current review and previous work~\cite{Li-2018-ABRC}.}
\label{table1.5}
\renewcommand\arraystretch{1.6}
\setlength{\tabcolsep}{1.5pt}
\begin{tabular}{cccccc}
\hline
\textbf{Characteristic} & \textbf{Previous} & \textbf{Current} & \textbf{Characteristic} & \textbf{Previous} & \textbf{Current} \\ \hline
Time Range              & 1992-2017         & 1992-2020        & Paper Number            & 55                & 96               \\
Related Task            & 3                 & 7                & Method Analysis         & Part              & Comprehensive    \\
Paper Detail            & Part              & Comprehensive    & Summary Table           & No                & Yes              \\
Statistic Analysis      & No                & Yes              & Potential Direction     & No                & Yes              \\ \hline
\end{tabular}
\end{table}

\subsection{Paper Searching and Screening}
\begin{figure}[htbp!]
\centering
\includegraphics[width=0.75\linewidth]{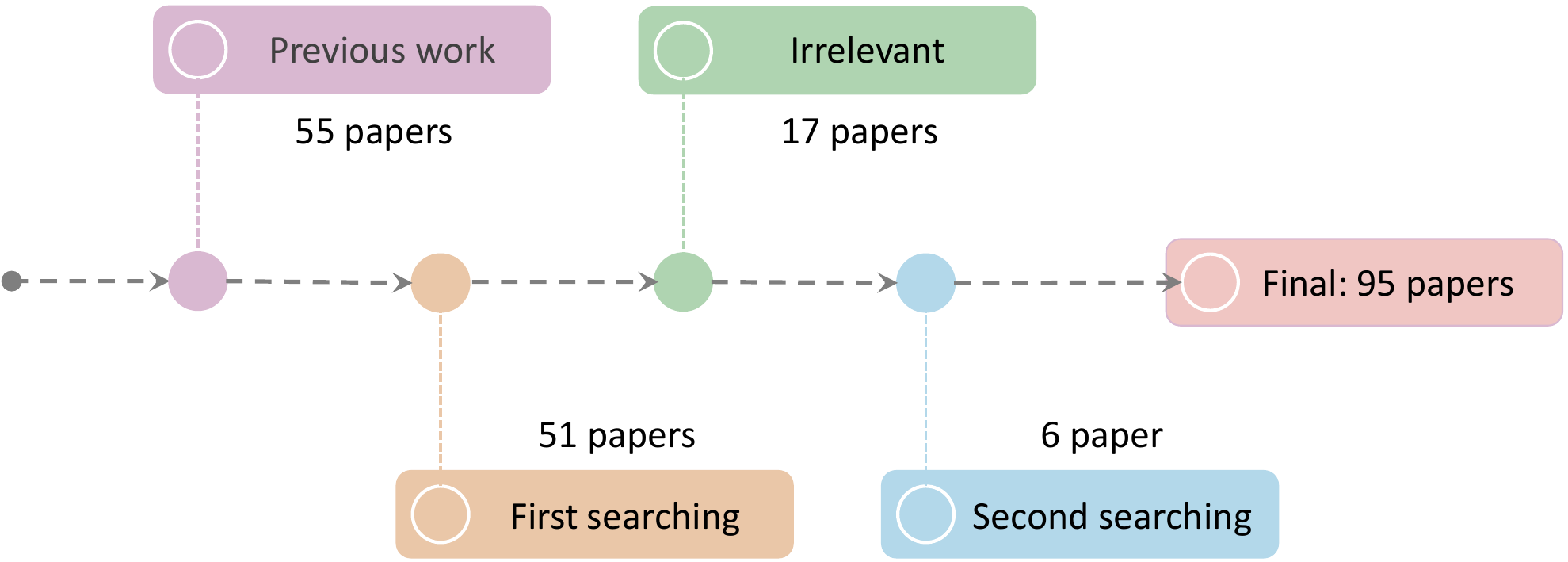}
\caption{The collection process of related papers.}
\label{fig62}
\end{figure}

To illustrate the collection process of related papers, Fig.~\ref{fig62} provides the details. Based on our previous review, there are 55 related papers from 1992 to 2017. We collect 51 papers from several databases, including Google Scholar, IEEE, Elsevier, Springer, ACM, and other academic databases. The keywords used for searching contains ``microorganism image analysis'', ``classification'', ``detection'', ``segmentation'', ``counting'', ``artificial neural network'', ``convolutional neural network'', and ``deep learning''. Their combinations are used to search related papers in these databases. Besides, a specific naming rule based on the publication year and article title is designed to avoid duplicate paper downloads, e.g., ``2010-Bacteria classification using neural network''. After carefully reading, 17 papers on other topics are excluded, and six related paper are added. Finally, a total of 95 research papers are retained for our review. 

\begin{figure}[htbp!]
\centering
\includegraphics[width=0.85\linewidth]{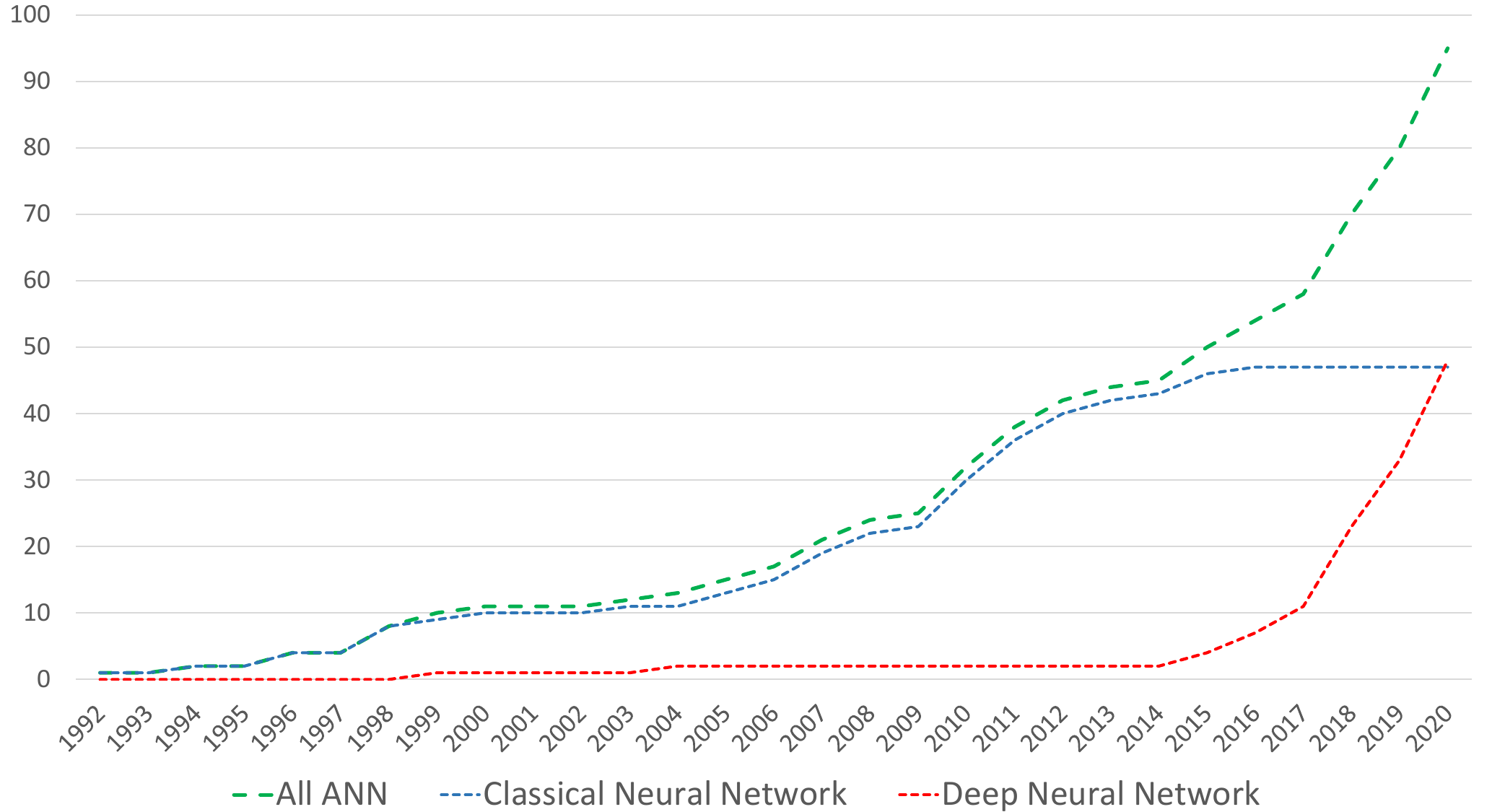}
\caption{The development trend of ANN methods for MIA tasks. The horizontal direction shows the time. The vertical direction shows the cumulative number of related works in each year.}
\label{fig3}
\end{figure}

The popularity and trend of ANN for the analysis of microorganism images are provided in Fig.~\ref{fig3}. Prior to the popularity of deep ANNs, most computer vision research works focus on developing feature engineering, which is usually based on specific domain knowledge~\cite{Liu-2020-DLGO}. After the proposal of AlexNet, deep learning methods (especially deep CNNs), which can learn powerful feature representations with multiple levels of abstraction directly from raw images, start to lead the research trend of computer vision~\cite{Liu-2020-DLGO}. Many pieces of research are devoted to studying new network structures, loss functions, attention mechanisms, and other hotspot issues. The development trend of ANNs also can be reflected in MIA tasks. Before 2014, most works adopted classical ANNs based on feature engineering. After, most researches employ deep ANNs to perform MIA tasks, especially deep CNNs.

\subsection{Structure of This Review}
This review is structured as follows: we begin by introducing the development of ANN and some representative networks used in MIA tasks in Sec.~\ref{ANN}; Sec.~\ref{CNN} introduces the MIA work using classical ANN methods; in Sec.~\ref{DNN}, deep ANN methods employed in the MIA tasks are summarized; Sec.~\ref{MA} presents the method analysis and potential directions; Sec.~\ref{CFW} concludes this paper.

\section{Artificial Neural Networks}
\label{ANN}

In this section, to better understand the development of ANN in the MIA task, we briefly introduce the evolution history of ANN. Besides, some representative ANN structures in MIA tasks are introduced.

\subsection{Evolution of Artificial Neural Networks}
\begin{figure}[htbp!]
\centering
\includegraphics[width=1\linewidth]{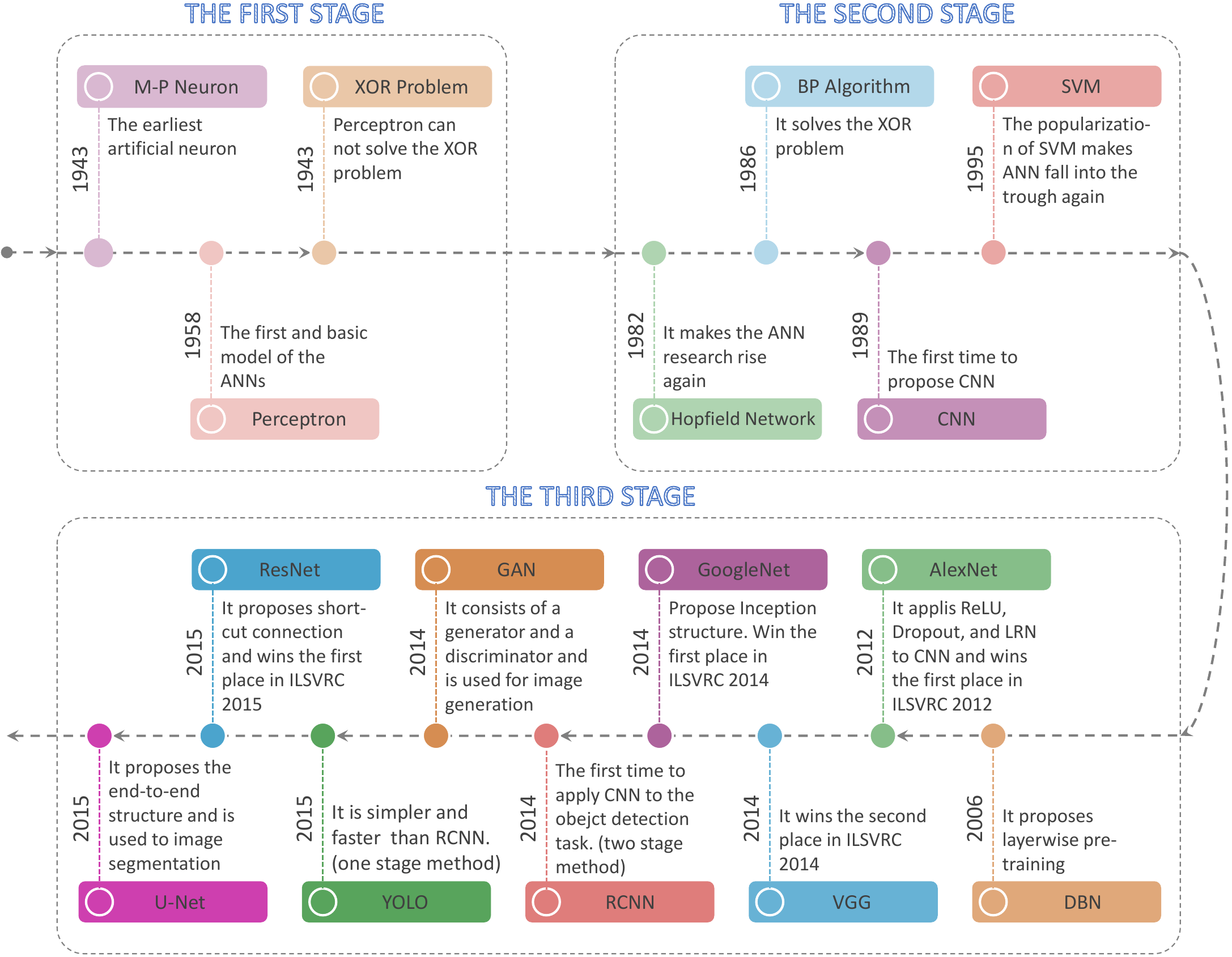}
\caption{The evolution of ANN.}
\label{fig4}
\end{figure}

The development of ANN has a long history. As shown in Fig.~\ref{fig4}, its course can be divided into three stages~\cite{Yamashita-2016-AIGD}. The ANN research can be traced back to M-P Neuron, designed according to the biological neuron, proposed in~\cite{Mcculloch-1943-ALCI} in 1943. This model is a physical model made up of elements such as resistors. Perceptron, whose learning rule is based on the original M-P Neuron, is proposed in~\cite{Rosenblatt-1958-PAPM}. After training, perceptron can determine the connection weights of neurons~\cite{Yamashita-2016-AIGD}. Since then, the first upsurge of ANN has started. However, Minsky et al. point out that perceptron can not be applied to solve the XOR problem (the linearly inseparable problem) in~\cite{Minsky-2017-PAIC}, which makes the ANN research fall into the trough.

In the second stage, with the proposal of the Hopfield network~\cite{Hopfield-1982-NNPS} in 1982, the study of ANN attracts attention again. After that, with the proposal of the \emph{Back-Propagation} (BP) algorithm~\cite{Rumelhart-1986-LRBE}, the XOR problem can be solved by training MLP using the BP algorithm. Besides, LeCun et al. propose CNN by introducing the convolutional layer, inspired by the biological primary visual cortex, into the neural network~\cite{Lecun-1989-BAHZ,Lecun-1998-GLAD}. However, limited by the computer performance and neural network theory at that time, although BP enables CNN to be trained, there are still problems such as too long training time and easy over-fitting. With the popularization of SVM, the ANN research falls into the trough again.

Although the ANN research falls into the trough, Hinton and Bengio et al. still focus on the research of ANNs~\cite{Hinton-2002-TPEM,Hinton-2006-AFLA,Hinton-2006-RDDN,Nair-2010-RLUI,Salakhutdinov-2009-DBM,Bengio-2009-LDAA,Bengio-2007-GLTD,Glorot-2010-UDTD,Le-2008-RPRB}. Benefit from their research progresses, ANNs show overwhelming advantages in speech and image recognition~\cite{Krizhevsky-2017-ICDC}. Since then, the third rise of ANN has begun. Different from the second stage, the computer performance is significantly improved. Training a deep neural network does not need such a long time as before. Besides, with the popularization of the internet, more and more data can be used for training, reducing the over-fitting problem.

The ANN-based MIA research also follows the development trend of ANN. Some representative ANN structures in MIA tasks are introduced in the following part.

\subsection{Representative Artificial Neural Networks}
\label{RANN}

ANN plays an essential role in the MIA task. We investigate related research papers and find that early MIA tasks based on classical neural networks rely on the ``Feature Engineering + Classifier'' workflow. They usually extract features from images by the existing experience or domain knowledge. After extraction, these features are used for training classifiers to perform corresponding tasks. Among these classifiers, MLP is the most widely used. With the widespread use of CNNs, the MIA task can no longer rely on feature engineering. Among these CNNs, AlexNet~\cite{Krizhevsky-2017-ICDC}, VGGNet~\cite{Simonyan-2014-VDCN}, ResNet~\cite{He-2016-DRLI}, and Inception~\cite{Szegedy-2015-GDC} are popular in the microorganism image classification task, U-Net~\cite{Ronneberger-2015-UCNB} and its improved networks are also widely used for the microorganism segmentation task, and YOLO~\cite{Redmon-2016-YOLO} is also widely used in the microorganism detection task. To understand the characteristics of these networks, we provide brief descriptions of MLP, AlexNet, VGGNet, ResNet, Inception, U-Net, and YOLO below.

\subsubsection{Multilayer Perception}
\label{MLP}

\begin{figure}[htbp!]
\centering
\includegraphics[width=0.6\linewidth]{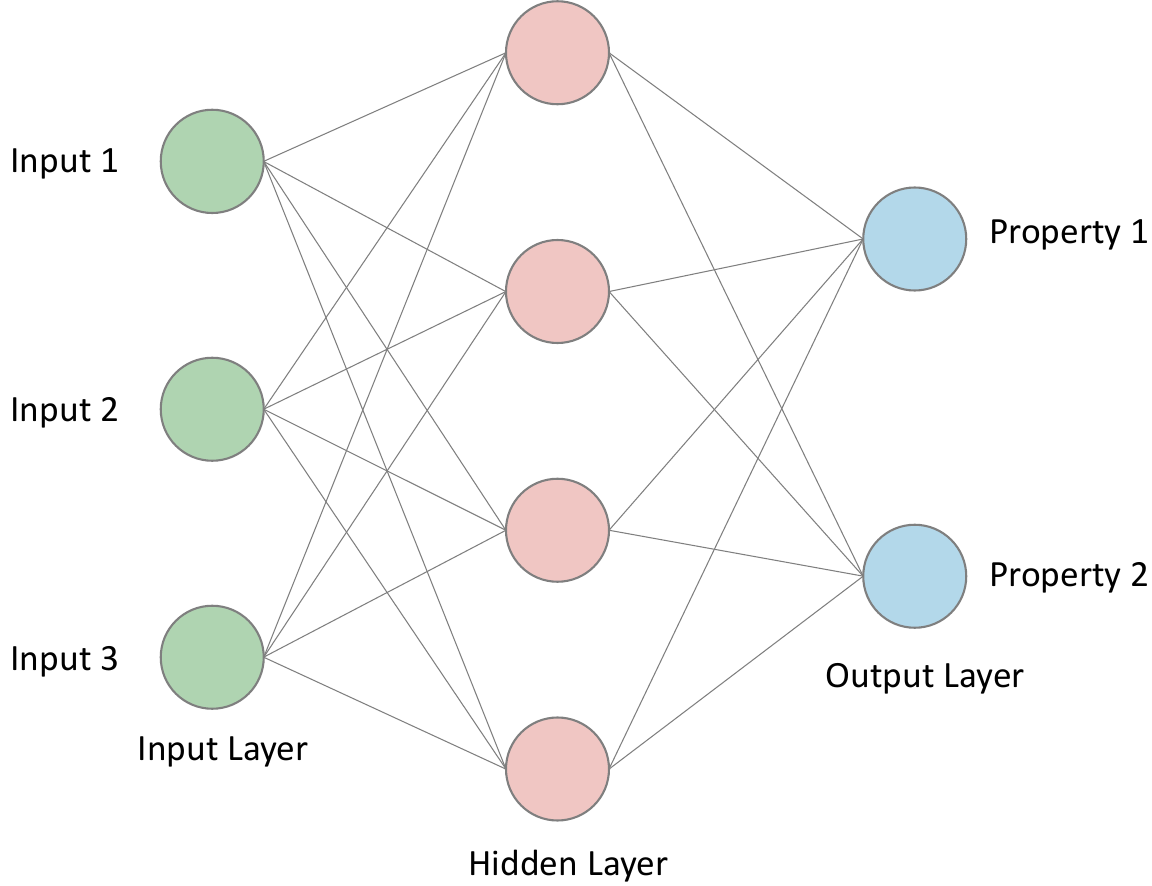}
\caption{An example of Multilayer Perceptron.}
\label{fig5}
\end{figure}

As is shown in Fig.~\ref{fig5}, an MLP usually consists of three layers: an input layer, a hidden layer, and an output layer. The early MLP is a class of feed-forward ANN. Like the perceptron, the early MLP can determine the connection weight between two layers by the error correction learning, which adjusts the connection weight according to the error between the expected output and the actual output. However, error correction learning cannot work between multiple layers. For this reason, the early MLP uses the random number to determine the connection weight between the input layer and the hidden layer, and the error correction learning is used to adjust the connection weight between the hidden layer and the output layer. With the proposal of the BP algorithm, MLP can adjust the connection weight layer by layer.

\subsubsection{AlexNet}
\label{AlexNet}
\begin{figure}[htbp!]
\centering  
\subfigure[AlexNet]{
\label{fig6_1}
\includegraphics[width=0.2\textwidth]{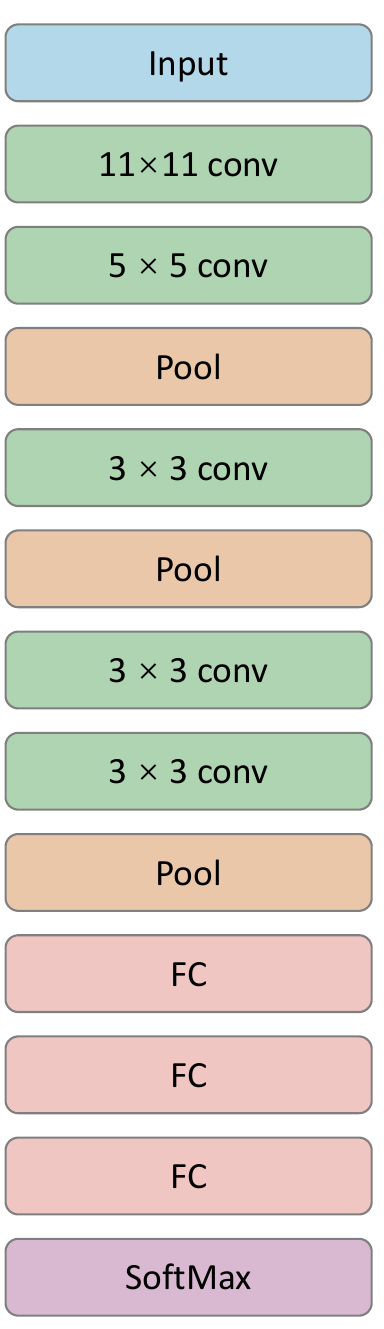}}
\subfigure[VGG-16]{
\label{fig6_2}
\includegraphics[width=0.2\textwidth]{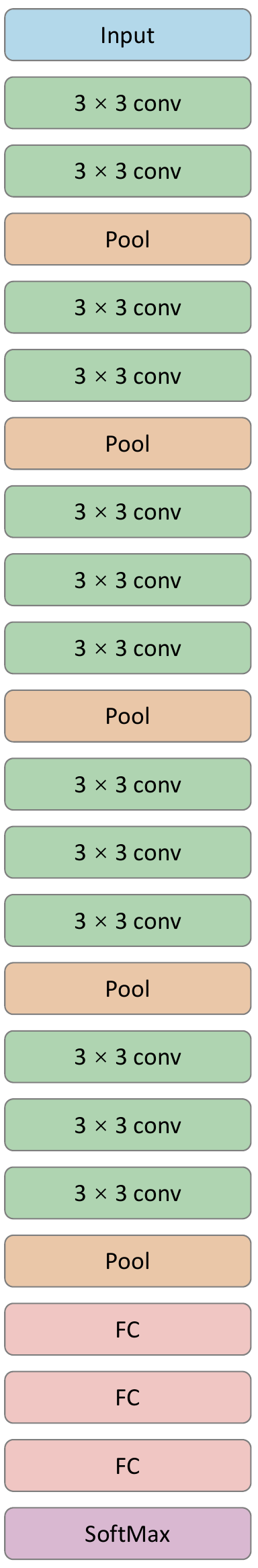}}
\subfigure[GoogLeNet]{
\label{fig6_3}
\includegraphics[width=0.238\textwidth]{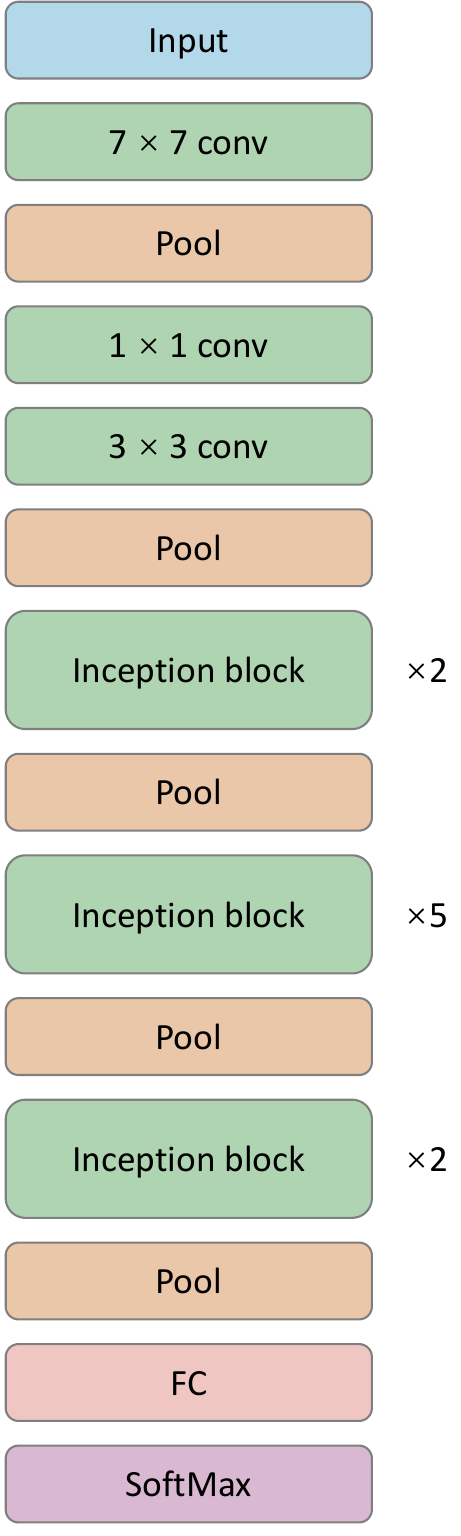}}
\subfigure[ResNet]{
\label{fig6_5}
\includegraphics[width=0.26\textwidth]{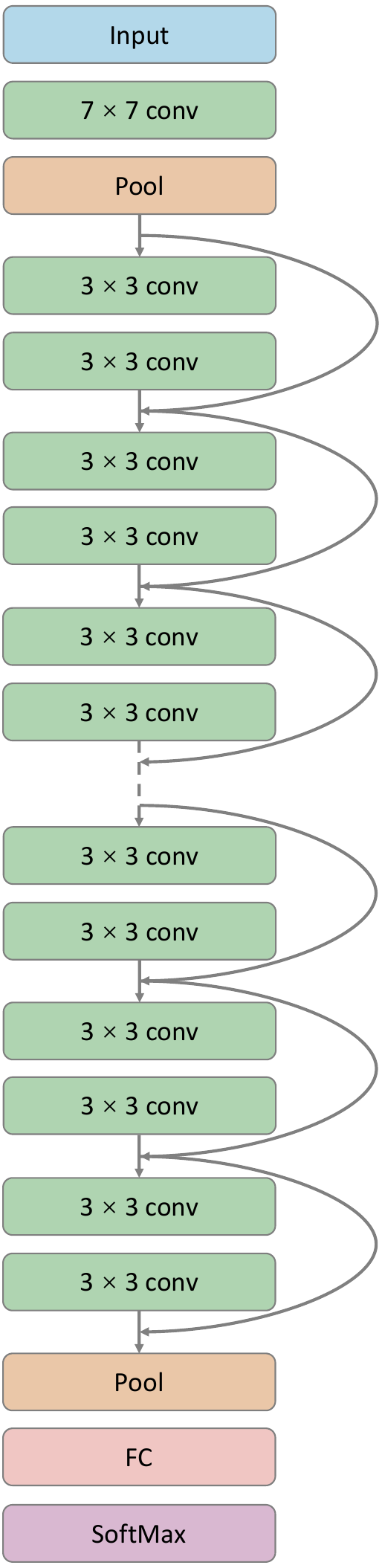}}
\subfigure[Inception block]{
\label{fig6_4}
\includegraphics[width=0.5\textwidth]{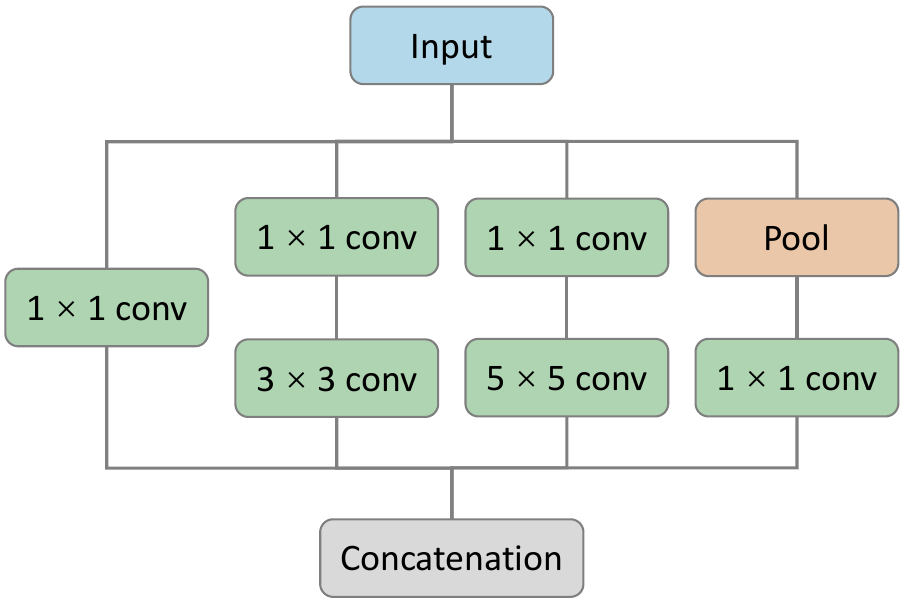}}
\subfigure[Residual learning block]{
\label{fig6_6}
\includegraphics[width=0.4\textwidth]{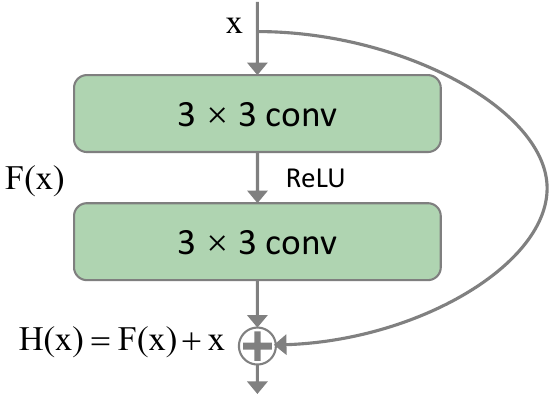}}
\caption{DNN architectures. (a) is AlexNet, (b) is VGG-16, (c) is GoogLeNet, (d) is ResNet, (e) is the Inception block used in GoogLeNet, and (f) is the residual learning block used in ResNet.}
\label{fig6}
\end{figure}

AlexNet adopts an architecture with consecutive convolutional layers. It is the first ANN to win the ILSVRC 2012~\cite{Rahaman-2020-ASCC}. After its triumphant performance, the champion architectures in the following years are all deep CNNs. As shown in Fig.~\ref{fig6_1}, AlexNet consists of eight layers, including five convolutional layers and three fully connected layers. The significance of AlexNet is that they use the \emph{Rectified Linear Unit} (ReLU) as an activation function instead of the sigmoid or hyperbolic tangent function~\cite{Rahaman-2020-ASCC}. AlexNet is trained by multi-GPU, which can cut down on the training time. Besides, different from the conventional pooling methods, overlapping pooling is introduced to AlexNet. AlexNet has 60 million parameters~\cite{Russakovsky-2015-ILSV}. To avoid the overfitting problem, data augmentation and dropout are employed.

\subsubsection{VGGNet}
\label{VGGNet}
\cite{Simonyan-2014-VDCN} proposes a CNN model named VGGNet, which wins second place in ILSVRC 2014. VGGNet is characterized by its simplicity, using only the $3\times3$ convolutional filter, which is the smallest size to capture the spatial information of left/right, up/down, center. The layer number of VGGNet could be 16 or 19. As the architecture of VGG-16 is shown in Fig.~\ref{fig6_2}, the network consists of 13 convolutional layers, five max pooling layers, three fully connected layers, and a softmax layer. VGG-19 has three more convolution layers than VGG-16. VGG-16 and VGG-19 comprise 138 and 144 million parameters~\cite{Simonyan-2014-VDCN}. The significance of VGGNet is the use of $3\times3$ convolutional filters, whose stack could obtain the same receptive filed as the bigger convolutional filter used in AlexNet. Besides, this stack, which has fewer parameters than the bigger convolutional filter used in AlexNet, allows VGGNet to have more weight layers. It can make better performance~\cite{Rahaman-2020-ASCC}.

\subsubsection{Inception}
\label{Inception}
\cite{Szegedy-2015-GDC} proposes GoogLeNet, which wins first place in ILSVRC 2014. Fig.~\ref{fig6_3} shows that it consists of 22 convolutional layers and five pooling layers. Nevertheless, this network has only 7 million parameters~\cite{Rahaman-2020-ASCC}. The significance of GoogLeNet is that it first introduces the Inception structure. As the Inception structure is shown in Fig.~\ref{fig6_4}, it first uses the idea of using the $1\times1$ convolutional filter to reduce the channel number of the previous feature map to reduce the total number of parameters of the network. Besides, the structure uses convolutional filters of different sizes to obtain multi-level features to improve the performance~\cite{Zhang-2020-AMCF}.

In the Inception-V2, batch normalization is added, and a stack of two $3\times3$ convolutional filters is employed instead of a $5\times5$ convolutional filter to increase the network depth and reduce the parameters~\cite{Ioffe-2015-BNAD}. In Inception-V3, a stack of $1\times n$ and $n \times1$ convolutional filters is used to replace the $n \times n$ convolutional filter~\cite{Szegedy-2016-RIAC}. In Inception-V4, the idea of residual learning block is incorporated~\cite{Szegedy-2017-IIIR}.

\subsubsection{ResNet}
From AlexNet (five convolutional layers), VGGNet (16 or 19 convolutional layers), to GoogLeNet (22 convolutional layers), the structure of CNNs is getting deeper and deeper. Deep CNNs allow more complex feature extraction, which leads to better results in theory. However, extending the depth of CNN by simply adding layers may lead to gradient vanishing or explosion problems~\cite{Bengio-1994-LLDG,Glorot-2010-UDTD}. To solve these problems, \cite{He-2016-DRLI} proposes ResNet, whose structure is shown in Fig.~\ref{fig6_5}. ResNet employs a novel approach, called identity mapping, to increase network depth for improving performance. Fig.~\ref{fig6_6} shows the residual learning block-based on identity mapping. In this block, $H(x)$ denotes an underlying mapping to be fit by the stacked layers, and the input feature map of these layers is denoted as $x$. The residual function can be denoted as $F(x) = H(x) - x$. The residual network's primary purpose is to make a deeper network from a shallow network by copying weight layers in the shallow network and setting other layers in the deeper network to be identity mapping.

\subsubsection{U-Net}
\label{U-Net}
\begin{figure}[htbp!]
\centering
\includegraphics[width=0.95\linewidth]{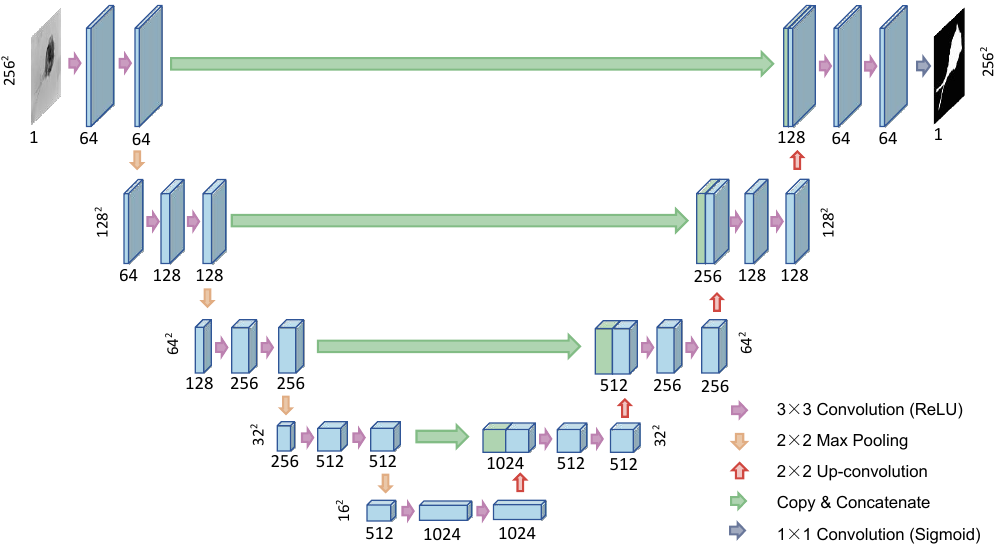}
\caption{U-Net architecture.}
\label{fig12}
\end{figure}

U-Net is a CNN initially used to perform the microscopic image segmentation task~\cite{Zhang-2021-LANL}. As the structure is shown in Fig.~\ref{fig12}, U-Net is symmetrical, consisting of a contracting path (left side) and an expansive path (right side), which gives it the u-shaped architecture. The training strategy of U-Net relies on the strong use of data augmentation to make more effective use of the available annotated samples~\cite{Ronneberger-2015-UCNB}. Besides, the end-to-end structure of U-Net can retrieve the shallow information of the network~\cite{Ronneberger-2015-UCNB}.

\subsubsection{YOLO}
\begin{figure}[htbp!]
\centering
\includegraphics[width=0.95\linewidth]{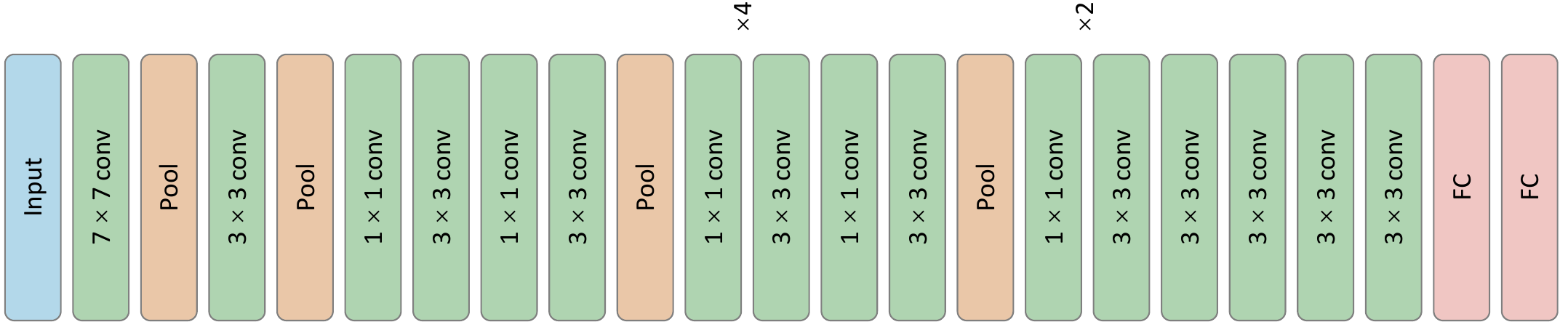}
\caption{YOLO architecture.}
\label{fig13}
\end{figure}

\cite{Redmon-2016-YOLO} proposes a novel framework called YOLO, which uses the whole topmost feature map to predict confidences for multiple categories and bounding boxes~\cite{Zhao-2019-ODDL}. YOLO's main idea is to divide the input image into an $ S \times S$ grid, and the grid cell is responsible for detecting the object centered in that cell~\cite{Redmon-2016-YOLO}. Each grid cell predicts $B$ bounding boxes and confidence scores for those boxes~\cite{Redmon-2016-YOLO}. These confidence scores reflect how confident the model is that the box contains an object and how accurate it thinks the box is that it predicts~\cite{Redmon-2016-YOLO}. Each grid cell also predicts $C$ conditional class probabilities. It should be noticed that only the contribution from the grid cell containing an object is calculated~\cite{Zhao-2019-ODDL}. The structure of YOLO is shown in Fig.~\ref{fig13}. It consists of 24 convolutional layers and two fully connected layers. Instead of the Inception used by GoogLeNet, they use $1\times 1$ reduction layers followed by $3\times3$ convolutional layers.

\section{Classical Neural Network for Microorganism Image Analysis}
\label{CNN}
An overview of the MIA task using classical neural networks is discussed in this section. We divide the tasks into different categories according to their tasks. Due to the classification works occupying most of all MIA tasks using classical neural networks, the related works are analyzed according to their used networks. For other tasks, the work is analyzed independently because each task contained a limited number of papers. Besides, we provide a summary to summarize the characters of the MIA based on classical neural networks and a table for readers to find relevant research works conveniently.

\subsection{Classification Task}

\subsubsection{Classification based on MLP}
\label{CMLP}
Through the investigation of MIA based on classical neural networks, it can be found that MLP is the most widely used network. Around 30 papers adopt MLP-based methods to perform the microorganism image classification task, but they involve various microorganism categories, with differences in image pre-processing, feature extraction, and MLP structure. Therefore, we critically and comparably analyze these papers according to the details of their methods.

Since the direct use of MLP to process image data requires a large number of neurons to connect each pixel in the image, most of the related work adopts the MLP based on feature engineering. Before feature extraction, image data usually need to be pre-processed. Therefore, to comprehensively understand the characteristic of MLP-based microorganism image classification methods, the analysis of the image pre-processing, feature extraction, and MLP implementation detail are provided, respectively.

\paragraph{\textbf{Image Pre-processing:}}
Most MLP-based microorganism image classification methods employ image pre-processing before extracting features. In most of these related works, the purpose of image pre-processing is to segment the \emph{Region of Interest} (ROI). The segmentation methods can be summarized into thresholding-based, edge detection-based, region growing-based, active contour-based, and manual methods.

\cite{Ginoris-2007-RPMU,Amaral-2008-SPII,Xiaojuan-2007-ANBC,Xiaojuan-2007-ANBR,Xiaojuan-2008-ANWB,Cunshe-2008-ANWB,Xiaojuan-2009-AIBN,Danping-2013-IPMS,Kruk-2015-CCSI} employ thresholding-based segmentation methods to split the microorganism objects from the images. \cite{Ginoris-2007-RPMU} proposes a framework to recognize protozoa and metazoa. The rough ROIs are selected manually after the image contrast enhancement and denoising first. Then, the accurate ROIs are generated by thresholding-based approach, whose threshold value can be defined by manual method, Otsu, or Entropy method. As an extension work, \cite{Amaral-2008-SPII} uses a similar segmentation approach to identify stalked protozoa. \cite{Xiaojuan-2007-ANBC,Xiaojuan-2007-ANBR} propose a framework to perform the bacteria classification task. An approach named ITSMM is presented to detect the edges of bacteria, which combines the iterative threshold value segmentation with mathematical morphology edge detection. As the extension works, \cite{Xiaojuan-2008-ANWB,Cunshe-2008-ANWB,Xiaojuan-2009-AIBN} use similar segmentation approaches for different microorganism edge detection tasks. \cite{Danping-2013-IPMS} focus on the identification task of powdery mildew spores. The image segmentation process contains Otsu-based binarization, image smoothing, connected domain analysis. Before segmentation, sequential pre-processing operations are applied first, including illumination compensation, greying, and image enhancement. Analogously, \cite{Kruk-2015-CCSI} uses Otsu-based segmentation method before the feature extraction in the soil microorganism recognition task.

\cite{Culverhouse-1996-ACFD,Blackburn-1998-RDBA,Mosleh-2012-APSA} adopt edge-based segmentation methods. \cite{Culverhouse-1996-ACFD} compare different \emph{Dinoflagellates} classification methods, including human experts, two ANN methods (MLP and RBF), and two ML methods (KNN and QDA). Sobel edge detection is applied in ANN and ML classification methods to segment the \emph{Dinoflagellate} specimen from the background, debris, and clutter. \cite{Blackburn-1998-RDBA} utilizes the Marr-Hildreth operator to perform edge detection-based segmentation for classifying individual bacteria and non-bacteria. \cite{Mosleh-2012-APSA} introduces a preliminary study on automated freshwater algae recognition and classification system, in which image segmentation is applied to isolate the individual objects by using Canny edge detection algorithm after image contrast enhancement, image greying, and image binarization.

\cite{Culverhouse-2000-DAMV,Schulze-2013-PAAS,Osman-2011-TBDZ,Osman-2012-OSEL} use region-based methods to perform the segmentation tasks. \cite{Culverhouse-2000-DAMV} develops a recognition system to automatically categorize marine \emph{Dinoflagellate}, which uses local maximum detection and region growing algorithm to perform pre-processing operations after the wavelet transform of the input image. \cite{Schulze-2013-PAAS} develops an automated analysis system for the identification of phytoplankton. It uses a region growing approach to separate the organisms from the background of the microscopic image. The edge detection using the Sobel operator and contrast adjustment is performed first to generate better segmentation results. Then, the watershed segmentation is used to obtain the rough segmentation results. Finally, region growing is applied for accurate segmentation. \cite{Osman-2011-TBDZ,Osman-2012-OSEL} develop methods based on the image analysis technique and ANN to detect the \emph{Mycobacterium tuberculosis} in the tissue section. CY-based color filter and $k$-mean clustering are first applied to remove pixels unrelated to red color. Then, the median filter and region growing are used to obtain the segmentation results.

In addition to using thresholding-based, edge detection-based, and region growing-based for segmentation, some works use active contour-based methods and manual methods. For the active contour-based method, \cite{Priya-2016-AOIL} proposes a digital tuberculosis object-level and image-level classification method based on MLP activated by the SVM learning algorithm. In the pre-processing operations, the non-uniform illumination correction technique is applied to reduce the influence caused by non-uniform illumination. Then, the active contour segmentation method is used to identify the presence of bacilli and outliers. For manual methods, \cite{Embleton-2003-ACPP} develops a phytoplankton identification and counting system, which uses manually selected threshold values to segment the ROIs. Additionally, the pre-processing applied in~\cite{Weller-2005-SCSO,Weller-2007-TSNN} also contains image segmentation, but the detailed methods are not mentioned.

Although most papers utilize image pre-processing for object segmentation, some papers use pre-processing without segmentation before feature extraction. For instance, \cite{Culverhouse-1994-ACFS} proposes an automatic taxonomic classification method, in which the feature extraction is applied after the \emph{Cymatocylis} image pre-processing, which only includes digitization, binarization, and manual noise reduction. Besides, \cite{Veropoulos-1998-IPNC} introduces an automatic method to detect TB in sputum smears. The pre-processing operations, including edge detection, region labeling and removal, edge pixel linking, and boundary tracing, are applied one by one for subsequent shape feature extraction. Additionally, \cite{Widmer-2005-UANN} proposes a MLP-based method for the identification task of these two protozoa. The image data is directly clipped to fit the input of the networks.

Through the comparative analysis of the pre-processing of classification tasks based on MLP above, it can be found that although the methods implemented are variable, most of the pre-processing methods aim to achieve the segmentation of microorganism objects. Most of the remaining pre-processing methods focus on reducing noise in the image data and highlighting the objects, which can facilitate effective feature extraction. However, there are some MLP-based works~\cite{Balfoort-1992-AIAN,Rulaningtyas-2011-ACTB,Kiranyaz-2011-CRMI} performing the microorganism image classification task without image pre-processing. In addition, some works~\cite{Gerlach-1998-IRSM,Kay-1999-TAAS,Ginoris-2006-RPMU,Zeder-2010-AQAA} do not indicate whether image pre-processing operations are used.

\paragraph{\textbf{Feature Extraction:}}
Most MLP-based microorganism image classification methods employ feature extraction in their workflows, which is a crucial step for classification performance. Therefore, to summarize their characteristics, we analyze them from different feature categories, including shape, texture, and color.

The shape is an important visual feature, which is widely utilized to describe microorganism image content in MLP-based classification methods. The shape features used in MLP-based methods mainly include geometrical descriptors, spectral transform, invariant moments, and junction descriptors. Geometrical descriptors are common and simple shape features~\cite{Zhang-2004-RSRD}, which usually contain area, perimeter, eccentricity, circularity, and other descriptors. Spectral descriptors can overcome the problem of noise sensitivity and boundary variations by analyzing shape in the spectral domain~\cite{Zhang-2004-RSRD}. Fourier descriptor is one of the most widely used spectral descriptors. The invariant moment is also widely used to describe the shape information. Besides, some works extract junction descriptors as the shape features.

Geometrical descriptors are widely used in MLP-based microorganism classification tasks, but the implementation details are various. Some MLP-based classification works only or mainly adopt geometric descriptors. For instance, \cite{Gerlach-1998-IRSM} adopts simple and common geometrical descriptors (area, eccentricity, circularity, and length of the skeleton) as the microorganism features. \cite{Rulaningtyas-2011-ACTB} extract geometrical features for TB classification, including perimeter, area, radii, circularity, compactness, eccentricity, and tortuosity. Besides, \cite{Osman-2012-OSEL,Siena-2012-DATB} also employ geometrical features for TB classification. Additionally, different from simple measurement of objects, \cite{Kay-1999-TAAS} measures the geometrical details of microorganism structures, including hooks, dorsal bar, and other structures. Nevertheless, some works adopt multiple types of features, which contain not only geometrical features but also other features. For example, \cite{Embleton-2003-ACPP} extracts not only geometrical descriptors but also color and other shape features, like color distribution and Fourier descriptor, for the sequential analysis. \cite{Weller-2005-SCSO,Weller-2007-TSNN} also extract multiple types of features as the original features, which include the axis, compactness, circularity, and other geometrical descriptors. After extracting the original features, a classification tree algorithm is applied to select the effective features from original features to train the neural networks. \cite{Xiaojuan-2007-ANBC,Xiaojuan-2007-ANBR} extract geometrical features, contour invariant moments, and texture features in their works. \emph{Principal Component Analysis} (PCA) is used for dimensionality reduction of features. \cite{Ginoris-2006-RPMU,Ginoris-2007-RPMU} adopt not only geometrical descriptors (area, perimeter, length, width, etc.) but also other shape features to perform the semi-automatic recognition of several protozoa and metazoa commonly used in sewage treatment. \cite{Balfoort-1992-AIAN,Xiaojuan-2008-ANWB,Cunshe-2008-ANWB,Amaral-2008-SPII,Xiaojuan-2009-AIBN,Kiranyaz-2011-CRMI,Mosleh-2012-APSA,Danping-2013-IPMS} also use geometric features mixed with other features in their works.

Spectral descriptors contain Fourier descriptors and wavelet descriptors, which are derived from spectral transforms on 1-D shape signatures~\cite{Zhang-2004-RSRD}. Fourier descriptor is mainly used in MLP-based microorganism classification works. \cite{Culverhouse-1994-ACFS} adopts the 2-D Fast Fourier transform for shape feature extraction in \emph{Cymatocylis} classification tasks. \cite{Veropoulos-1998-IPNC} uses the discrete Fourier transform to calculate the Fourier coefficients needed to represent the shape of each object for TB classification. \cite{Blackburn-1998-RDBA} employs LabVIEW’s built-in signal-processing library to perform Fourier transform to obtain shape features for bacteria classification. \cite{Priya-2016-AOIL} also utilizes 15 Fourier descriptors to describe the shape information of TB objects. Nevertheless, different from previous works, fuzzy entropy is applied to select prominent Fourier descriptors. Additionally, \cite{Culverhouse-1996-ACFD,Culverhouse-2000-DAMV,Weller-2005-SCSO,Schulze-2013-PAAS,Ginoris-2006-RPMU,Ginoris-2007-RPMU} extract multiple types of features, which also contain Fourier descriptors to represent shape features. In addition to geometrical and spectral features, some works employ invariant moments and junction descriptors as shape features. \cite{Xiaojuan-2007-ANBC,Xiaojuan-2007-ANBR} extract Hu invariant moments for bacteria classification, which are independent of changes of shift, measurement, and rotation. \cite{Danping-2013-IPMS} also uses Hu invariant moments for the identification of powdery mildew spores. \cite{Osman-2011-TBDZ} extract six affine moment invariants to train ANN to perform the TB classification task. Besides, \cite{Culverhouse-1996-ACFD,Culverhouse-2000-DAMV} adopt junction descriptor as one of their features.

The texture is a vital element of human visual perception, and texture feature has widespread applications in many computer vision systems~\cite{Humeau-2019-TFEM}. It also is widely used in MLP-based classification tasks. In addition to using shape features, \cite{Culverhouse-1996-ACFD} also extract texture features for \emph{Dinoflagellates} classification. Similarly, \cite{Culverhouse-2000-DAMV} adopts the same texture features. \cite{Weller-2005-SCSO,Weller-2007-TSNN} extracts multiple types of features, which contain texture features. \cite{Xiaojuan-2007-ANBC,Xiaojuan-2007-ANBR} extract not only the shape but also texture features. Analogously, \cite{Mosleh-2012-APSA} extract both shape and texture features for algae identification.

In addition to shape and texture features, the color feature also plays an essential role in MLP-based microorganism classification tasks. For example, \cite{Balfoort-1992-AIAN} uses an optical plankton analyzer to measure color features to distinguish \emph{Cyanobacteria} from other algae. Besides extracting shape and texture features, \cite{Weller-2005-SCSO,Weller-2007-TSNN} also use color features in their works. \cite{Kruk-2015-CCSI} adopts color features to assist the identification of soil microorganisms.

Through the analysis of the feature extraction of MLP-based microorganism classification tasks, it can be found that shape features are the most frequently used feature type. Texture and color features are also used in some works. Most extracted features are directly fed to the subsequent classifiers. This strategy cannot explore the effective features and is inefficient for the classification of microorganisms. However, there are some exceptions. Some works adopt the tree algorithm to select effective features from original features. It is of great significance to explore efficient features for microorganism image classification.

\paragraph{\textbf{MLP Implementation Details:}}
MLP is the most widely-used network among classical neural networks in MIA tasks. The basic information and characteristic of MLP are provided in Sec.~\ref{MLP}. Through investigation of MLP-based microorganism classification tasks, it can be found that most of them adopt the three-layer structure based on the BP algorithm, which contains an input layer, a hidden layer, and an output layer. Nevertheless, there are some exceptions. In this part, the MLP-based works are introduced by their implementation details.

Among microorganism classification tasks, most three-layer MLPs use the BP algorithm in their training process. For instance, \cite{Culverhouse-1994-ACFS} adopts a three-layer MLP based on the BP algorithm to perform the \emph{Cymatocylis} classification task. The network consists of 15 input, three hidden, and five output neurons. \cite{Blackburn-1998-RDBA} employs the BP-based three-layer MLP network, which has 15 input, five hidden, and three output neurons, to classify individual bacteria and non-bacteria objects. \cite{Xiaojuan-2007-ANBC,Xiaojuan-2007-ANBR} also adopt the three-layer MLP using the BP algorithm for training to perform bacterial image classification. \cite{Gerlach-1998-IRSM} adopts the three-layer MLP based on the BP algorithm to perform the classification task for investigating the reactors' influence on the fungal morphology. Besides, \cite{Mosleh-2012-APSA,Siena-2012-DATB,Danping-2013-IPMS} adopt BP-based 21-8-5, 2-15-2, and 7-3-1 (Input-Hidden-Output neurons) MLP structures for algae, TB, and powdery mildew spores classification tasks, respectively. Unlike directly using the BP algorithm for MLP training, \cite{Xiaojuan-2008-ANWB,Cunshe-2008-ANWB,Xiaojuan-2009-AIBN} adopt an adaptive accelerated BP algorithm to train MLP for bacteria image classification tasks.

Through the above discussion, it can be found that the MLP structures are various. This is because the number of input neurons usually depends on the dimension of the feature vector~\cite{Embleton-2003-ACPP}. However, for the work without feature engineering, the number of input neurons depends on how many pixels the input image contains. For instance, \cite{Widmer-2005-UANN} adopts MLPs to identify \emph{Cryptosporidium parvum} and \emph{Giardia lamblia}. Due to the input image sizes (40 by 40 and 95 by 95 pixels for \emph{Cryptosporidium parvum} and \emph{Giardia lamblia}, respectively), the MLPs have 1600 and 9025 input neurons, respectively. The number of output neurons often depends on the category number of classification~\cite{Weller-2005-SCSO}. Nevertheless, there are some exceptions, such as \cite{Embleton-2003-ACPP} adopting the BP-based MLP, whose output layer includes only one neuron, for phytoplankton classification. The output neuron value ranges from -1 to 1, which can indicate the classification result. Besides, \cite{Danping-2013-IPMS} utilizes a similar output layer. Nevertheless, for the hidden layer, most works do not provide the details for using the relevant number of neurons.

The configuration setting of MLP aims to achieve good performance. Therefore, some works evaluate the performance of different MLP configurations. \cite{Weller-2005-SCSO,Weller-2007-TSNN} evaluate the classification performance of sedimentary organic matter based on different four-layer MLP configurations. Similarly, in the classification and retrieval of macroinvertebrate images, \cite{Kiranyaz-2011-CRMI} evaluate the classification accuracy of different MLP configurations, which include not only three-layer but also four-layer structures. In addition to~\cite{Weller-2005-SCSO,Kiranyaz-2011-CRMI}, \cite{Schulze-2013-PAAS} develops an automatic image analysis system to recognize phytoplankton, which employs the four-layer MLP. Besides the four-layer and three-layer structures, there are some works that adopt the two-layers MLP without any hidden layer. \cite{Ginoris-2007-RPMU} proposes a semi-automatic image analysis program for protozoa and metazoa classification, which uses a two-layer feed-forward MLP based on the BP algorithm. \cite{Amaral-2008-SPII} introduces a semi-automatic method to perform the recognition task of stalked protozoa species in sewage. It also uses the two-layer MLP with 15 input and ten output neurons.

In addition to the above works, there are some works that compare the performance of MLP with other ML methods. For example, in~\cite{Culverhouse-1996-ACFD}, human experts' performance in identifying \emph{Dinoflagellates} is compared to that achieved by two ANN classifiers (MLP and RBF) and two other statistical techniques, KNN and \emph{Quadratic Discriminant Analysis} (QDA). \cite{Veropoulos-1998-IPNC} proposes an automatic method to detect TB in sputum smears, in which the classifiers from discriminant methods of statistics or neural networks are compared. The results show that the MLP based on the BP algorithm performs best. \cite{Kay-1999-TAAS} develops a classification system based on statistical methods, which can successfully discriminate closely pathogen species within the same host. The statistical methods contain linear discriminant analysis, nearest neighbors, MLP, and projection pursuit regression. \cite{Culverhouse-2000-DAMV} compares the four methods of MLP, KNN, RBF, and QDA in the classification of \emph{Dinoflagellates}. The experimental results show that human performance declines while the performance of BPN rises most obviously when the number of species increases. 
Similar comparisons works contain \cite{Ginoris-2006-RPMU,Kiranyaz-2011-CRMI,Kruk-2015-CCSI,Priya-2016-AOIL}.

The above discussion shows that the three-layer MLP based on the BP algorithm occupies the leading position in microorganism image classification tasks. Besides that, some works adopt two-layer or four-layer MLP, but there is no work employing the MLP more than four layers. This is because the deep structure has more training parameters and may lead to the over-fitting problem under insufficient training data. However, at that time, the performance of computer hardware is limited, and the microorganism image data is difficult to collect. It is worth noting that the number of MLP input and output neurons usually depends on the dimension of the feature vector and classification category number. However, the setting of hidden layers in most works does not go through detailed theoretical or experimental analysis. In addition, some works do not provide the MLP implementation details. \cite{Balfoort-1992-AIAN,Zeder-2010-AQAA} do not explain whether they adopt the BP algorithm in the training process. \cite{Rulaningtyas-2011-ACTB,Osman-2011-TBDZ,Osman-2012-OSEL} do not provide the detailed configuration of MLP.

\subsubsection{Classification based on RBF}

In addition to MLP, RBF is also one of the most popular networks in MIA tasks based on classical neural networks. More than ten papers employ RBF-based methods to perform the microorganism image classification task. These methods involve several microorganism categories, with differences in image pre-processing, feature extraction, and MLP structure. Therefore, like Sec.~\ref{CMLP}, we analyze the image pre-processing, feature extraction, and MLP implementation detail, respectively.

\paragraph{\textbf{Image Pre-processing:}}
Like the characteristic of the pre-processing in MLP-based microorganism image classification methods, most RBF-based methods apply pre-processing operations before feature extraction. Almost all pre-processing operations are aimed to obtain the ROI. Since there are only about ten related works and some belong to a series of works, the pre-processing details are directly introduced in sequence.

\cite{Hiremath-2010-AICB} proposes a method for the classification of different cell growth phases. In this method, the adaptive global thresholding is applied to segment the cell images after several morphological operations, erosion, reconstruction, and dilation. As the extension works, \cite{Hiremath-2011-ICCB,Hiremath-2011-DMIA} employ the same segmentation method. \cite{Hiremath-2011-ICCB} introduces an image analysis method for cocci bacterial cell classification. \cite{Hiremath-2011-DMIA} focuses on the classification of spiral bacterial cells, which contain three categories: \emph{vibrio}, \emph{spirillum}, and \emph{spirochete}. However, as a series of works, \cite{Hiremath-2010-DIAC} adopts the active contour method to obtain the segmented images in the analysis of cocci bacterial cells. \cite{Hiremath-2012-SBCI} also adopts the active contour method in the segmentation part of spiral bacterial cell analysis. Besides, \cite{Priya-2015-AITO} uses the active contour segmentation method using level set formulation and Mumford-Shah technique in the identification of TB in digital sputum smear images. There are some works that compare multiple ANN classifiers in the microorganism classification tasks. For the works involving both RBF and MLP, including~\cite{Culverhouse-1996-ACFD,Culverhouse-2000-DAMV,Kruk-2015-CCSI}, the pre-processing details are provided in Sec.~\ref{CMLP}. Besides, the pre-processing applied in~\cite{Weller-2007-TSNN} also contains image segmentation, but the detailed methods are not mentioned.

Same as MLP, although the pre-processing operations are variable, all aim to achieve the segmentation of microorganism objects and reduce the influence of noise. Additionally, \cite{Kiranyaz-2011-CRMI} performs the macroinvertebrate image classification task without image pre-processing.

\paragraph{\textbf{Feature Extraction:}}
Most RBF-based microorganism image classification methods extract image features after the pre-processing. The feature extraction and selection have a significant influence on the performance of RBF. Same as the analysis of pre-processing, we also analyze the feature extraction of related works in sequence.

\cite{Hiremath-2010-AICB} uses geometrical features to train RBF and other classifiers. The geometrical features contain circularity, compactness, eccentricity, tortuosity, and length-width ratio. As the extension works, \cite{Hiremath-2010-DIAC,Hiremath-2011-ICCB,Hiremath-2011-DMIA,Hiremath-2012-SBCI} also adopt geometrical features, but there are some differences. \cite{Hiremath-2010-DIAC,Hiremath-2011-ICCB,Hiremath-2011-DMIA} employ the same five geometrical features, but \cite{Hiremath-2012-SBCI} only uses three of them: circularity, eccentricity, and tortuosity. \cite{Priya-2015-AITO} also uses the geometrical feature. Nevertheless, unlike other works, it selects the most significant features from the original features by student’s ‘$t$’ test. Finally, compactness, eccentricity, circularity, and tortuosity are chosen for training the following classifier. As the extension of~\cite{Weller-2005-SCSO}, \cite{Weller-2007-TSNN} extracts the same original features, which contains shape(geometrical descriptors, Fourier descriptors, and others), texture, and color features. After the initial extraction, a classification tree algorithm is applied to select the effective features. \cite{Kiranyaz-2011-CRMI} uses both geometrical statistical features. Moreover, due to \cite{Culverhouse-1996-ACFD,Culverhouse-2000-DAMV,Kruk-2015-CCSI} evaluating both MLP and RBF in their works, the feature extraction details are provided in Sec.~\ref{CMLP}.

After the summary of the feature extraction in RBF-based microorganism classification tasks, it can be found that most works directly extract the features from the pre-processed image data without more refined and efficient feature screening. However, \cite{Priya-2015-AITO,Weller-2007-TSNN} adopt some methods to screen significant features from the original features. Using more specific features can improve the efficiency of the microorganism classification system.

\paragraph{\textbf{RBF Implementation Details:}}
RBF is the second widely-used ANN in microorganism image classification tasks. It is a commonly used ANN for function approximation problems~\cite{Amir-2016-NMS}. The popular RBF form is a three-layer feed-forward neural network, which contains input, hidden, and output layers. The hidden layer comprises several radial basis function nonlinear activation units~\cite{Faris-2017-ERBF}. Activation functions in RBFs are conventionally implemented as Gaussian functions~\cite{Faris-2017-ERBF}. It distinguishes itself from other ANNs due to its universal approximation and faster learning speed~\cite{Amir-2016-NMS}. Same as \emph{Image Pre-processing} and \emph{Feature Extraction}, the implementation details are introduced in sequence.

\cite{Hiremath-2010-AICB} evaluates four kinds of classifiers in the classification of bacilli bacterial cell growth phases. RBF classifier is compared with 3$\sigma$, KNN, and fuzzy classifiers. The experimental results show that all classifiers but the fuzzy classifier perform worse than RBF. As the extension works, \cite{Hiremath-2010-DIAC,Hiremath-2011-ICCB} evaluate 3$\sigma$, KNN, and RBF classifiers in the classification of cocci bacterial cells. The results show that RBF outperforms others. \cite{Hiremath-2011-DMIA} evaluates RBF, 3$\sigma$, KNN, and fuzzy classifiers in analyzing spiral bacterial cell groups. RBF, 3$\sigma$, and fuzzy classifiers achieve 100\% classification accuracy for different bacterial cells. Different from \cite{Hiremath-2011-DMIA}, the adaptive neuro-fuzzy inference system is used to replace the fuzzy classifier in~\cite{Hiremath-2012-SBCI}. The results show that RBF and the neuro-fuzzy classifier obtain 100\% accuracy for different spiral bacterial cells classification. \cite{Priya-2015-AITO} adopts RBF to identify TB in digital sputum smear. RBF is compared with the \emph{Adaptive Neuro-fuzzy Inference System} (ANFIS) and \emph{Complex-valued Adaptive Neuro-fuzzy Inference System} (CANFIS). The results show that the CANFIS achieves better accuracy than RBF and ANFIS. \cite{Weller-2007-TSNN} uses RBF to replace some MLPs in~\cite{Weller-2005-SCSO} to classify sedimentary organic matter. The experimental results show that in the best case, the correct recognition rate is improved by 4\% to just over 91\%. \cite{Kiranyaz-2011-CRMI} focuses on the automatic classification and retrieval of macroinvertebrate images. Four classifiers are evaluated in this work, which contain MLP and RBF. The results show that MLP has the best performance. Besides, \cite{Culverhouse-1996-ACFD,Culverhouse-2000-DAMV,Kruk-2015-CCSI} also evaluate both MLP and RBF in their works, in which all the RBFs outperform MLPs.

Through the above discussion, we can find that the RBF performs better than the MLP in most works. However, in~\cite{Kiranyaz-2011-CRMI}, MLP achieves the best performance. It's worth noting that several configurations of MLP, including different layer structures, are evaluated to find out the best configuration. However, RBF only contains one hidden layer. That may be the reason that MLP outperforms RBF.

\subsubsection{Classification based on other classical ANNs}

In addition to MLP and RBF, some works adopt other classical ANNs in the classification of microorganism images. Nevertheless, due to the limited number of each network, we directly analyze each related paper and provide detailed discussions about the research motivation, contribution, data, workflow, and result.

Image recognition methods are behindhand with the wide application of optical imaging samplers in plankton ecology research. However, most methods need manual post-processing to correct their results. To optimize this situation, \cite{Hu-2006-AAQT} proposes a dual-classification method. In this method, four kinds of shape features, including moment invariants, morphological measurements, Fourier descriptors, and granulometry curves, are used to train a \emph{Learning Vector Quantization Neural Network} (LVQ-NN) in the first classification, and texture-based features (co-occurrence matrix) are used to train the SVM to achieve the second classification. The LVQ-NN has two layers, namely, a competitive layer and a linear output layer. The complexity of the network depends on the number of training images and the number of taxa in the classifier. Only when the two classifiers' results are consistent will the data to be classified be divided into one class, which effectively reduces the false positive rate. The image data used in the experiments contain seven plankton categories and about 20000 images. The results show that compared with previous methods, the dual-classification system after correction can effectively achieve abundance estimation and reduce the error by 50\% to 100\%.

The conventional manual method for detecting \emph{Mycobacterium tuberculosis} is an ineffective but necessary part of diagnosing tuberculosis disease. In~\cite{Osman-2010-DMTZ}, a method based on the image analysis technique and ANN is introduced to detect the \emph{Mycobacterium tuberculosis} in the tissue section. Fifteen tissue slides are collected in the Department of Pathology, USM Hospital, Kelantan, and each slide is captured to generate 30 to 50 images. A total of 607 objects, which contains 302 definite TB and 305 possible TB, are obtained. In the experiments, the moving $k$-means clustering is applied to segment the images, and geometrical features of Zernike moments are extracted. Then, the \emph{Hybrid Multilayered Perceptron} (HMLP) is used to test the performance of different feature combinations in the detection task. The experimental results show that the HMLP with the best feature combination can achieve an accuracy of 98.07\%, a sensitivity of 100\%, and a specificity of 96.19\%.

TB detection in tissue is more complex and challenging than detection in sputum. As the extension of~\cite{Osman-2011-TBDZ}, \cite{Osman-2011-HMPN} introduces a method based on the image processing technique and ANN to detect and classify TB in tissue. The 1603 objects consist of three categories: TB, overlapped TB, and non-TB, which are collected from 150 tissue slide images. The captured image is segmented first, and then the six affine moment invariants are extracted to train the network to perform the classification task. It uses the HMLP network trained by integrating both Modified Recursive Prediction Error algorithm and Extreme Learning Machine to replace single-layer feed-forward neural network trained by Extreme Learning Machine in~\cite{Osman-2011-TBDZ}. The results show that this network can achieve the highest testing accuracy of 77.33\% and average testing accuracy of 74.62\% for 35 hidden nodes.

The traditional microorganism detection and identification method is laborious and time-consuming. In~\cite{Kumar-2010-RDMU}, a rapid and cost-effective method based on PNN is applied to classify the microorganism image classification task. The features, including various geometrical, optical, and textural parameters, are extracted from the pre-processed images to train the network in this method. In the experiments, the dataset includes five categories: \emph{Bacillus thuringiensis}, \emph{Escherichia coli K12}, \emph{Lactobacillus brevis}, \emph{Listeria innocua}, and \emph{Staphylococcus epidermis}. The experimental results show that the PNN based on nine kinds of features ($45^{\circ}$ run length non-uniformity, width, shape factor, horizontal run length non-uniformity, mean grey level intensity, ten percentile values of the grey level histogram, 99 percentile values of the grey level histogram, sum entropy, and entropy) can classify the microorganisms with 100\% accuracy.

The classification of algae is one of the fundamental problems in water resources management. In~\cite{Coltelli-2014-WMAR}, an automatic real-time microalgae identification and enumeration method based on image analysis is introduced. This method integrates segmentation, shape features extraction, pigment signature determination, and ANN classification. The ANN used here is a \emph{Self-Organizing Map} (SOM), whose details can be found in~\cite{Rissino-2009-RSTC,Sap-2008-HSOM,Silva-2007-AHPS}. The neuron number of SOM used in this study is fixed as the estimate of category number, and the neurons are interconnected through neighborhood relations. The data shown in the experiment is private. The experimental results show that the accuracy of this method is 98.6\% on 23 categories of 53869 images.

\subsection{Segmentation Task}
%分割
Continuous monitoring of cells during the fermentation process can significantly increase the fermentation yield. In~\cite{Shabtai-1996-MMMC}, a prototype \emph{Self-Tunning Vision System} (STVS) is developed to monitor morphologic changes of the yeast-like fungus Aureobasidium pullulans during the fermentation of pullulan polysaccharide. The workflow of this system contains several steps: pre-processing, segmentation, intermediate processing, feature extraction, and classification. A SOM network is employed for image segmentation. The experimental results show that the system is reliable, accurate, and versatile.

The Zeihl-Neelsen tissue slide image segmentation is a crucial part of diagnosing tuberculosis disease by the computer-aided method. A method based on HMLP is introduced to perform the segmentation task in~\cite{Osman-2010-STBZ}. In the experiments, there are a number of 5608 data used for training and numbers of 2268, 1634, and 2800 used for samples A, B, and C for testing. In the segmentation process, the hue and saturation components of each pixel in the kernel of $3\times3$ pixels are used as the inputs of the HMLP network. The center pixel of the kernel is used as the output of the HMLP network. If the testing results show that the pixel belongs to the segmentation object, the center pixel will be assigned as '1'; otherwise, it will be '0'. The experimental results show that this method can obtain the accuracy of 98.72\%, 99.45\%, and 97.75\% for samples A, B, and C, respectively.

The conventional tuberculosis diagnosis method based on the microscope is widely used in developing countries. Nevertheless, it is time-consuming. Therefore, in~\cite{Costa-2015-AITM}, a novel method based on image processing techniques is proposed to assist the diagnosis. It contains three steps: image acquisition, segmentation, and post-processing. In the segmentation part, SVM and ANN (three-layer feed-forward neural network) are used. The input variables of classifiers are combinations of pixel color features selected from four color spaces. The best feature is selected by a scalar feature selection technique. The output of the segmentation part is to determine whether it is a bacilli. The function of the post-processing step is to remove the non-bacilli parts. The training data used in the segmentation part consists of 1200 bacilli pixels and 1200 background pixels from 120 images. The experimental results show that SVM achieves the best sensitivity of 96.80\% with an error rate of 3.38\%. The ANN classifier achieves the best sensitivity of 91.53\% and an error rate of 5.20\%.

\subsection{Counting Task}
%计数
In~\cite{Wit-1998-AAAN}, a feed-forward MLP, which contains 81 input neurons, ten hidden neurons, and one output neuron, is proposed for the image enhancement and subsequent enumeration of microorganisms adhering to the solid substrate. Unlike traditional classical ANNs, this network does not rely on feature engineering. The network's input directly employs grey values from $9\times9$ pixel sub-images taken from the microscope image ($512\times512$ pixel). These grey values are processed to yield an output value for every pixel in the original image, whereafter microorganisms are indicated with a marking circle. For high-contrast images, nearly all adhering organisms can be detected correctly. However, in the lower-quality images, for metal or silicone rubber substrata, image enhancement by the ANN yielded enumeration of the adhering bacteria with an accuracy of 93\%-98\%. After ANN enhancement, 98\% of the yeast cells adhering on silicone rubber substrata were enumerated correctly. It can be concluded that for low quality and complicated images, ANNs have outstanding performance.

\subsection{Feature Extraction Task}

The manual method for bacteria classification is complex and ineffective. In~\cite{Zhu-2010-BCUN}, a bacteria image classification method based on the features generated by \emph{Pulse Coupled Neural Network} (PCNN), which is a biologically inspired ANN based on the work~\cite{Eckhorn-1989-ANNF}, is introduced. In the workflow, the acquired images are pre-processed by cropping the image and saving the region of interest first. Then, the entropy sequence features are generated by PCNN. In the end, the classification is done by a classifier based on Euclid distance. The experimental results show that this method is feasible and efficient.

\subsection{Summary}

According to the above discussion, MIA based on the classical neural network has the following characteristics. Firstly, due to the wide range of microorganism applications, such as medical diagnosis, environmental management, food production, and other fields, most datasets used in related papers are privately collected data. The lack of public datasets makes it challenging to develop the general MIA algorithm. It also makes the comparative evaluation of different algorithms difficult. Secondly, classical neural networks usually cannot be applied directly to the image data, so the workflow of MIA based on the classical neural network usually contains three steps: pre-processing, feature extraction, and classification. Therefore, feature engineering has a significant influence on the algorithm's performance. Besides, classical neural networks used in MIA tasks usually have shallow structures. This is because deep networks have numerous connections between different layers, which can result in abundant parameters and hard training problems. Thirdly, the most tasks of MIA based on classical neural networks are classification. Among other tasks, there are examples of directly applying classical neural networks to process image data, which are different from CNN that appears later.

In addition, to help readers find and understand relevant works quickly, we briefly summarize these papers of MIA based on classical neural network in Tab.~\ref{table3}, in which years, application tasks, references, datasets, object species, category number, methods, and results are provided.

\begin{landscape}\scriptsize
\renewcommand\arraystretch{1.2}
\setlength{\tabcolsep}{1.1pt}
\centering
\begin{longtable}{cccccccccccccccccccccccc}

\caption{Summary of the MIA works based on classical neural network. (Classification (C), Segmentation (S), Enhancement (E), Detection (D), Feature Extraction(FE))}

\endfirsthead
\caption[l]{Continue: Summary of the MIA works based on classical neural network.}
\endhead

\hline
Year & Task & Ref.                   & Data           & Object                                         & Class            & Method                                                         & Result                                                                                                                                                                            \\ \hline
1992 & C    & \cite{Balfoort-1992-AIAN}     & Private          & Algae                                          & 8                & MLP                                                            & Average identification rate \textgreater 93\%                                                                                                                                     \\
1994 & C    & \cite{Culverhouse-1994-ACFS}  & Private          & \textit{Cymatocylis}                           & 5                & MLP                                                            & Identification rate of  28\% trials\textgreater 70\%                                                                                                                              \\
1996 & C    & \cite{Culverhouse-1996-ACFD}  & Private          & Algae                                          & 23               & \begin{tabular}[c]{@{}c@{}}MLP\\ RBF\end{tabular}              & \begin{tabular}[c]{@{}c@{}}MLP classification rate = 66\%\\ RBF classification rate = 83\%\end{tabular}                                                                           \\
1996 & S    & \cite{Shabtai-1996-MMMC}      & Private          & \textit{Aureobasidium pullulans}               & 2                & SOM                                                            & \textbackslash{}                                                                                                                                                                  \\
1998 & C    & \cite{Veropoulos-1998-IPNC}   & Private          & \textit{Tuberculosis}                          & 2                & MLP                                                            & Accuracy = 97.9\%                                                                                                                                                                 \\
1998 & C    & \cite{Gerlach-1998-IRSM}      & Private          & \textit{Aspergillus awamori}                   & 4                & MLP                                                            & Best classification results \textgreater 90\%                                                                                                                                     \\
1998 & C    & \cite{Blackburn-1998-RDBA}    & Private          & Bacteria                                       & 2                & MLP                                                            & \textbackslash{}                                                                                                                                                                  \\
1998 & E    & \cite{Wit-1998-AAAN}          & Private          & Microorganisms                                 & 2                & MLP                                                            & Accuracy \textgreater 93\%                                                                                                                                                        \\
1999 & C    & \cite{Kay-1999-TAAS}          & Private          & \textit{Gyrodactulus}                          & 5                & MLP                                                            & Detection rate = 100\%                                                                                                                                                            \\
2000 & C    & \cite{Culverhouse-2000-DAMV}  & DIB              & \textit{Dinoflagellates}                       & 23               & \begin{tabular}[c]{@{}c@{}}MLP\\ RBF\end{tabular}              & \begin{tabular}[c]{@{}c@{}}RBF Accuracy = 83\%\\ MLP Accuracy = 65\%\end{tabular}                                                                                                 \\
2003 & C    & \cite{Embleton-2003-ACPP}     & Private          & Phytoplankton                                  & 4                & MLP                                                            & Within 10\% of manual analysis                                                                                                                                                    \\
2005 & C    & \cite{Widmer-2005-UANN}       & Private          & \textit{Cryptosporidium parvumGiardia lamblia} & 2                & MLP                                                            & \textit{\begin{tabular}[c]{@{}c@{}}Cryptosporidium: Identification rate  = 91.8\%\\ Giardia: Identification rate  = 99.6\%\end{tabular}}                                          \\
2005 & C    & \cite{Weller-2005-SCSO}       & Private          & Sedimentary organic matter                     & 11               & MLP                                                            & Average classification rate = 87\%                                                                                                                                                \\
2006 & C    & \cite{Hu-2006-AAQT}           & Private          & Plankton                                       & 7                & LVQ-NN                                                         & Reduce the error by 50\% to 100\%                                                                                                                                                 \\
2006 & C    & \cite{Ginoris-2006-RPMU}      & Private          & Protozoa and metazoa                           & 5                & MLP
 & \textbackslash{}                                                                                                                                                                  \\
2007 & C    & \cite{Ginoris-2007-RPMU}      & Private          & Protozoa and metazoa                           & 22               & MLP                                                            & Overall recognition rate \textgreater 80\%                                                                                                                                        \\
2007 & C    & \cite{Xiaojuan-2007-ANBC}     & NMCR             & Bacteria                                       & 8                & MLP                                                            & Accuracy = 82.3\%                                                                                                                                                                 \\
2007 & C    & \cite{Xiaojuan-2007-ANBR}     & NMCR             & Bacteria                                       & 8                & MLP                                                            & Accuracy = 86.3\%                                                                                                                                                                 \\
2007 & C    & \cite{Weller-2007-TSNN}       & Private          & Sedimentary organic matter                     & 11               & \begin{tabular}[c]{@{}c@{}}RBF\\ MLP\end{tabular}              & Average classification rate = 91\%                                                                                                                                                \\
2008 & C    & \cite{Xiaojuan-2008-ANWB}     & CECC             & Bacteria                                       & 4                & MLP                                                            & Accuracy = 85.5\%.                                                                                                                                                                \\
2008 & C    & \cite{Cunshe-2008-ANWB}       & CECC             & Bacteria                                       & 4                & MLP                                                            & Accuracy = 85.5\%.                                                                                                                                                                \\
2008 & C    & \cite{Amaral-2008-SPII}       & Private          & Protozoa                                       & 8                & MLP                                                            & \textbackslash{}                                                                                                                                                                  \\
2009 & C    & \cite{Xiaojuan-2009-AIBN}     & CECC             & Bacteria                                       & 4                & MLP                                                            & Accuracy = 85.5\%.                                                                                                                                                                \\

2010 & C    & \cite{Hiremath-2010-AICB}     & Private          & Bacterial cell growth phases                   & 3                & RBF                                                            & Average accuracy = 97\%                                                                                                                                                           \\

2010 & C    & \cite{Hiremath-2010-DIAC}     & Private          & Cocci bacteria                                 & 6                & RBF                                                            & Average accuracy = 98.83\%                                                                                                                                                        \\
2010 & C    & \cite{Kumar-2010-RDMU}        & Private          & Microorganism                                  & 5                & PNN                                                            & Accuracy = 100\%                                                                                                                                                                  \\

\hline
Year & Task & Ref.                   & Data             & Object                                         & Class            & Method                                                         & Result                                                                                                                                                                            \\ \hline
2010 & C    & \cite{Zeder-2010-AQAA}        & Private          & Environmental bacteria                         & 3                & MLP                                                            & Identification rate = 94\%.                                                                                                                                                       \\
2010 & C    & \cite{Osman-2010-DMTZ}        & Private          & Mycobacterium tuberculosis                     & 2                & HMLP                                                           & \begin{tabular}[c]{@{}c@{}}Accuracy = 98.07\%\\ Sensitivity = 100\%\\ Specificity = 96.19\%\end{tabular}                                                                          \\
2010 & S    & \cite{Osman-2010-STBZ}        & Private          & Tuberculosis                                   & 3                & HMLP                                                           & Overall accuracy = 98.64\%                                                                                                                                                        \\
2010 & FE    & \cite{Zhu-2010-BCUN}          & \textbackslash{} & Bacteria                                       & \textbackslash{} & PCNN                                                           & \textbackslash{}                                                                                                                                                                  \\
2011 & C    & \cite{Hiremath-2011-ICCB}     & Private          & Cocci bacteria                                 & 6                & RBF                                                            & Average accuracy = 98.5\%                                                                                                                                                         \\
2011 & C    & \cite{Hiremath-2011-DMIA}     & Private          & Spiral bacteria                                & 3                & RBF                                                            & Accuracy = 100\%                                                                                                                                                                  \\
2011 & C    & \cite{Osman-2011-TBDZ}        & Private          & Tuberculosis                                   & 3                & MLP                                                     & Accuracy = 77.25\%                                                                                                                                                                \\
2011 & C    & \cite{Osman-2011-HMPN}        & Private          & Tuberculosis                                   & 3                & HMLP                                                           & Accuracy = 74.62\%                                                                                                                                                                \\
2011 & C    & \cite{Rulaningtyas-2011-ACTB} & \cite{Forero-2004-ITBB} & Tuberculosis                 & 2                & MLP                                                            & Mean square error = 0.000368                                                                                                                                                      \\
2011 & C    & \cite{Kiranyaz-2011-CRMI}     & Private          & Macroinvertebrate                              & 8                & \begin{tabular}[c]{@{}c@{}}MLP\\ RBF\end{tabular} & MLP performs best                                                                                                                                                                 \\
2012 & C    & \cite{Osman-2012-OSEL}        & Private          & Tuberculosis                                   & 3                & MLP                                                     & Accuracy =  91.33\%                                                                                                                                                               \\
2012 & C    & \cite{Mosleh-2012-APSA}       & Private          & Algae                                          & 5                & MLP                                                            & Overall accuracy = 93\%                                                                                                                                                           \\
2012 & C    & \cite{Hiremath-2012-SBCI}     & Private          & Spiral bacteria                                & 3                & RBF                                                            & Accuracy = 100\%                                                                                                                                                                  \\
2012 & C    & \cite{Siena-2012-DATB}        & Private          & Tuberculosis                                   & 2                & MLP                                                            & Accuracy about 88\%                                                                                                                                                               \\
2013 & C    & \cite{Danping-2013-IPMS}      & Private          & \textit{Powdery mildew spores}                 & 2                & MLP                                                            & Correct rate = 63.6\%                                                                                                                                                             \\
2013 & C    & \cite{Schulze-2013-PAAS}      & Private          & Phytoplankton                                  & 10               & MLP                                                            & \begin{tabular}[c]{@{}c@{}}Average recognition rate = 94.7\%\\ Average error rate = 5.5\%\end{tabular}                                                                            \\
2014 & C    & \cite{Coltelli-2014-WMAR}     & Private          & Microalgae                                     & 23               & SOM                                                            & \begin{tabular}[c]{@{}c@{}}Average recognition rate = 94.7\%\\ Average error rate = 5.5\%\end{tabular}                                                                            \\
2015 & C    & \cite{Priya-2015-AITO}        & Private          & Tuberculosis                                   & 2                & \begin{tabular}[c]{@{}c@{}}RBF\\ ANFIS\\ CANFIS\end{tabular}   & CANFIS is effective                                                                                                                                                               \\
2015 & C    & \cite{Kruk-2015-CCSI}         & Private          & Soil microorganisms                            & 12               & \begin{tabular}[c]{@{}c@{}}MLP\\ RBF\end{tabular}              & \begin{tabular}[c]{@{}c@{}}MLP accuracy = 98.97\%\\  RBF accuracy = 94.78\end{tabular}                                                                                            \\
2015 & S    & \cite{Costa-2015-AITM}        & Private          & Tuberculosis                                   & 2                & MLP                                                            & \begin{tabular}[c]{@{}c@{}}Sensitivity = 91.53\%\\  Error rate = 5.20\%\end{tabular}                                                                                              \\
2016 & C    & \cite{Priya-2016-AOIL}        & Private          & Tuberculosis                                   & 2                & MLP          & \begin{tabular}[c]{@{}c@{}}Object-level: Accuracy = 91.3\%\\  Image-level: Accuracy = 92.5\%\end{tabular}                                                                         \\ \hline

\label{table3}
\end{longtable}
\end{landscape}

\section{Deep Neural Network for Microorganism Image Analysis}
\label{DNN}

The MIA tasks based on deep neural networks are introduced in this section. The discussion is provided according to different tasks, including classification, segmentation, feature extraction, detection,  counting, and data augmentation. Same as the MIA works based on classical neural networks, papers related to the classification task occupy most of the MIA works based on the deep neural network. Therefore, the detailed analysis is presented according to different neural networks. In the segmentation part, U-Net is the most popular network. Thus, the details are provided from U-Net and other ANNs. Nevertheless, the networks used in other tasks are more fragmented, so the related papers are discussed individually. Additionally, a summary is provided to summarize the characters of the MIA based on deep neural networks and a table for readers to find relevant papers conveniently.

\subsection{Classification Task}

\subsubsection{Classification based on CNN}

Among the MIA tasks based on deep neural networks, the networks consisting of convolutional filters are divided into the category of CNN, excluding already named networks, such as AlexNet, ResNet, and DeepSea. Due to the application of convolutional filters in CNNs, the MIA analysis is able to no longer rely on feature engineering, and the workflow usually includes little or no image pre-processing and feature extraction. Therefore, the introduction of microorganism image classification based on CNN mainly focuses on microorganism category, methodology, and performance.

\begin{figure}[htbp!]
\centering
\includegraphics[width=0.7\linewidth]{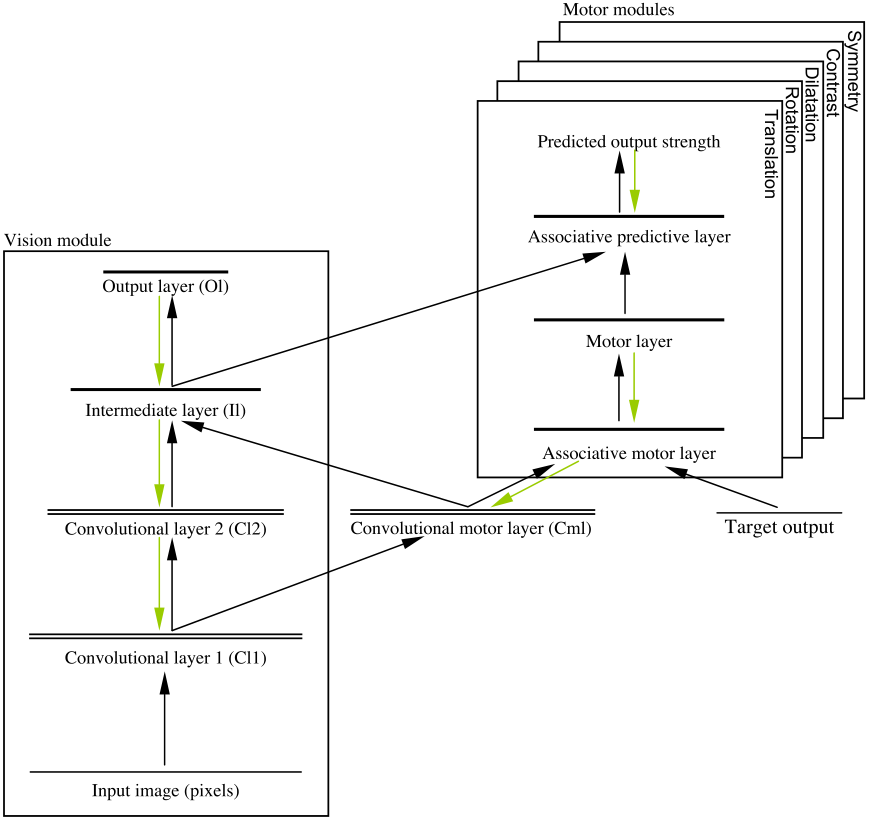}
\caption{The novel CNN in~\cite{Beaufort-2004-ARCD}.}
\label{fig18}
\end{figure}

The recognition system of plankton is challenging to design, mainly because most traditional methods rely on feature engineering. When the recognition task relates to multiple categories, extracting and selecting practical features will be a tough challenge. To optimize this situation, \cite{Dollfus-1999-FNNR} proposes a recognition system, called SYRACO2, to perform the classification task of coccoliths. A simple CNN containing two convolutional layers and two fully connected layers is designed in this system. Unlike classical neural networks, the application of the convolutional filter enables the system to no longer rely on feature engineering. The coccolith data used in the experiment contains 13 species and one non-coccolith class. Each coccolith species has about 100 images, and the non-coccolith class has about 600 images. The average effective recognition rate of the system in 13 species is 86\%. As an extension, \cite{Beaufort-2004-ARCD} introduces a novel CNN with a parallel structure to improve the identification rate. The structure is shown in Fig.~\ref{fig18}. Based on the network proposed in SYRACO2~\cite{Dollfus-1999-FNNR}, motor modules are introduced in the parallel neural network. The motor modules can dynamically achieve the translation, rotation, dilatation, contrast, and symmetry of the images through training. Besides, the novel system introduces secondary classification, which can achieve better classification. The results show that it recognizes approximately 96\% of coccoliths belonging to 11 Pleistocene taxa during routine work.

\begin{figure}[htbp!]
\centering
\includegraphics[width=0.95\linewidth]{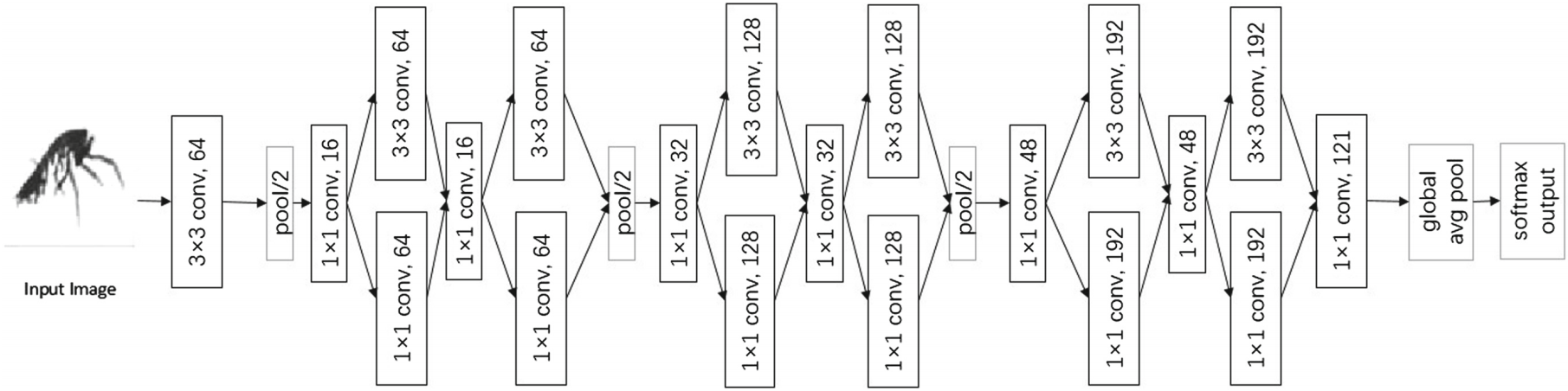}
\caption{The proposed CNN in~\cite{Yan-2017-AMEC}.}
\label{fig39}
\end{figure}

Besides, \cite{Lee-2016-PCIL,Py-2016-PCDC,Yan-2017-AMEC,Luo-2018-APIA} also employ CNNs for the classification of plankton. \cite{Lee-2016-PCIL} adopts a plankton image dataset, WHOI-Plankton, which consists of 3.4 million expert-labeled plankton images with 103 classes. Nevertheless, It has a seriously imbalanced problem. Therefore, a CNN based on transfer learning is introduced to solve it. A balanced sub-dataset is extracted from WHOI-Plankton to pre-train the network. Then, the original dataset is used to fine-tune the network. The results show that this method achieves the classification accuracy and F1-score of 92.80\% and 33.39\%. \cite{Py-2016-PCDC} converts the original image into multiple sizes to train a simple CNN in parallel so that it can adapt to different inputs. The dataset used in this study is Plankton Set 1.0. The best model achieves a softmax loss of 0.613. The performance needs to be further improved. Kaggle data science competition in 2015 released a plankton dataset, which has 30336 images of 121 categories. \cite{Yan-2017-AMEC} proposes a novel CNN based on the exploration of classical CNN models, such as CaffeNet, VGG-19, and ResNet. The structure is shown in Fig.~\ref{fig39}. The results show that the CNN significantly reduces the model size and improves the frame rate while achieving similar performance. \cite{Luo-2018-APIA} uses a CNN, which consists of 13 convolutional and 12 pooling layers, to perform plankton classification. The Kaggle plankton dataset is used to evaluate the proposed CNN. The results show that an utterly random classification assessment obtains an average accuracy of 84\% and a recall of 40\% for all groups.

Bacteria classification plays an essential role in many applications, such as disease diagnosis and food production. However, manual methods are ineffective. Thus, there are some works using CNNs to assist this process. \cite{Balagurusamy-2019-DDBD} develops a smartphone optical device, which uses a CNN-based method to detect bacteria and analyze their motion. The CNN consists of two convolution layers and two fully connected layers. The results show that it can achieve an accuracy of 83\% in the classification of \emph{E.coli} and \emph{B.subtilis}. \cite{Polap-2019-BSCU} introduces a model based on region covariance and CNN to classify bacteria. The region covariance is used to segment the input image, in which texture comparison is used. These segments are fed into CNN for classification. The experimental data contain three categories: rod-shaped bacteria, round or nearly round shape bacteria, and mixed bacteria strains. The results show that the classification accuracy values in the rod-shaped bacteria and spherical or nearly spherical shape bacteria exceed 91\% and 78\%, respectively. \cite{Bliznuks-2020-ENNS} proposes a system that uses 3D CNN to identify bacterial growth. Different from others, laser speckle images are the objects of analysis. The 3D CNN can encode spatial speckle variance and their changes in time. The results show that it reaches an accuracy of 0.95. \cite{Chopra-2020-NMBC} evaluate several different CNN training strategies to find out the most suitable one in classifying bacteria images in four categories. The evaluated models contain simple CNN, simple CNN optimized by particle swarm optimization, autoencoders pre-training the CNN, autoencoders pre-training the CNN optimized by particle swarm optimization, and CNN optimized by genetic algorithms. The results show that autoencoders pre-training the CNN optimized by particle swarm optimization performs best with the accuracy and F1-score of 94.9\% and 95.6\%. In~\cite{Mhathesh-2020-A3CN}, a method based on 3D CNN is presented to perform bacterial image classification. The results show that it obtains an accuracy of 95\%. \cite{Tamiev-2020-ACBC} proposes a CNN-based method to classify bacterial cells. \emph{B. subtilis} image data is used in this study. In the CNN-based method, the original image is first processed with a binary segmentation algorithm and annotated manually, and then these images are augmented to train and test the CNN. The results show that the proposed CNN can achieve 86\% accuracy.

\begin{figure}[htbp!]
\centering
\includegraphics[width=0.95\linewidth]{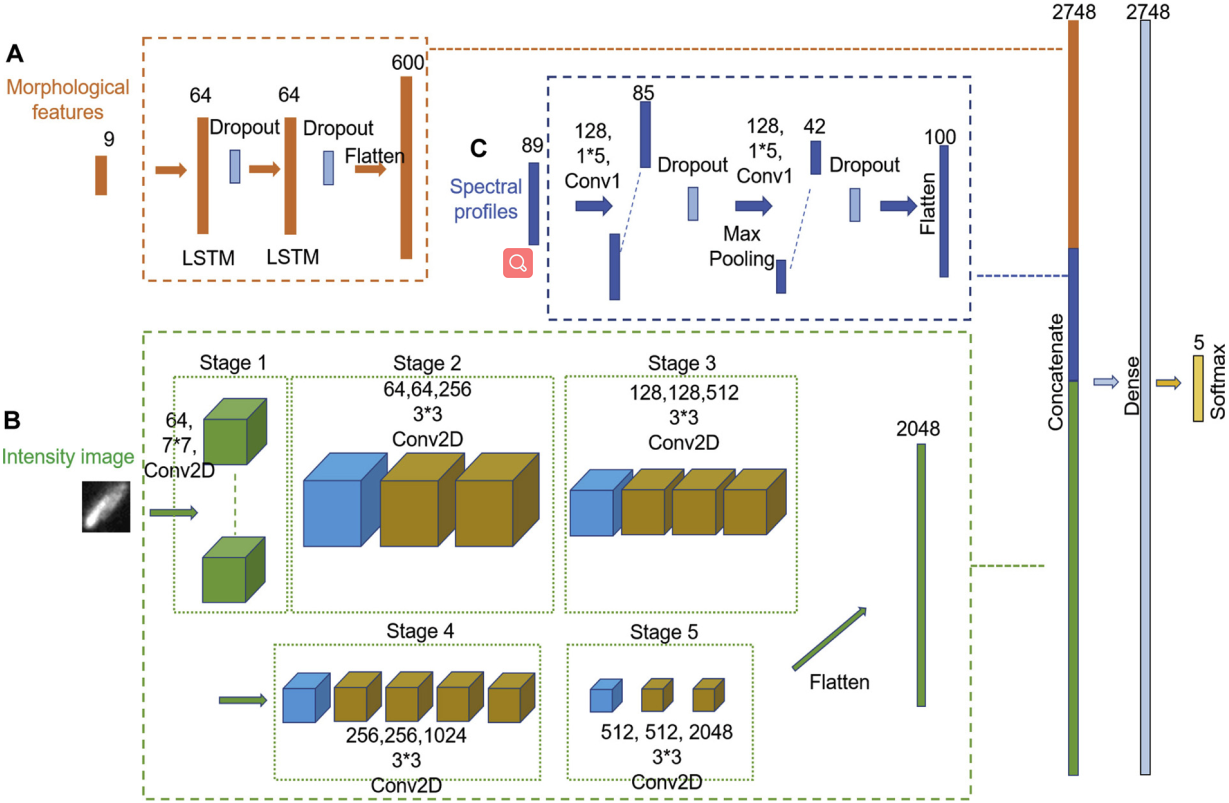}
\caption{Fusion-Net framework in~\cite{Kang-2020-SCFP}. (A) LSTM framework for morphological feature analysis; (B) ResNet framework for intensity image analysis; (C) 1D-CNN framework for spectral profile analysis.}
\label{fig61}
\end{figure}

Besides, some bacteria classification works adopt both CNNs and other networks. In~\cite{Nie-2015-ADFB}, a framework for bacteria segmentation and classification is proposed. In this framework, the original images are cropped into patches, and \emph{Convolutional Deep Belief Network} (CDBN) is used to extract the patch-level features for training SVM to achieve the patch-level segmentation, which aims to distinguish foreground and background. After that, the foreground patches are further classified by a simple CNN. The predicted labels vote for the final bacterial category. The data consists of 17 species. The results show that the framework achieves an accuracy of 97.14\% in patch-level segmentation and an accuracy of 62.10\% in classification. In~\cite{Kang-2020-CFBU}, a food-borne pathogenic bacteria classification method based on CNNs and hyperspectral microscope imaging is proposed. The data contains five species. Hyperspectral microscope imaging technology can obtain both spatial and spectral information. There are two CNN models used in this work: U-Net and 1D CNN. U-Net is first applied to segment the ROIs, and 1D CNN is used to classify the bacteria using the spectral information extracted from the ROIs. The results show that U-Net achieves the average accuracy of 96\% and the average mIOU of 88\%, and 1D CNN obtains the classification accuracy of 90\%. As an extension of~\cite{Kang-2020-CFBU}, \cite{Kang-2020-SCFP} proposes a hybrid deep learning framework defined as ``Fusion-Net". In the proposed framework, the first step is data acquisition, and then three features, including morphological features, intensity images, and spectral profiles, are extracted from the image data. These features are used to train three networks: LSTM, ResNet, and 1D CNN. The schematic diagram is shown in Fig.~\ref{fig61}. The results show that the three networks achieve classification accuracies of 92.2\%, 93.8\%, and 96.2\%, respectively. After the fusion, the classification accuracy is increased to 98.4\%.

In addition to plankton and bacteria, CNN is also used in TB classification. TB is the pathogenic bacterium that causes tuberculosis. \cite{Swetha-2020-CNNB} proposes a framework for automatic and rapid classification of TB in sputum images. In this framework, the image is pre-processed by applying noise reduction and intensity modification, and then the segmentation is done by the \emph{Channel Area Thresholding} (CAT). After that, the HOG and SURF features are extracted from the segmented images to train the CNN classier. The experimental results show that the proposed framework achieves an accuracy of 99.5\%, a sensitivity of 94.7\%, and a specificity of 99\%.

Through the above discussion, it can be found that some CNN-based works~\cite{Dollfus-1999-FNNR,Beaufort-2004-ARCD,Balagurusamy-2019-DDBD} adopt simple and shallow structures. This is because of the limitation of hardware performance and available image data. Due to the significant improvement of computer performance and available data, most CNN structures adopted in the recent decade are deeper and more complex, such as the CNNs in~\cite{Yan-2017-AMEC,Luo-2018-APIA}. Besides the difference in network structures, \cite{Lee-2016-PCIL} introduces CNN-based transfer learning to solve the imbalance problem of microorganism image data. Additionally, one of the most significant advances of CNN is that the convolutional filter enables the network to no longer rely on feature engineering. It should be noted that although the convolutional filter can make CNN no longer rely on feature engineering, there are still some works that use feature engineering to further improve the performance of CNN, such as~\cite{Swetha-2020-CNNB}.

\subsubsection{Classification based on AlexNet}
AlexNet is one of the most classical deep neural networks. Its basic information and structure are provided in Sec.~\ref{AlexNet}. It is also widely used in MIA tasks, especially the classification task. In~\cite{Dai-2016-AHCN}, a hybrid CNN model is proposed for plankton classification. The network consists of three branches based on AlexNet, and their inputs are original images, global feature images, and local feature images. At the bottom of the network, fully connected layers connect these branches, and softmax is applied for classification. The dataset used here is WHOI-Plankton. The experimental results show that the hybrid CNN based on AlexNet achieves an accuracy of 95.83\%. Similarly, \cite{Cui-2018-TSIF} introduces a novel texture feature extraction method and a hybrid CNN based on AlexNet for plankton classification. The original image is used to generate texture and shape images, and they are concatenated with the original image. The concatenated image is used for AlexNet training. The dataset used in this study is also WHOI-Plankton, and the experimental results show that the network with three inputs obtained the best accuracy of 96.58\% in 30 categories and 94.32\% in 103 categories. \cite{Pedraza-2017-ADCA} utilizes the transfer learning technology based on AlexNet pre-trained by ImageNet to perform the classification of diatom. The performance of the network fine-tuned by different forms of data is studied. The results show that when the original data are combined with the normalized data, the network achieves the best average accuracy of 99.51\%.

\subsubsection{Classification based on VGGNet}
\begin{figure}[htbp!]
\centering
\includegraphics[width=0.95\linewidth]{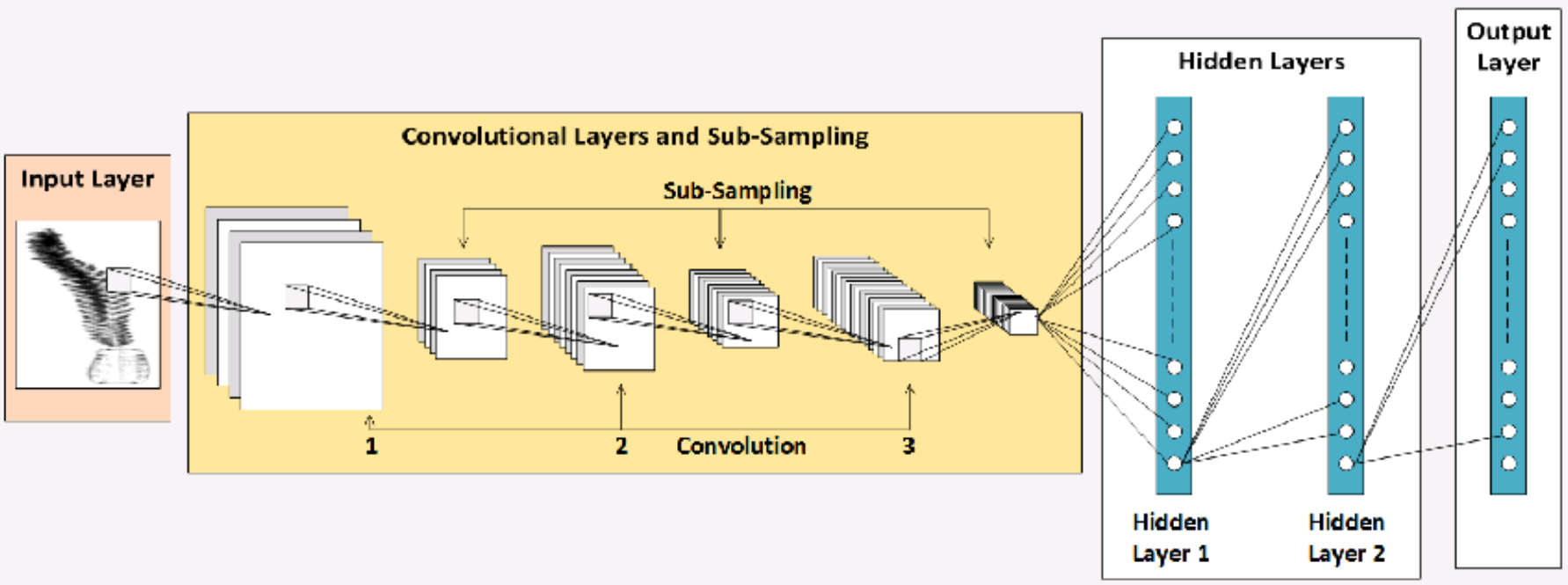}
\caption{The proposed network structure in~\cite{Al-2018-IPIC}.}
\label{fig41}
\end{figure}

VGGNet is also popular in many image analysis tasks. Its characteristic is provided in Sec.~\ref{VGGNet}. VGGNet has several different types of structures. Their related classification works are summarised in this part. In~\cite{Al-2018-IPIC}, a CNN based on VGG-16 is proposed to classify plankton, and its structure is shown in Fig.~\ref{fig41}. To evaluate the network's performance, three sub-datasets from SIPPER are used. The experimental results show that it achieves an accuracy of 80.54\% on SIPPER-77. \cite{Wang-2018-TPCN} presents a transferred parallel neural network for large-scale imbalanced plankton dataset classification. It consists of a pre-trained model based on small classes and an untrained model. In the training process of the network, the data is fed into the network for training, and the pre-trained model is used as a feature extractor to enhance the features of small classes to improve the classification ability on the imbalanced dataset. The dataset is WHOI-Plankton, and the results show that the transferred parallel neural network based on VGG-16 achieves the best accuracy of 94.98\% and F1-score of 54.44\%. Besides, in~\cite{Zawadzki-2020-DLAC}, several popular CNN models with or without pre-trained weights are compared and analyzed in the microorganism (bacteria and fungi) classification task. These models include VGG-16, VGG-19, Xception, ResNet-50, Inception-V3, MobileNetV2, and DenseNet201. The experimental results show that Xception and ResNet-50 perform well, and it is not always beneficial to use the pre-trained weights of CNN models.

\subsubsection{Classification based on Inception Series}
Inception series also play an essential role in many image analysis tasks. The introduction of Inception is provided in Sec.\ref{Inception}. Some works use Inception-related networks to perform the microorganism image classification task. For instance, \cite{Wahid-2018-CMIB} uses the modified deep CNN model based on Inception-V1 to realize bacterial classification. The network is pre-trained by a million images, and then the network is re-trained by the dataset, which has five bacteria species selected from several online resources. Results show that the network achieves an accuracy of about 95\%. As an extension, the Xception-based bacteria classification method is tested in~\cite{Wahid-2019-DCNN}. Xception is pre-trained by ImageNet and then fine-tuned by experimental data, which collects a total of 740 images in seven categories from several online datasets. The experimental results show that the Xception-based method achieves an accuracy of 97.5\%.  Besides, \cite{Zawadzki-2020-DLAC} also compares Xception and Inception-V3 with other networks. The details are provided in the last part.

\subsubsection{Classification based on other deep ANNs}
In addition to the above networks, some works adopt other deep ANNs to classify microorganism images. Nevertheless, due to the limited number of each network, we directly analyze each related paper and provide detailed discussions about the research motivation, contribution, data, workflow, and result.

Rapid identification of microbial pathogens is of great significance in treating infection. A method to distinguish bacterial species using 3D  refractive index images is presented in~\cite{Kim-2018-AIBU}. DenseNet and \emph{Wide Residual Network} (WRN)~\cite{Zagoruyko-2016-WRN} are compared in this study. The data used in the experiment consisted of seven categories, each containing more than a thousand images. The experimental results show that the highest accuracy obtained by WRN is 73.2\%, and the highest accuracy obtained by DenseNet is 85\%.

Plankton image classification is of great significance to the study of plankton. In~\cite{Liu-2018-DPRN}, a CNN called PyramidNet for plankton image classification is proposed. The dataset used in this study is WHOI-Plankton. The experimental results show that PyramidNet performs better than AlexNet, GoogLeNet, VGG-16, and ResNet. It gets the best performance with 86.30\% accuracy and 41.64\% F1-score.

Analysis of viruses using transmission electron microscopy images is an essential step in understanding the formation mechanism of infectious virions. In~\cite{Devan-2019-DHCT}, an investigation for comparing the performance of a CNN that was trained from scratch with pre-trained CNN models as well as existing image analysis methods are produced. The private data, which contains 190 images for training and verification and 21 images for testing, is used in this study. The experimental results show that the pre-trained ResNet50 fine-tuned by virus image data obtains the best performance with an accuracy of 95.44\% and an F1-score of 95.22\%.

Different species of bacteria have different effects on humans, so it is essential to distinguish them. In~\cite{Rujichan-2019-BCUI}, deep CNN is used to classify 33 bacteria species in the DIBaS dataset. In this work, to improve the classification accuracy, some pre-processing operations, including color masking and image augmentation, are applied first, and then the MobileNetV2 based on ImageNet weight is used as the base model for classification. The experimental results show that the model achieves an average accuracy of 95.09\%.

The possibility of using image classification and deep learning methods to recognize the standard and high-resolution bacteria and yeast images is studied in~\cite{Treebupachatsakul-2020-MIRB}. The standard resolution data used in this study is provided by the authors and includes three classes of bacteria and one class of yeast, each with more than 200 images. High-resolution images are taken from similar bacteria and yeasts in~\cite{Zielinski-2017-ALAB}. The network used in the experiment is LeNet. The experimental results show that more than 80\% accuracy can be obtained from standard resolution data.

\subsection{Segmentation Task}
In addition to classification task, segmentation task plays a vital role in MIA tasks based on deep neural networks. U-Net is the most popular network used in the segmentation task. Besides, there are some works adopting other networks for segmentation. To facilitate the analysis, we introduce the segmentation from U-Net and others.
 
\subsubsection{Segmentation based on U-Net}
\begin{figure}[htbp!]
\centering
\includegraphics[width=0.95\linewidth]{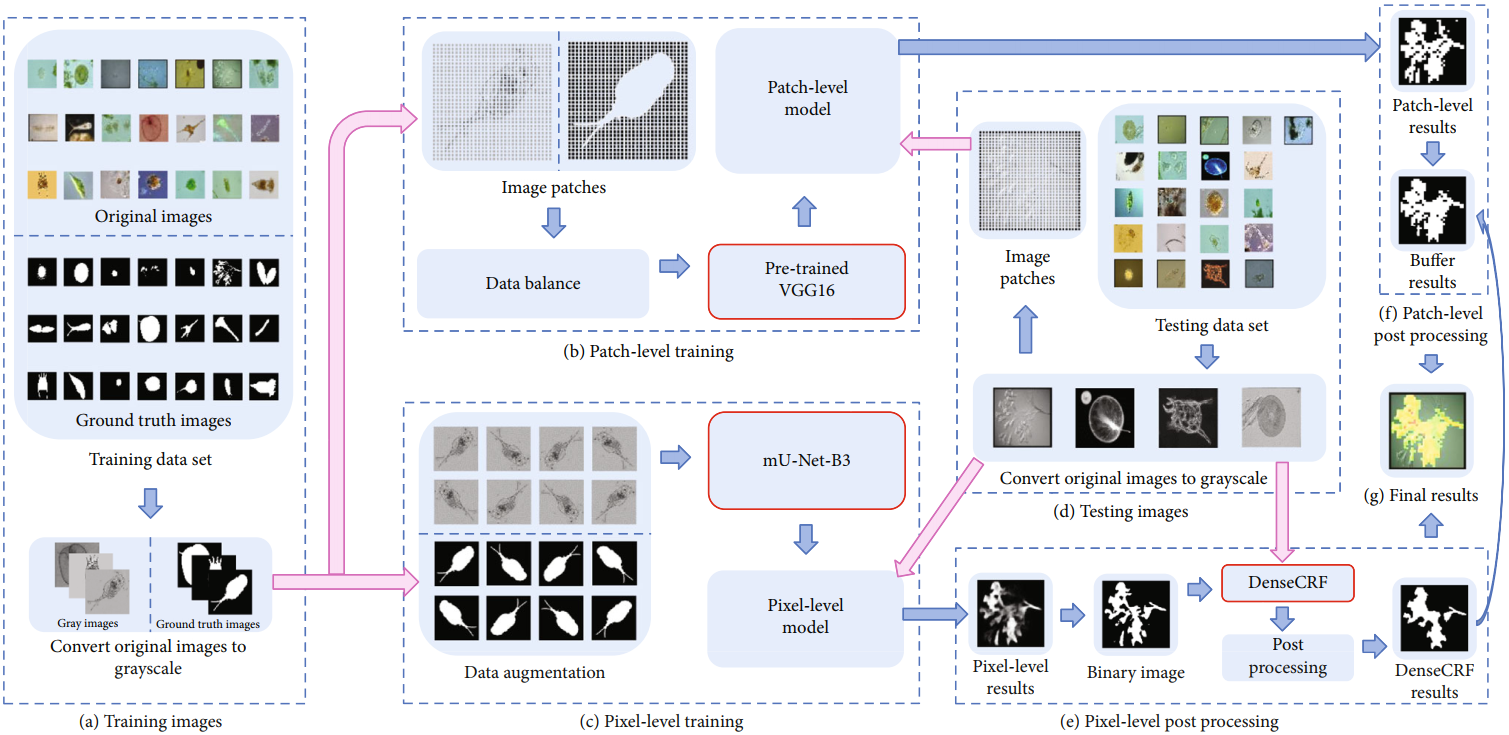}
\caption{The CNN-CRF proposed in~\cite{Zhang-2020-AMCF}.}
\label{fig59}
\end{figure}

U-Net is widely used in many image segmentation tasks, including microorganism image segmentation. Its essential information and structure are provided in Sec.~\ref{U-Net}. Due to the expensive image segmentation annotation, \cite{Matuszewski-2018-MATS} proposes a simple U-Net, which uses the minimum annotation, to perform the Rift Valley virus segmentation task. The dataset contains 143 TEM images~\cite{Kylberg-2012-SVPC}. The minimal annotation requires only simple annotation of the virus center, and then the ground truth images are generated by dilating operation. The results show that the network achieves a Dice of 90\% and an IOU of 83.1\%. In~\cite{Dietler-2020-ACNN}, a method for boundary identification of irregular yeast cells is proposed. In this method, U-Net is used to segment the cell objects from the background, but it cannot solve the adhesion problem of cells. Therefore, the watershed algorithm is combined with the prediction scores of U-Net to achieve the instance segmentation. The experimental results show that the proposed method achieves a mean accuracy of 94\%. \cite{Li-2020-MAMR} develops a \emph{Multiple Receptive Field U-Net} (MRFU-Net) to perform the environmental microorganism image segmentation task. This network is inspired by GoogLeNet and U-Net. The EMDS-5 dataset~\cite{Li-2020-ANMD,Li-2021-EMDS,Kulwa-2022-ANPD}, which contains 21 microorganism categories, is used to evaluate the segmentation performance. Experimental results show that the Dice, Jaccard, recall, accuracy, and VOE obtained by MRFU-Net are 87.23\%, 79.74\%, 87.65\%, 97.30\%, and 20.26\%, respectively. As an extension of work~\cite{Li-2020-MAMR}, a CCN-CRF framework is proposed to perform the environmental microorganism image segmentation task in~\cite{Zhang-2020-AMCF}. Fig.~\ref{fig59} provides the details of this framework. The framework includes two parts: pixel-level segmentation and patch-level segmentation. In pixel-level segmentation, a low-cost U-Net (called mU-Net-B3 or LCU-Net) is proposed to perform the pixel-level segmentation. It shows better performance and has less than one-third memory requirement than U-Net. In the patch-level segmentation, the transfer learning based on VGG-16 is used to perform the patch classification. These predicted labels are used to reconstruct the patch-level segmentation results. The experimental results show that the Dice, Jaccard, recall, accuracy, and VOE obtained in pixel-level segmentation are 87.13\%, 79.74\%, 87.12\%, 96.91\%, and 20.26\%, respectively.

\subsubsection{Segmentation based on other deep ANNs}

\begin{figure}[htbp!]
\centering  
\subfigure[Mosaic image]{
\includegraphics[width=0.35\textwidth]{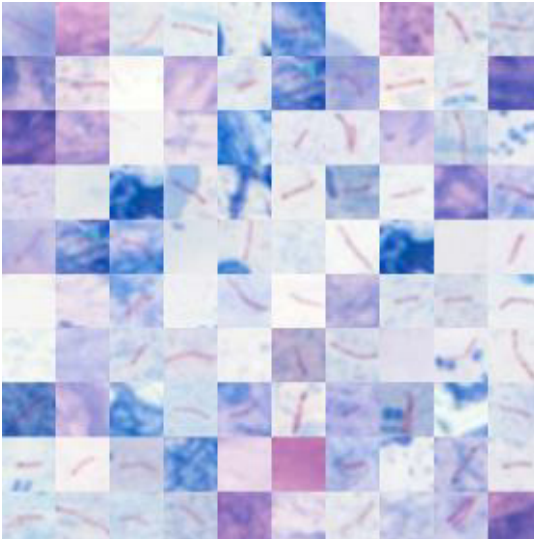}}
\subfigure[Ground truth]{
\includegraphics[width=0.35\textwidth]{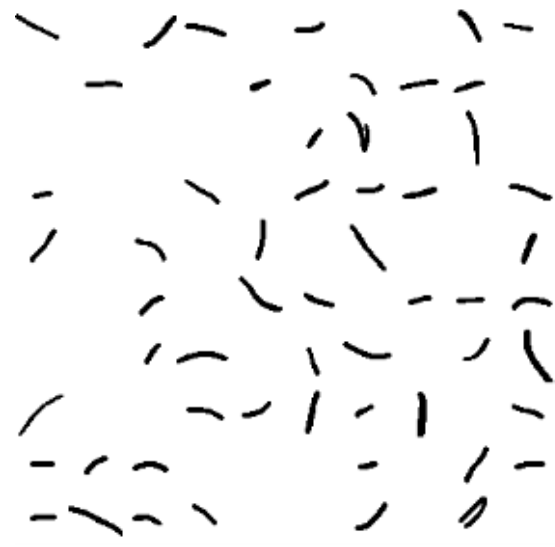}}
\caption{An example of mosaic image in~\cite{Serrao-2020-ABDL}.}
\label{fig51}
\end{figure}

\begin{figure}[htbp!]
\centering
\includegraphics[width=0.95\linewidth]{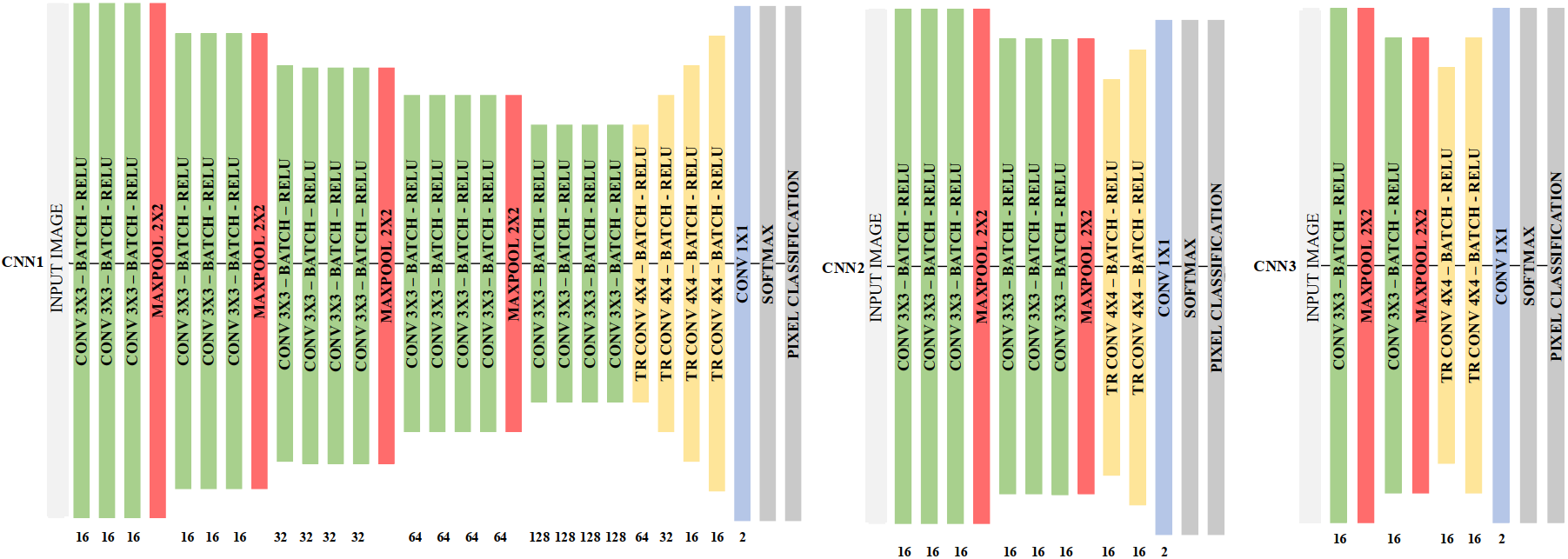}
\caption{The proposed network structures in~\cite{Serrao-2020-ABDL}.}
\label{fig53}
\end{figure}

Accurate segmentation in cell microscopy is one of the critical steps in cell analysis. In~\cite{Aydin-2017-CBTC}, a SegNet-based segmentation method is proposed and applied to the multi-modal fluorescent microscopy image data of yeast cells. The data used in the experiment is the fluorescence microscope images of yeast cell division, consisting of 6000 training samples, 1200 validation samples, and 1200 test samples. Experimental results show that the method achieves a mIOU of 71.72\%. Tuberculosis is one of the top 10 causes of death worldwide. In~\cite{Serrao-2020-ABDL}, a method based on CNN and a mosaic image approach is proposed. The data used in this study includes positive and negative patches. These patch data are used to generate a total of 5000 mosaic images. Each image comprises 100 patches, about half of which are negative, and another half are positive. Fig.~\ref{fig51} provides a mosaic image example. In the experiments, three CNNs are proposed to perform the segmentation task to achieve the bacillus detection, which aims to perform the bacilli count. Fig.~\ref{fig53} provides these networks' structures. The network performance is evaluated by counting the number of segmented bacilli. The experimental results show that the deepest CNN1 obtain the best performance with an accuracy of 99.665\%.

\subsection{Detection Task}
\begin{figure}[htbp!]
\centering
\includegraphics[width=0.85\linewidth]{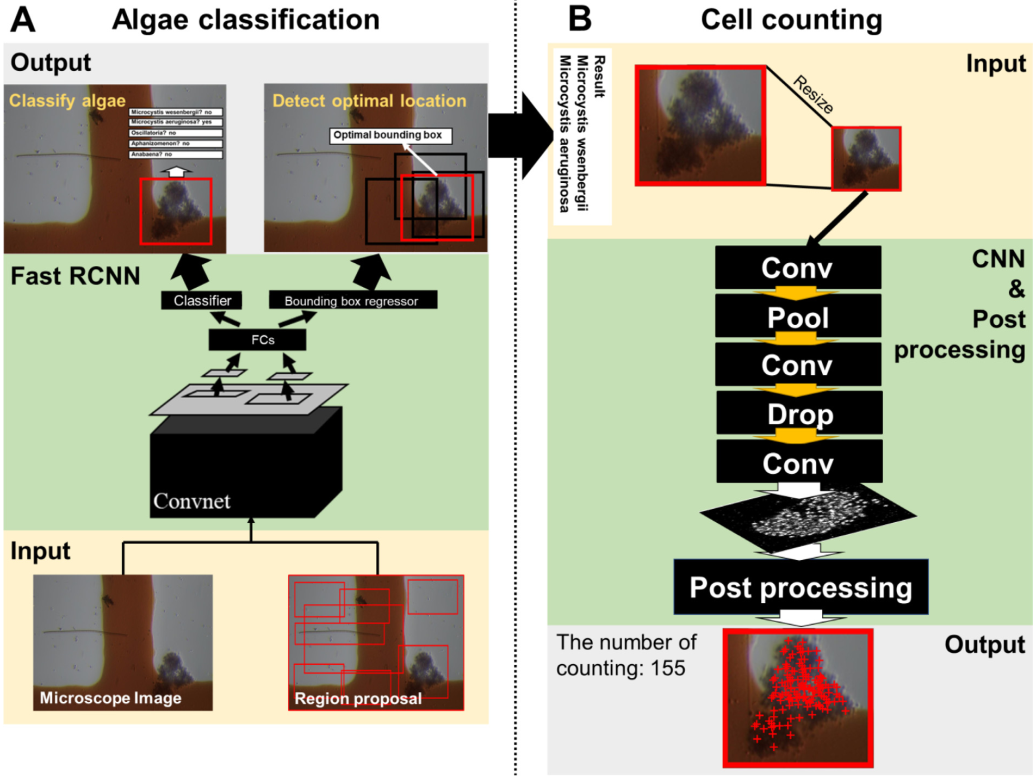}
\caption{The procedure of algae classification and cell counting in~\cite{Baek-2020-IECS}.}
\label{fig60}
\end{figure}

Due to microorganism image detection works adopting different deep ANNs and each network containing a limited number of works, the detailed analysis of each paper is presented individually. In~\cite{Hung-2017-AFRO}, a two-stage method of detection and classification is proposed to analyze malaria parasite images. In the detection stage, Faster RCNN is used to detect the target, which only distinguishes whether it is the red blood cell or not. In the second stage, AlexNet is used to classify the targets labeled as non-red blood cells. All networks are pre-trained by ImageNet and fine-tuned by the malaria parasite image data. The results show that an accuracy of 59\% is achieved in the detection stage, and 98\% is obtained in the second classification stage. Similarly, \cite{Baek-2020-IECS} introduces a framework based on Fast RCNN and CNN to classify and quantify five species of cyanobacteria. The framework is shown in Fig.~\ref{fig60}. Fast RCNN is used to detect and classify targets, and then CNN is used to perform the counting. The experimental results show that the classification accuracy values of fast RCNN for five species of cyanobacteria are 0.929, 0.973, 0.829, 0.890, and 0.890, and CNN obtained the $R^{2}$ value of 0.85, and RMSE of 23 cells in the counting task.

\begin{figure}[htbp!]
\centering
\includegraphics[width=0.95\linewidth]{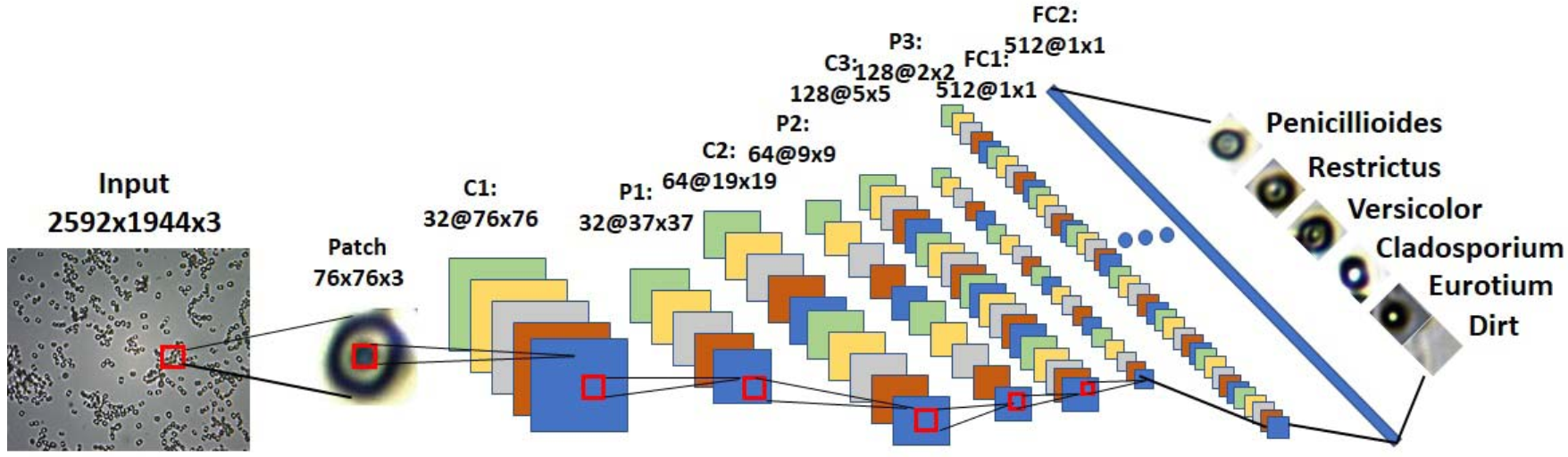}
\caption{The CNN for fungus detection in~\cite{Tahir-2018-AFSD}.}
\label{fig48}
\end{figure}

\cite{Tahir-2018-AFSD} proposes a new fungus dataset and develops a CNN-based method for fungus detection. The fungus dataset contains 40800 images of five classes of fungi and one extra class of dirt. This data is divided into 30000 training data and 10800 test data. The CNN proposed in this paper is shown in Fig.~\ref{fig48}, which can perform the detection of fungi and the classification of fungi. The experimental results show that the accuracy of detection is 94.8\%. In~\cite{Pedraza-2018-LPCN}, a comparison is conducted to test whether the latest deep learning network, including RCNN and YOLO, can adapt to the detection of diatoms. The data used here is private and contains nearly 11000 images in ten categories. The experimental results show that YOLO is more effective with an F1-score of 72\%, while the RCNN is only 10\%.

\subsection{Feature Extraction Task}
Besides the classification, segmentation, and detection tasks, some works employ deep neural networks to perform the image feature extraction task. The details are provided individually.

\begin{figure}[htbp!]
\centering
\includegraphics[width=0.98\linewidth]{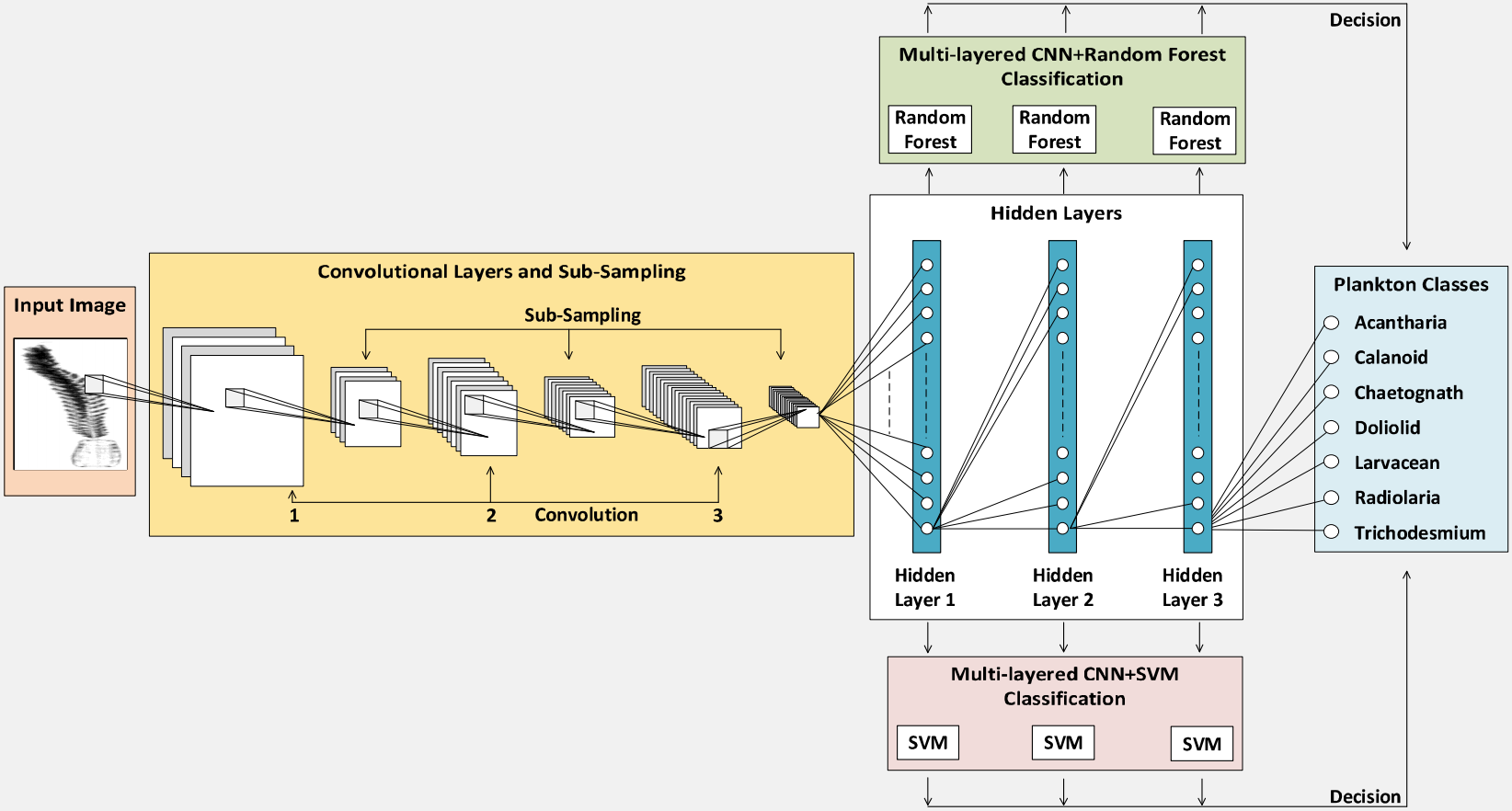}
\caption{The illumination of the hybrid algorithm in~\cite{Al-2015-PEHC}.}
\label{fig36}
\end{figure}

In~\cite{Al-2015-PEHC}, a hybrid plankton classification algorithm based on CNN is proposed. The illumination of the hybrid algorithm is shown in Fig.~\ref{fig36}. First, the plankton image data is used to train the CNN, which has three hidden layers. Then, the features generated by each hidden layer are combined with different classification methods, which include RF and SVM. The experimental results show that the SVM trained by the features generated by the first hidden layer achieves the best performance with an accuracy of 96.70\%.

\cite{Kosov-2018-EMCU} proposes a classification framework based on the deep neural network and CRF to classify environmental microorganisms. The EMDS-4 dataset is used, which contains 400 images of 20 categories. The framework combines the global and local features for building the CRF model. The local features are generated by DeepLab-VGG-16, which is a reorganized network of DeepLab and pre-trained VGG-16. The DeepLab-VGG-16 is trained by EMDS-4 to generate a feature vector for each pixel. Then, these feature vectors are fed into RF to perform as the unary potential of CRF. Compared with SIFT and simple features, the results show that the DeepLab-VGG-16 feature is superior.

In~\cite{Rodrigues-2018-ETLS}, transfer learning technology is used to remedy the problem of the small dataset. Two public datasets and one private dataset are used in this study. The ISIIS dataset is used to pre-train a DeepSea and the first AlexNet, while ImageNet is used to pre-train the second AlexNet. These pre-trained networks are used as feature extractors to obtain the image features of the small private dataset, and the SVM is used to evaluate these features. The results show that DeepSea with ISIIS performs best, achieving 84\% classification accuracy.

As an extension of works~\cite{Wahid-2018-CMIB}, a bacteria classification system based on Inception-V3 and SVM is presented in~\cite{Ahmed-2019-CDCN}. The data used in this study consists of seven categories. The pre-trained Inception-V3 is fine-tuned by about 800 training images in the system. After that, SVM performs the classification task with the fine-tuned Inception-V3 as the feature extractor. The experimental results show that the system achieves an accuracy of around 96\%.

In~\cite{Cheng-2019-ECNN}, an end-to-end framework for plankton image identification and enumeration is presented. An algorithm is proposed to extract and enhance the ROI from the input image first. Then, the local grayscale values are used to enhance the local features of ROIs. After that, CNN is used to extract the features from the enhanced ROIs. These features are fed into SVM for multi-class classification. In the experiments, CNNs, including AlexNet, VGGNet, GoogLeNet, and ResNet, are compared. The private data used in this study contains six plankton categories and one ‘other’ category. The experimental results show that the best accuracy (94.13\%) and recall (94.52\%) are obtained by ResNet50 with SVM.

In~\cite{Rawat-2019-ADLB}, a generic framework for plankton classification is proposed. The data used in this study contains 235 images in five categories. In this experiment, Inception-V3, VGG-16, and VGG-19 are used to extract common features, and then CNN, Logistic Regression, SVM, and KNN are used as the classifiers to test the performance of different features. The results show that the CNN with Inception-V3 feature extractor achieves the best accuracy of 99.5\%.

Plankton is one of the essential components in marine ecosystems. In~\cite{Lumini-2019-DLTL}, different transfer learning methods based on CNNs are investigated, aiming to design an ensemble plankton classifier based on their diversity. The transfer learning methods include one round tuning, two rounds tuning, and pre-processing tuning. The datasets used in this study include three public datasets (WHOI, ZooScan, and Kaggle) and one dataset used in two rounds of tuning. In the experiments, three transfer learning methods and the features extracted from the one round tuning are combined with SVM are tested. The experimental results show that the ensemble of models generated by one round tuning and two rounds tuning obtains the best performance. It achieves an accuracy of 0.9527 in the WHOI dataset, an accuracy of 0.8826 in the ZooScan dataset, and an accuracy of 0.9413 in the Kaggle dataset.

Since the convolution module is usually translational invariant, when the target rotates by a certain angle, the network will not recognize it. In~\cite{Cheng-2020-MTCN}, a method combining the translational and rotational features is proposed to address this problem. In this method, the original images (images in Cartesian coordinates) are transformed into polar images (images in polar coordinates) first. Then, the CNN models trained by polar coordinates and original images are used as a feature extractor. The classification is performed by SVM. In this study, the proposed method is tested on the in situ plankton dataset and the CIFAR-10 dataset. The experimental results show that the Densenet201+Polar+SVM model obtains the highest classification accuracy (97.989\%) and recall rate (97.986\%) on the in situ plankton dataset. On the CIFAR-10 dataset, it obtains the highest classification accuracy (94.91\%) and the highest recall rate (94.76\%).

\subsection{Data Augmentation Task}

Environmental microorganism image datasets usually contain limited data. In~\cite{Xu-2020-AEFG}, an enhanced framework of GANs is proposed to perform the environmental microorganism image data augmentation task. The dataset used in the experiment is EMDS-5, which contains 21 classes of microorganisms, with 20 images and their corresponding ground truth images for each class~\cite{Li-2021-EMDS}. Due to the different directions of microorganism images, it is challenging to generate images in various directions directly through GAN. Considering the small dataset, the images directly generated by GAN may lose many details. This framework transforms the original image in the same direction by combining it with the ground truth image to address the above problems. Besides, color space transformation is performed to increase the amount of training data and enable GAN to generate more details. To evaluate the framework's effectiveness, VGG-16 is trained by the generated images for classification, and the results show that the data generated by the proposed framework can effectively improve the classification performance.

\subsection{Summary}
After the above review, compared with the study based on classical neural network, we find that the MIA based on the deep neural network has the following characteristics: First, different from MIA based on classical neural networks, more open-access microorganism datasets are used, such as SIPPER, WHOI-Plankton, and EMDS series. Second, the deep neural networks used in MIA tasks are mainly CNN, whose convolutional filter can be applied directly to the image data. Therefore, unlike MIA based on classical neural networks, MIA based on deep neural networks usually do not rely on feature engineering. The neural network can automatically extract the effective features from the image through the convolutional filter, which is more flexible than artificial features. Besides, the structure of deep neural networks is usually deeper than that of classical neural networks because the convolution layer has fewer parameters than the fully connected layer. However, deep neural networks require lots of training data and high performance of the computer. Third, due to the flexibility of deep neural network architecture, the MIA tasks based on deep neural networks are more varied, such as microorganism image data augmentation, detection, and counting. In addition, compared with MIA based on classical neural networks, more MIA works based on deep neural networks focus on optimizing and improving the deep neural networks.

To help readers find and understand relevant works quickly, we briefly summarize the papers related to MIA based on deep neural networks in Tab.~\ref{table4}, in which years, application tasks, references, datasets, object species, category number, methods, and results are provided.

\section{Methodology Analysis and Potential Direction}
\label{MA}
The papers related to MIA based on classical and deep neural networks are analyzed in Sec.~\ref{CNN} and Sec.~\ref{DNN}. To further understand the characteristics of these works, a corresponding in-depth analysis is provided in this section. Besides, we also discuss the research limitations and potential development directions.

\subsection{Analysis of Methods Based on Classical Neural Networks}

The works related to MIA based on classical neural networks are analyzed according to their target tasks, including classification, segmentation, counting, and feature extraction tasks. Due to most classical neural networks relying on feature engineering, the analysis of most works is provided from three parts: image pre-processing, feature extraction, and network analysis.

In the classification tasks, the image pre-processing operations mainly include noise reduction, image enhancement, and segmentation. For example, the pre-processing operations used in~\cite{Culverhouse-1994-ACFS,Ginoris-2007-RPMU,Mosleh-2012-APSA} contain noise reduction. These noise reduction methods have both manual methods and image processing algorithms. Image enhancement technology is used in the pre-processing of~\cite{Danping-2013-IPMS}. Besides, most works adopt image segmentation in their pre-processing to highlight the object regions and reduce the influence caused by noise. This step is vital for subsequent feature extraction. The segmentation methods contain several types, such as edge-based, region-based, and thresholding-based methods. \cite{Blackburn-1998-RDBA,Weller-2005-SCSO,Xiaojuan-2007-ANBC,Hiremath-2010-AICB} employ edge detection, Iterative thresholding, and adaptive global thresholding segmentation methods. In feature extraction, the shape, texture, and color features are the most commonly used features. For example, shape features of images are extracted in~\cite{Veropoulos-1998-IPNC,Gerlach-1998-IRSM,Hu-2006-AAQT,Xiaojuan-2007-ANBR}, in which Fourier descriptors are extracted in~\cite{Veropoulos-1998-IPNC,Hu-2006-AAQT}, area, eccentricity, circularity, and other geometrical features are used in~\cite{Gerlach-1998-IRSM}, and the invariant moment is extracted in~\cite{Hu-2006-AAQT,Xiaojuan-2007-ANBR}. The texture features of images are extracted for training classifiers in~\cite{Culverhouse-1996-ACFD,Hu-2006-AAQT,Xiaojuan-2007-ANBR,Mosleh-2012-APSA}. \cite{Balfoort-1992-AIAN,Weller-2005-SCSO,Weller-2007-TSNN} use the color features for training the networks. In the network analysis, it can be found that classical neural networks used in MIA tasks usually have shallow structures. This is because the performance of computer hardware is limited in the early years, and deep structures have numerous connections between different layers, which can result in abundant parameters and hard training problems.

\begin{landscape}\scriptsize
\renewcommand\arraystretch{1.28}
\setlength{\tabcolsep}{1.0pt}
\centering
\begin{longtable}{cccccccccccccccccccccccc}

\caption{Summary of the MIA works based on deep neural network. (Classification (C), Feature Extraction (FE), Segmentation (S), Detection (D), Counting (Ct), Data Augmentation (DA))}
\endfirsthead
\caption[l]{Continue: Summary of the MIA works based on classical neural network.}
\endhead

\hline
Year & Task  & Ref.                        & Data & Object                                                   & Class                                                & Method                                                                                                                    & Result                                                                                                                                             \\ \hline
1999 & C     & \cite{Dollfus-1999-FNNR}           & Private
 & Plankton                                                 & 13                                                   & CNN                                                                                                                       & Recognition rate = 86\%                                                                                                                            \\
2004 & C     & \cite{Beaufort-2004-ARCD}          & Private                                                             & Plankton                                                 & 11                                                   & CNN                                                                                                   & Recognition rate = 96\%                                                                                                                            \\
2015 & FE     & \cite{Al-2015-PEHC}                & SIPPER                                                              & Plankton                                                 & 7                                                    & CNN+SVM                                                                                                                   & CNN+SVM accuracy = 96.70\%                                                                                                                         \\
2015 & FE\&C  & \cite{Nie-2015-ADFB}               & Private                                                             & Bacteria                                                 & 17                                                   & \begin{tabular}[c]{@{}c@{}}CDBN+SVM\\ CNN\end{tabular}                                                                    & \begin{tabular}[c]{@{}c@{}}FE: CDBN+SVM accuracy  = 97.14\%\\ C: Accuracy, Precision, Recall = 62.1\%, 83.76\%, 82.16\%\end{tabular}                \\
2016 & C     & \cite{Dai-2016-AHCN}               & WHOI-Plankton                                                       & Plankton                                                 & 30                                                   & Hybrid CNN based on AlexNet                                                                                               & Accuracy = 95.83\%                                                                                                                                 \\
2016 & C     & \cite{Lee-2016-PCIL}               & WHOI-Plankton                                                       & Plankton                                                 & 103                                                  & Transfer learning based on CNN                                                                                                                       & \begin{tabular}[c]{@{}c@{}}Accuracy = 92.80\%\\ F1-score = 33.39\%\end{tabular}                                                                    \\
2016 & C     & \cite{Py-2016-PCDC}                & Plankton Set 1.0                                                    & Plankton                                                 & \textbackslash{}                                     & CNN                                                                                                                       & Softmax loss = 61.30\%                                                                                                                             \\
2017 & S     & \cite{Aydin-2017-CBTC}             & Private                                                             & Yeast                                                    & 2                                                    & CNN based on SegNet                                                                                                                    & mIOU = 71.72\%                                                                                                                                     \\
2017 & D\&C  & \cite{Hung-2017-AFRO}              & Private                                                             & Malaria parasite                                         & 8                                                    & \begin{tabular}[c]{@{}c@{}}Faster RCNN\\ AlexNet\end{tabular}                                                             & \begin{tabular}[c]{@{}c@{}}D: Accuracy = 59\%\\ C: Accuracy = 98\%\end{tabular}                                                                    \\
2017 & C     & \cite{Pedraza-2017-ADCA}           & Private                                                             & \textit{Diatom}                                          & 80                                                   & Transfer learning based on AlexNet                                                                                                                   & Average accuracy = 99.51\%                                                                                                                         \\
2017 & C     & \cite{Yan-2017-AMEC}               & Plankton Dataset                                                    & Plankton                                                 & 121                                                  & CNN                                                                                                                       & \begin{tabular}[c]{@{}c@{}}Top-1 accuracy = 76.40\%\\ Top-5 accuracy = 96.10\%\end{tabular}                                                        \\
2018 & C     & \cite{Al-2018-IPIC}                & SIPPER                                                              & Plankton                                                 & 77                                                   & CNN based on VGG-16                                                                                                       & Accuracy = 80.54\%                                                                                                                                 \\
2018 & C     & \cite{Cui-2018-TSIF}               & WHOI-Plankton                                                       & Plankton                                                 & 103                                                  & Hybrid CNN based on AlexNet                                                                                               & Accuracy = 94.32\%                                                                                                                                 \\
2018 & C     & \cite{Kim-2018-AIBU}              & Private                                                             & Bacteria                                                 & 7                                                    & \begin{tabular}[c]{@{}c@{}}DenseNet\\ WRN\end{tabular}                                                                                                                                                                          & \begin{tabular}[c]{@{}c@{}}DenseNet: Accuracy = 85\%\\ WRN: Accuracy = 73.2\%\end{tabular}                                                         \\
2018 & FE     & \cite{Kosov-2018-EMCU}             & EMDS-4                                                              & Microorganism                                            & 20                                                   & DeepLab-VGG-16                                                                                                            & \textbackslash{}                                                                                                                                   \\
2018 & C     & \cite{Liu-2018-DPRN}               & WHOI-Plankton                                                       & Plankton                                                 & 103                                                  & PyramidNet                                                                                                                & \begin{tabular}[c]{@{}c@{}}Accuracy = 86.3\%\\ F1-score = 41.64\%\end{tabular}                                                                     \\
2018 & C     & \cite{Luo-2018-APIA}               & Kaggle                                                              & Plankton                                                 & 108                                                  & CNN                                                                                                                       & \begin{tabular}[c]{@{}c@{}}Average accuracy = 84\% \\ Average recall = 40\%\end{tabular}                                                           \\
2018 & S     & \cite{Matuszewski-2018-MATS}       & Rift Valley virus                                                   & Virus                                        & 2                                                    & CNN based on U-Net                                                                                                        & \begin{tabular}[c]{@{}c@{}}Dice = 90\%\\ IOU = 83.1\%\end{tabular}                                                                                 \\
2018 & D     & \cite{Pedraza-2018-LPCN}           & Private                                                             & \emph{Diatom}                                                   & 10                                                   & \begin{tabular}[c]{@{}c@{}}RCNN\\ YOLO\end{tabular}                                                                                                                                                                                                                                                                                          & \begin{tabular}[c]{@{}c@{}}RCNN: F1-score = 10\%\\ YOLO: F1-score = 72\%\end{tabular}                                                              \\
\hline
Year & Task  & Ref.                        & Data & Object                                                   & Class                                                & Method                                                                                                                    & Result                                                                                                                                             \\ \hline

2018 & FE     & \cite{Rodrigues-2018-ETLS}         & Private                                                             & Plankton                                                 & 20                                                   & Transfer learning based on DeepSea                                                                                                               & Accuracy = 84\%                                                                                                                                    \\

2018 & D     & \cite{Tahir-2018-AFSD}             & Private                                                             & Fungus                                                   & 6                                                    & CNN                                                                                                                       & Accuracy = 94.8\%                                                                                                                                  \\
2018 & C     & \cite{Wahid-2018-CMIB}             & Multiple sources                                                    & Bacteria                                                 & 5                                                    & CNN based on Inception-V1                                                                                                 & Accuracy approx 95\%                                                                                                                               \\
2018 & C     & \cite{Wang-2018-TPCN}              & WHOI-Plankton                                                       & Plankton                                                 & 103                                                  
&\begin{tabular}[c]{@{}c@{}}Transferred parallel neural\\ network based on VGG-16\end{tabular}                                                                                                                                                                                                                                                                                         & \begin{tabular}[c]{@{}c@{}}Accuracy = 94.98\%\\ F1-score = 54.44\%\end{tabular}                                                                    \\

2019 & FE     & \cite{Ahmed-2019-CDCN}             & Multiple sources                                                    & Bacteria                                                 & 7                                                    & Inception-V3 + SVM                                                                                                        & Accuracy $\approx$ 96\%                                                                                                                               \\

2019 & C     & \cite{Balagurusamy-2019-DDBD}      & Private                                                             & Bacteria                                                 & 2                                                    & CNN                                                                                                                       & Accuracy = 83\%                                                                                                                                    \\
2019 & C     & \cite{Bliznuks-2020-ENNS}          & Private                                                             & Microorganism                                            & 2                                                    & 3D CNN                                                                                                                    & Accuracy = 95\%                                                                                                                                    \\
2019 & FE     & \cite{Cheng-2019-ECNN}             & Private                                                             & Plankton                                                 & 7                                                    & ResNet50 + SVM                                                                                                            & \begin{tabular}[c]{@{}c@{}}Accuracy = 94.52\%\\ Recall = 94.13\%\end{tabular}                                                                      \\
2019 & C     & \cite{Devan-2019-DHCT}             & Private                                                             & \textit{Herpesvirus}                                     & 2                                                    & Transfer learning based on ResNet                                                                                                                  & \begin{tabular}[c]{@{}c@{}}Accuracy = 95.44\%\\ F1-score = 95.22\%\end{tabular}                                                                    \\
2019 & FE     & \cite{Lumini-2019-DLTL}            & \begin{tabular}[c]{@{}c@{}}WHOI\\ ZooScan\\ Kaggle\end{tabular}     & Plankton                                                 & \begin{tabular}[c]{@{}c@{}}22\\ 20\\ 38\end{tabular} & Ensemble of CNN models                                                                                                    & \begin{tabular}[c]{@{}c@{}}WHOI: Accuracy = 95.27\%\\ ZooScan: Accuracy = 88.26\%\\ Kaggle: Accuracy = 94.13\%\end{tabular}                        \\
2019 & C     & \cite{Polap-2019-BSCU}             & \textbackslash{}                                                    & Bacteria                                                 & 3                                                    & CNN                                                                                                                       & \begin{tabular}[c]{@{}c@{}}Rod shaped: Classification rate \textgreater 91\%\\ Spherical shape: Classification rate \textgreater 78\%\end{tabular} \\
2019 & FE     & \cite{Rawat-2019-ADLB}             & from Internet                                                       & Plankton                                                 & 5                                                    & Inception-V3 + CNN                                                                                                        & Accuracy = 99.5\%                                                                                                                                  \\
2019 & C     & \cite{Rujichan-2019-BCUI}          & DIBaS                                                               & Bacteria                                                 & 33                                                   & MobileNetV2                                                                                                               & Average accuracy = 95.09\%                                                                                                                         \\
2019 & C     & \cite{Wahid-2019-DCNN}             & Multiple sources                                                    & Bacteria                                                 & 7                                                    & CNN based on Xception                                                                                                     & Accuracy = 97.5\%                                                                                                                                  \\
2020 & C\&Ct & \cite{Baek-2020-IECS}              & Private                                                             & Cyanobacteria                                            & 5                                                    & \begin{tabular}[c]{@{}c@{}}Fast RCNN\\ CNN\end{tabular}                                                                   & \begin{tabular}[c]{@{}c@{}}C: Average accuracy = 90.22\%\\ Ct: R2, RMSE = 0.85,  23\end{tabular}                                                   \\
2020 & FE     & \cite{Cheng-2020-MTCN}             & \begin{tabular}[c]{@{}c@{}}In situ plankton\\ CIFAR-10\end{tabular} & Plankton                                                 & \textbackslash{}                                     & Densenet201 + Polar + SVM                                                                                                 & \begin{tabular}[c]{@{}c@{}}In situ plankton: Accuracy, Recall = 97.989\%, 97.986\%\\ CIFAR-10: Accuracy, Recall  = 94.91\%, 94.76\%\end{tabular}   \\
2020 & C     & \cite{Chopra-2020-NMBC}            & \textbackslash{}                                                    & Bacteria                                                 & 4                                                    & CNN                                                                                                                       & \begin{tabular}[c]{@{}c@{}}Accuracy = 94.9\%\\ F1-score = 95.6\%\end{tabular}                                                                      \\
2020 & S     & \cite{Dietler-2020-ACNN}           & Private                                                             & Yeast                                                    & 2                                                    & U-Net                                                                                                                     & Accuracy =  94\%                                                                                                                                   \\

2020 & S\&C  & \cite{Kang-2020-CFBU}              & Private                                                             & Bacteria                                                 & 5                                                    & \begin{tabular}[c]{@{}c@{}}U-Net\\ 1D CNN\end{tabular}                                                                    & \begin{tabular}[c]{@{}c@{}}S: Average accuracy, mIOU= 96\%, 88\%\\ C: Accuracy = 90\%\end{tabular}                                                 \\

\hline
Year & Task  & Ref.                        & Data & Object                                                   & Class                                                & Method                                                                                                                    & Result                                                                                                                                             \\ \hline

2020 & C     & \cite{Kang-2020-SCFP}              & Private                                                             & Bacteria                                                 & 5                                                    & \begin{tabular}[c]{@{}c@{}}Fusion-Net based on LSTM, \\ ResNet, and 1D CNN\end{tabular}                                                                                                                                                                                   & Accuracy =  98.4\%                                                                                                                                 \\

2020 & S     & \cite{Li-2020-MAMR}                & EMDS-5                                                              & Microorganism                                            & 2                                                    & MRFU-Net                                                                                                                  & \begin{tabular}[c]{@{}c@{}}Dice = 87.23\%\\ Jaccard = 79.74\%\\ Accuracy = 97.30\%\end{tabular}                                                    \\
2020 & C     & \cite{Mhathesh-2020-A3CN}          & Private                                                             & Bacteria                                                 & 2                                                    & CNN                                                                                                                       & Accuracy = 95\%                                                                                                                                    \\
2020 & S     & \cite{Serrao-2020-ABDL}            & Private                                                             & Tuberculosis                                             & 2                                                    & CNN                                                                                                                       & Accuracy = 99.665\%                                                                                                                                \\
2020 & C     & \cite{Swetha-2020-CNNB}            & Private                                                             & Tuberculosis                                             & 3                                                    & CNN                                                                                                                       & \begin{tabular}[c]{@{}c@{}}Accuracy = 99.5\%\\ Sensitivity = 94.7\%\\ Specificity = 99\%\end{tabular}                                              \\
2020 & C     & \cite{Tamiev-2020-ACBC}            & Private                                                             & Bacteria                                     & 11                                                   & CNN                                                                                                                       & Accuracy =  86\%                                                                                                                                   \\

2020 & C     & \cite{Treebupachatsakul-2020-MIRB} & Multiple sources                                                    & \begin{tabular}[c]{@{}c@{}}Bacteria\\ Yeast\end{tabular} & 4                                                    & LeNet                                                                                                                     & Accuracy \textgreater 80\%                                                                                                                         \\

2020 & DA    & \cite{Xu-2020-AEFG}                & EMDS-5                                                              & Microorganism                                            & 21                                                   & GAN                                                                                                                       & \textbackslash{}                                                                                                                                   \\
2020 & C     & \cite{Zawadzki-2020-DLAC}          & Multiple sources                                                    & \begin{tabular}[c]{@{}c@{}}Bacteria\\ Fungi\end{tabular} & 4                                                    & \begin{tabular}[c]{@{}c@{}}Xception\\ ResNet-50\\ Inception-V3\\ MobileNetV2\\ DenseNet201\\ VGG-16\\ VGG-19\end{tabular} & Xception and ResNet-50 perform well                                                                                                                \\
2020 & S     & \cite{Zhang-2020-AMCF}             & EMDS-5                                                              & Microorganism                                            & 2                                                    & CNN-CRF                                                                                                                   & \begin{tabular}[c]{@{}c@{}}Dice = 87.13\%\\ Jaccard = 79.74\%\\ Accuracy = 96.91\%\end{tabular}                                                    \\ \hline

\label{table4}
\end{longtable}
\end{landscape}

In the other tasks, the workflow is different from the classification tasks. Image segmentation is usually performed by conventional image processing algorithms or manual methods in the classification tasks, but ANNs are applied to perform it in the segmentation tasks. For example, the SOM is applied to segment yeast-like fungus images in~\cite{Shabtai-1996-MMMC}, and HMLP is used to perform the Zeihl-Neelsen tissue slide image segmentation in~\cite{Osman-2010-STBZ}. In addition, different from the shape and texture features used in the classification tasks, some works adopt ANNs to generate the features for the subsequent analysis in the feature extraction tasks. For instance, in~\cite{Zhu-2010-BCUN}, PCNN extracts the entropy sequence features for the classification based on Euclidean distance.

\begin{figure}[htbp!]
\centering
\includegraphics[width=0.85\linewidth]{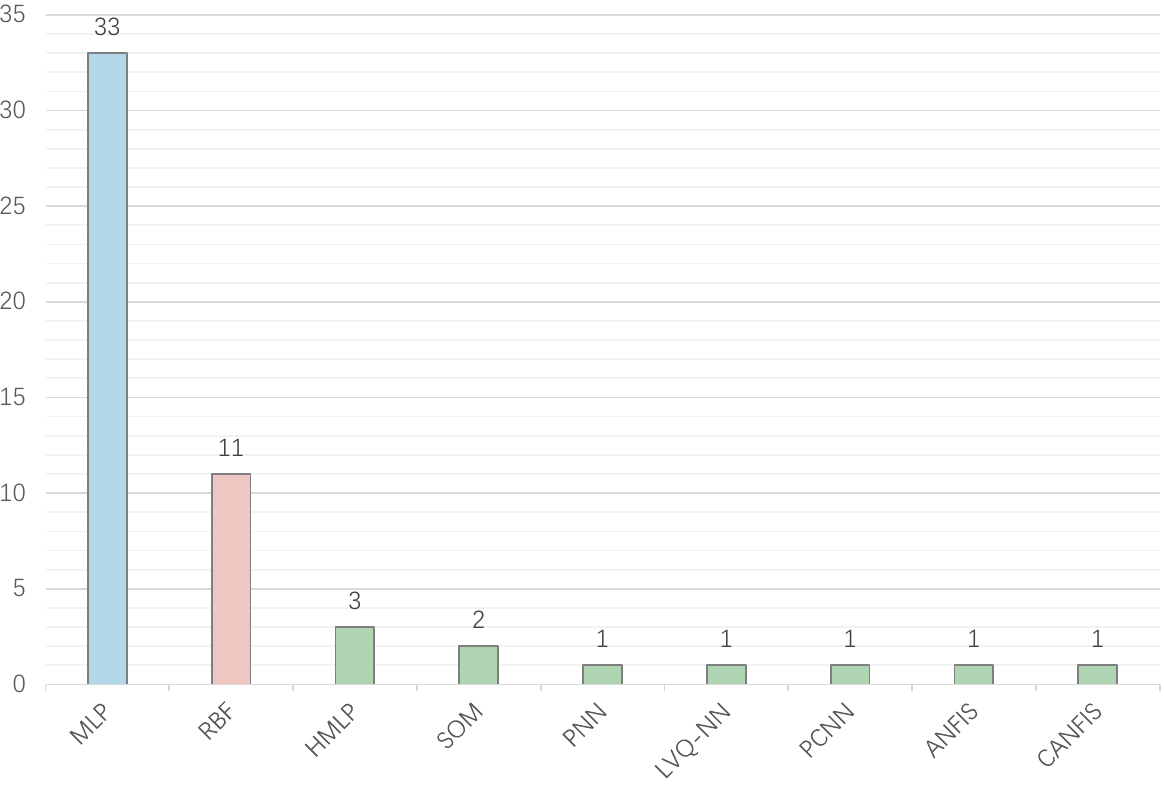}
\caption{The statistics of the classical neural networks used in MIA task.}
\label{fig63}
\end{figure}

After discussing MIA tasks based on classical neural networks, the statistic analysis is presented. As the statistic shown in Fig.~\ref{fig63}, we can find that MLP and RBF are the most widely used networks. The basic information and structure of MLP are provided in Sec.~\ref{MLP}. Among the MLP structures used in MIA, MLP with three layers is the most famous structure. This is because MLP with three layers is suitable for classifying nonlinearly separable patterns and approximating functions~\cite{Mirjalili-2011-MOAT}. It has been proven that three-layer MLPs can approximate any continuous and discontinuous function~\cite{Chen-1994-ARBL}. Besides, MLP is the most widely used network in early research about ANNs~\cite{Pistikopoulos-2011-ESCA}. RBF is an ANN that uses the radial basis function as the activation function. The output of RBF is a linear combination of radial basis functions of the inputs and neuron parameters. It also is one of the most famous classical neural networks due to its short training process and efficient mapping any nonlinear input-output relationships~\cite{Behera-2010-WTST}.

In conclusion, it can be found that most MIA works based on classical neural networks adopt feature engineering. The performance relies not only on the networks but also on feature extraction and selection. Therefore, this character makes it difficult to directly transplant the methods developed in these works to different microorganism analysis tasks. However, there are also cases where the classical neural network is directly applied to the microorganism image without any feature engineering among these classical neural networks. For example, a feed-forward MLP used in~\cite{Wit-1998-AAAN} does not rely on feature engineering but is directly applied to the image for training and classification. This innovation provides a new idea for the subsequent microorganism image analysis based on CNN.

\subsection{Analysis of Methods Based on Deep Neural Network}
As the overviews of works related to MIA based on deep neural networks are introduced in Sec.~\ref{DNN}, we can find that the analysis tasks include classification, segmentation, detection, counting, feature extraction, and data augmentation. Unlike the works related to MIA based on classical neural networks, the deep neural networks are mainly CNNs, which can directly extract the potentially useful feature from the image data by the convolutional filter. That means MIA based on deep neural networks is not limited by feature engineering. Therefore, the CNN structure plays a vital role in related task performance.

\begin{figure}[htbp!]
\centering
\includegraphics[width=0.85\linewidth]{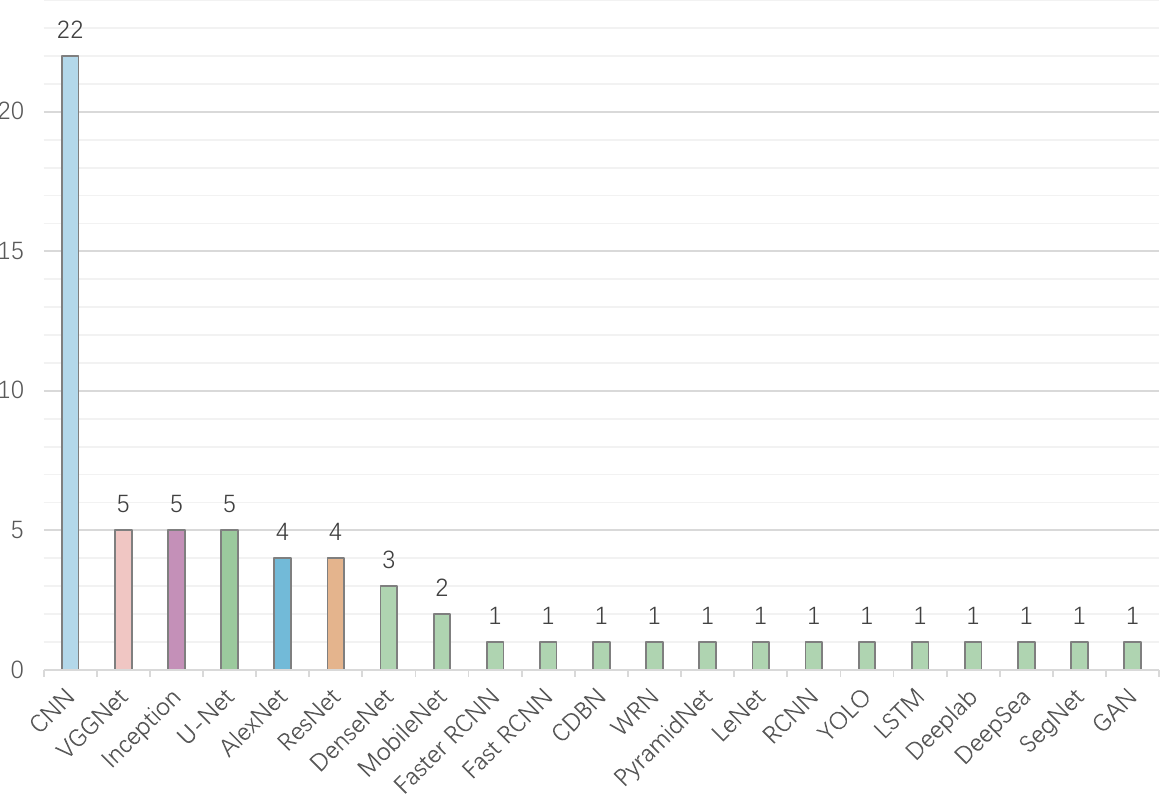}
\caption{The statistics of the deep neural networks used in MIA task.}
\label{fig64}
\end{figure}

A statistic of the deep neural networks used in MIA tasks is conducted in Fig.~\ref{fig64}. In this statistic, we divide the self-designed CNN into the CNN category. The network optimized based on the public network is divided into the original category of this public network. As we can find from Fig.~\ref{fig64}, CNNs are the most widely used. Among them, some characteristic self-designed networks are proposed. For example, the motor modules proposed in the parallel neural network can dynamically achieve translation, rotation, dilatation, contrast, and symmetry of the images by training in~\cite{Beaufort-2004-ARCD}, and a CNN is proposed to perform not only the detection but also the classification of fungi in~\cite{Tahir-2018-AFSD}. CNN can be widely applied in MIA due to the proposal of convolutional filter, which can directly take image data as input and no longer need manual feature extraction. Moreover, the parameter number of CNN is usually less than that of a fully connected neural network with the same depth. In addition, various optimization algorithms and training approaches are proposed to make it easier to train CNNs. This makes the self-design and training of CNNs more flexible.

When it comes to the public networks, the top five widely used networks are VGGNet, Inception, U-Net, AlexNet, and ResNet. There are some characteristic optimized networks based on these public networks. For example, a network called LCU-Net, which optimizes the original U-Net by increasing the diversity of the receptive field and decreasing the memory requirement of the network, is proposed to perform the microorganism image segmentation task in~\cite{Zhang-2020-AMCF,Zhang-2021-LANL}, and a transferred parallel neural network, which combines a pre-trained deep learning model using as a feature extractor and an untrained model, is introduced in~\cite{Wang-2018-TPCN}. These public networks can be widely used due to their outstanding performance in image analysis. AlexNet is the first CNN to win the champion place in ILSVRC 2012, and it is one of the symbols of the rise of CNN. GoogLeNet (Inception-V1) and VGGNet win the first and second places in ILSVRC 2014, respectively. ResNet is the champion in ILSVRC 2015. U-Net is one of the most popular segmentation networks. These networks have a significant influence on the development of deep neural networks. Their characteristics are provided in Sec.~\ref{RANN}.

In addition to optimizing these networks, there are also some other innovative points in the MIA tasks based on deep neural networks. For instance, in~\cite{Cui-2018-TSIF}, the original plankton images are converted into texture and shape images to improve the feature diversity. These three kinds of images are concatenated for the subsequent AlexNet training. Besides, in~\cite{Cheng-2020-MTCN}, the translational and rotational features are applied to address the unrecognizable problem caused by the target angle change. The rotational features are extracted from the polar images transformed from the original images (from Cartesian to polar coordinates) by CNN.

Besides, transfer learning also plays an essential role in the MIA tasks based on deep neural networks. According to the statistic, we find more than 15 papers involving transfer learning technology. Transfer learning focuses on storing knowledge learned from one task and applying it to a different but related task~\cite{West-2007-SRPA,Doodfellow-2016-DL}. Transfer learning has many applications, such as it can be used to perform the task that has limited training data~\cite{Zhang-2020-AMCF}, and it can be applied as the feature extractor for extracting the potential high-level features~\cite{Wang-2018-TPCN}. Among the MIA tasks based on transfer learning, there are some characteristic works. For example, in~\cite{Lee-2016-PCIL}, to address the imbalanced problem, a balanced sub-dataset is made from the original dataset for pre-training the CNN, and then the original dataset is used for fine-tuning the network. In~\cite{Rodrigues-2018-ETLS}, the plankton dataset ISIIS and the public dataset ImageNet are used to pre-train the deep learning models to determine the influence of different pre-trained datasets. 

\subsection{Potential Direction}
\label{PD}

In this part, we present three potential directions from the perspectives of fusion of existing MIA methods, dataset characteristics, and advanced methods in other fields.

\subsubsection{Potential Direction Based on Existing Methods}

In existing methods, the enhanced framework of GANs is proposed to achieve the data augmentation of microorganism images in~\cite{Xu-2020-AEFG}, but it suffers from the influence caused by different object directions. To solve this problem, it uses the corresponding segmentation ground truth images to uniform the directions of these objects. This approach is limited for the datasets without segmentation ground truth images. The work~\cite{Cheng-2020-MTCN} also faces the same challenge caused by different object directions. However, another approach is proposed to address this problem. The method used in~\cite{Cheng-2020-MTCN} converts the original images (images in Cartesian coordinates) into polar images (images in polar coordinates) to reduce the influence of the object direction. Therefore, this thought can be applied to improve further the framework used in~\cite{Xu-2020-AEFG}. Besides, in~\cite{Pedraza-2017-ADCA,Li-2020-MAMR,Zhang-2020-AMCF}, data augmentation is one of the steps in their workflows, the methods used in these works are rotation and flipping. The enhanced frameworks of GANs~\cite{Xu-2020-AEFG,Xu-2020-MIAU} may be helpful in these works. Although some existing MIA methods can be fused for optimization, many advanced techniques should be imported from other computer vision fields.

\subsubsection{Potential Direction Based on Dataset Characteristics}

From the dataset's perspective, most microorganism image datasets used in MIA tasks based on classical neural networks are private, and a few open-access datasets are used in MIA tasks based on deep neural networks. Some datasets, such as the EMDS series, have limited data samples. The existing methods using the EMDS series usually applied data augmentation to solve this problem. In recent years, few-shot learning and domain adaptation have become hot topics in the research about AI. Few-shot learning aims to develop some methods from data (augment the training data set by prior knowledge), model (constrain the hypothesis space by prior knowledge), or algorithm (alter the search strategy for the parameter of the best hypothesis in hypothesis space by prior knowledge) to achieve the good performance with limited training data~\cite{Wang-2020-GAFE}. Domain adaptation aims to learn a model for the target domain by leveraging knowledge from a labeled source domain~\cite{Deng-2021-IFDU,Deng-2021-JCDF}. Although few-shot learning and domain adaptation may help solve the limitation caused by insufficient data, they still face data acquisition challenges. Ideal few-shot learning and domain adaptation algorithms can address the insufficient data problem. However, there is still a performance gap between current few-shot learning and domain adaptation technologies and conventional supervised (deep learning) algorithms.

\subsubsection{Potential Direction Based on Advanced Methods}

In recent years, the self-attention mechanism-based transformer~\cite{Vaswani-2017-AIAY}, which is widely used in natural language processing, has become a new and hot spot in computer vision. That is because CNN is concerned with detecting certain features but does not consider their positioning with respect to each other. Besides, the pooling operations used in CNN lose much valuable information, such as the precise location of the adequate feature descriptor. However, the transformer-based self-attention mechanism has a more robust ability of global information representation. Additionally, the transformer has fewer parameters and low computation but performs well with the image analysis tasks compared with CNN. Therefore, it also can be a development direction in MIA tasks.

\begin{figure}[htbp!]
\centering
\includegraphics[width=0.85\linewidth]{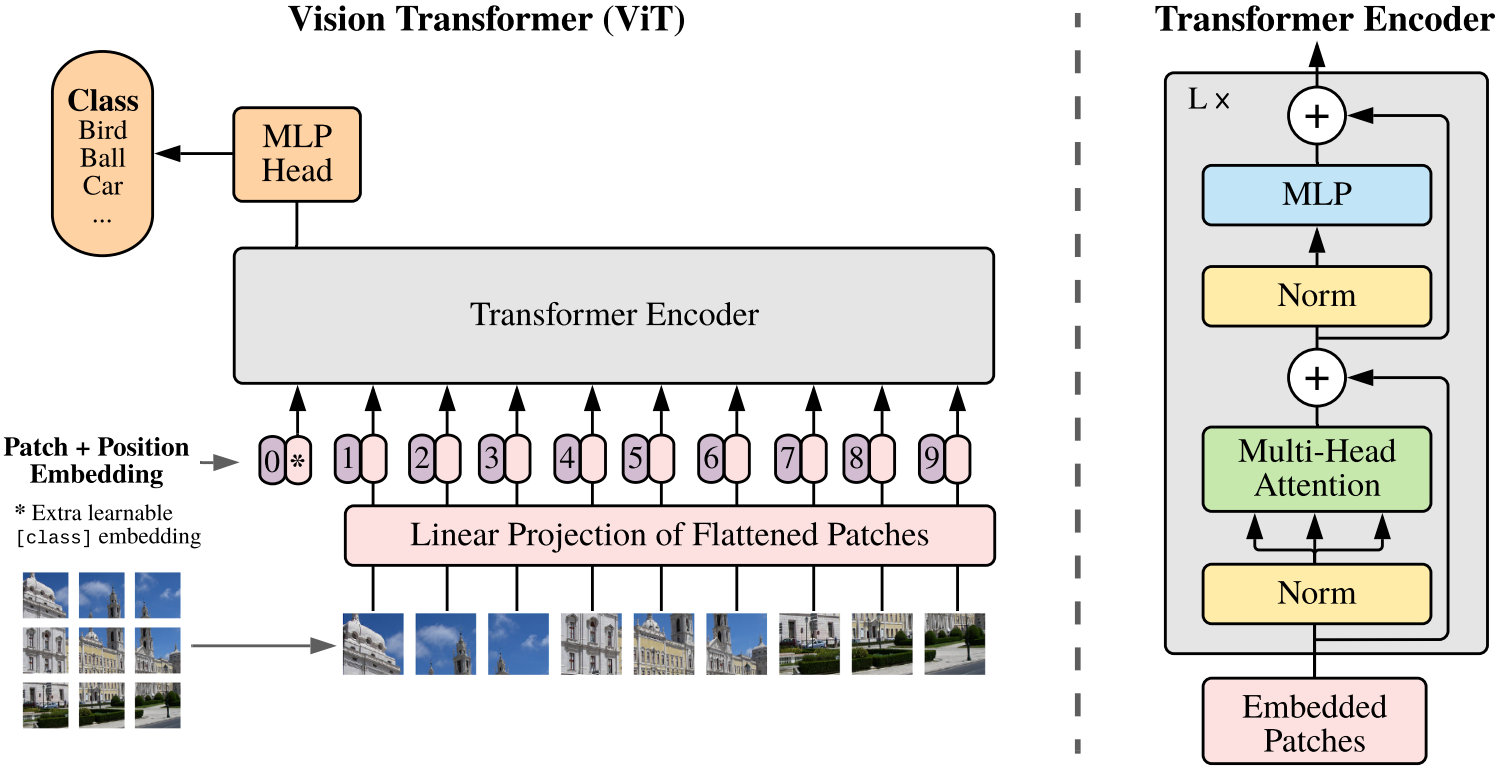}
\caption{The architecture of ViT in~\cite{Dosovitskiy-2020-AIIW}.}
\label{fig66}
\end{figure}

Considering the advantage of the transformer, \emph{Vision Transformer} (ViT) is proposed to perform the image classification task in~\cite{Dosovitskiy-2020-AIIW}. It is one of the most remarkable visual transformer methods, which directly applies sequences of image patches (with position information) as input first. The ViT projects the patches to the original transformer encoder and classifies the images with a multi-head attention mechanism as it works in natural language processing tasks. Fig.~\ref{fig66} provides the architecture of ViT. It is a potential method for microorganism image classification.

\begin{figure}[htbp!]
\centering
\includegraphics[width=0.99\linewidth]{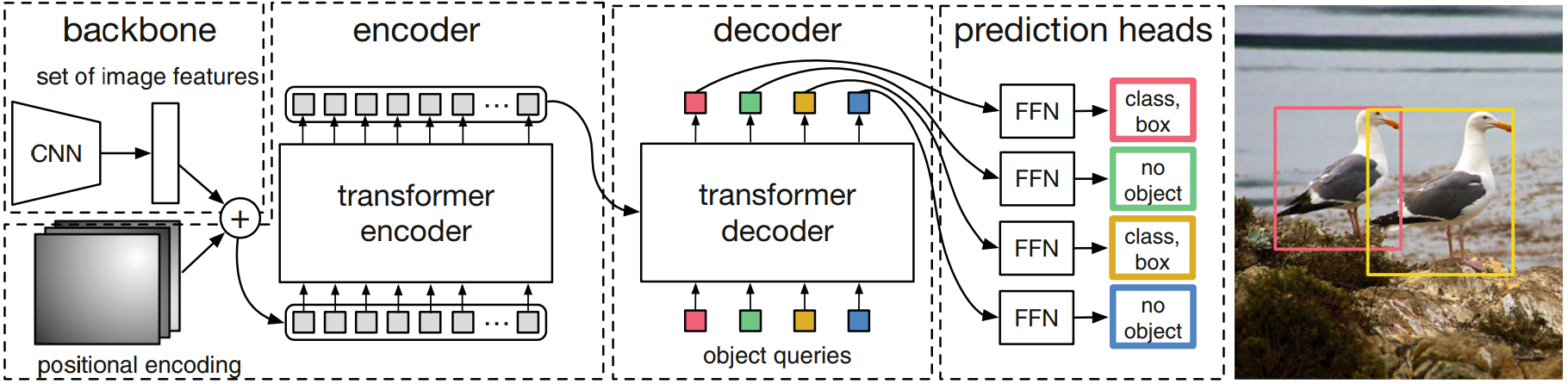}
\caption{The architecture of DETR in~\cite{Carion-2020-EODT}.}
\label{fig67}
\end{figure}

In addition to the classification task, transformer also performs well in the image detection task. In~\cite{Carion-2020-EODT}, a novel network named \emph{DEtection TRansformer} (DETR) is proposed for object detection. As the architecture of DETR is shown in Fig.~\ref{fig67}, it consists of a CNN backbone for feature extraction, an encoder-decoder transformer, and a detection prediction network based on a feed-forward network. Experiment results on COCO indicate that DETR and Faster R-CNN are comparable. It means DETR has the potential to be employed in microorganism image detection.

Although transformer shows great potential in some computer vision tasks, it still faces data acquisition challenge. The most well-known flaw of the transformer is that it requires massive training data. The original transformer is proposed to perform natural language processing tasks, and it is easier to obtain sufficient training data than computer vision tasks. Many computer vision tasks adopt pre-trained transformers in their tasks. Therefore, the applications of pure transformers in MIA still have limitations.

\section{Conclusion and Future Work}
\label{CFW}

In this paper, we conduct a review of the MIA based on classical and deep neural networks. A total of 95 papers are collected and reviewed in this paper. In Sec.~\ref{intro}, the background of MIA based on ANNs, motivation of this paper, literature collection process, and organization of this paper are introduced. In Sec.~\ref{ANN}, the development trend of ANNs and some representative networks, including MLP, VGGNet, Inception, ResNet, U-Net, and YOLO, are presented. Sec.~\ref{CNN} introduces the MIA based on classical neural networks. Related papers are summarized according to their applied tasks. Besides, a table summarized the characteristics of related papers is provided in the end. In Sec.~\ref{DNN}, the MIA based on the deep neural network is introduced from the perspectives of different tasks. The summary table of related papers is presented at the end of this section. Sec.~\ref{MA} provides analysis and statistics of methods used in MIA tasks based on classical and deep neural networks and discusses the potential development directions from the perspectives of fusion of existing methods, characteristics of datasets, and advanced methods in other fields.

Through the summary of MIA methods in this review, it can be found that this review does not have a dataset summary part. This is because most of the datasets used in related research are private data, making there no unified standard dataset for evaluating different MIA methods. Our group has been developing open-access microorganism image datasets for several years to address this limitation. Recently, several open-access microorganism image datasets have been released by our group, including EMDS-6~\cite{miamia-2021-EMDS6,Zhao-2022-ACSD}, EMDS-7~\cite{yang-2021-EMDS7}, and a SARS-CoV-2 (COVID-19) microscopic image dataset~\cite{Li-2020-ASMI}. In the future, developing more open-access microorganism image datasets and practical MIA algorithms based on ANNs is significant.

\section*{Acknowledgements}
This work is supported by the ``National Natural Science Foundation of China'' (No. 61806047) and the ``Fundamental Research Funds for the Central Universities'' (No. N2019003). We also thank Miss. Zixian Li and Mr. Guoxian Li for their important discussion in this work. Chen Li is both the co-first author and corresponding author of this paper.

\section*{Conflict of Interest}
The authors declare that they have no conflict of interest.

\bibliographystyle{spmpsci}      % mathematics and physical sciences
\bibliography{ReviewSummary}   % name your BibTeX data base

%\begin{thebibliography}{}
%%
%% and use \bibitem to create references. Consult the Instructions
%% for authors for reference list style.
%\end{thebibliography}

\end{document}